\documentclass[final]{elsarticle}

\usepackage{lineno,hyperref}
\usepackage{booktabs}
\usepackage{mathrsfs}
\usepackage{amsmath}
\usepackage{graphicx}
\usepackage{enumitem}
\usepackage{amssymb}
\usepackage{balance}
\usepackage{lscape} 
\usepackage{lipsum}  
\usepackage{url}
\usepackage{color}
\usepackage{makecell}
\usepackage{multirow}
\usepackage{array}
\newcolumntype{x}[1]{>{\centering\arraybackslash\hspace{0pt}}p{#1}}

\journal{Journal of Biomedical Informatics}

\begin{document}
\newcommand{\refalg}[1]{Algorithm \ref{#1}}
\newcommand{\refeqn}[1]{Equation \ref{#1}}
\newcommand{\reffig}[1]{Figure \ref{#1}}
\newcommand{\reftbl}[1]{Table \ref{#1}}
\newcommand{\refsec}[1]{Section \ref{#1}}

\newcommand{\reminder}[1]{\textcolor{red}{[[ #1 ]]}\typeout{#1}}
\newcommand{\reminderR}[1]{\textcolor{gray}{[[ #1 ]]}\typeout{#1}}

\newcommand{\add}[1]{\textcolor{red}{#1}\typeout{#1}}
\newcommand{\remove}[1]{\sout{#1}\typeout{#1}}

\newcommand{\m}[1]{\mathcal{#1}}
\newcommand{\bmm}[1]{\bm{\mathcal{#1}}}
\newcommand{\real}[1]{\mathbb{R}^{#1}}

\newcommand{\method}{\textsc{MedType}}

\newcommand{\ctakes}{cTAKES}
\newcommand{\scispacy}{ScispaCy}
\newcommand{\metamap}{MetaMap}
\newcommand{\metamaplite}{MetaMapLite}
\newcommand{\quickumls}{QuickUMLS}

\newcommand{\ncbi}{NCBI}
\newcommand{\cdr}{Bio CDR}
\newcommand{\sharecorpus}{ShARe}
\newcommand{\medmen}{MedMentions}
\newcommand{\wiki}{\textsc{WikiMed}}
\newcommand{\pubmed}{\textsc{PubMedDS}}

\newcommand{\basecnn}{Type-CNN}
\newcommand{\basefc}{DeepType-FC}
\newcommand{\basernn}{DeepType-RNN}
\newcommand{\baseattn}{AttentionNER}

\newcommand{\problem}{DD}
\newcommand{\problemfull}{Document Dating}

\newtheorem{theorem}{Theorem}[section]
\newtheorem{claim}[theorem]{Claim}

\newcommand{\tensor}{\mathcal{X}}
\newcommand{\Real}{\mathbb{R}}

\newcommand{\tuples}{\mathbb{T}}

\newcommand{\argmax}{arg\,max}

\newcommand\norm[1]{\left\lVert#1\right\rVert}

\newcommand{\note}[1]{\textcolor{blue}{#1}}

\newcommand*{\Scale}[2][4]{\scalebox{#1}{$#2$}}%
\newcommand*{\Resize}[2]{\resizebox{#1}{!}{$#2$}}%

\newcommand{\medlinker}{MedLinker}

%%% Tensor
%\DeclareMathAlphabet\ten{OMS}{cmsy}{b}{n} %%usage: \mathbfcal{W}
%% Matrix
\def\mat#1{\mbox{\bf #1}}%% usage: \mat{W}.

\begin{frontmatter}

\title{Improving Broad-Coverage Medical Entity Linking with Semantic Type Prediction and Large-Scale Datasets}

\author[cmu]{Shikhar Vashishth*}
\cortext[mycorrespondingauthor]{Corresponding author}
\ead{shikharvashishth@gmail.com}
\author[pitt]{Denis Newman-Griffis}
\author[cmu]{Rishabh Joshi}
\author[cmu]{Ritam Dutt}
\author[cmu]{Carolyn P. Ros\'{e}}
%\address{}

%\author[mymainaddress]{Carnegie Mellon University}

%\author[mysecondaryaddress]{University of Pittsburgh}

% \address[mic]{Microsoft Research, Bangalore, India}
\address[cmu]{Carnegie Mellon University, 5000 Forbes Ave, Pittsburgh, PA, USA}
\address[pitt]{University of Pittsburgh, 5607 Baum Blvd, Pittsburgh, PA, USA}

\begin{abstract}
\small
% Biomedical NLP tools are increasingly being applied for a variety of purposes, from clinical research to quality improvement. (Maybe start the Intro with this...)
{\bf Objectives:} Biomedical natural language processing tools are increasingly being applied for broad-coverage information extraction---extracting medical information of all types in a scientific document or a clinical note. In such broad-coverage settings, linking mentions of medical concepts to standardized vocabularies requires choosing the best candidate concepts from large inventories covering dozens of types. This study presents a novel semantic type prediction module for biomedical NLP pipelines and two automatically-constructed, large-scale datasets with broad coverage of semantic types.

\noindent{\bf Methods:} We experiment with five off-the-shelf biomedical NLP toolkits on four benchmark datasets for medical information extraction from scientific literature and clinical notes. All toolkits adopt a staged approach of mention detection followed by two stages of medical entity linking: (1) generating a list of candidate concepts, and (2) picking the best concept among them. We introduce a {\it semantic type prediction} module to alleviate the problem of overgeneration of candidate concepts by filtering out irrelevant candidate concepts based on the predicted semantic type of a mention. We present \method{}, a fully modular semantic type prediction model which we integrate into the existing NLP toolkits. To address the dearth of broad-coverage training data for medical information extraction, we further present \wiki{} and \pubmed{}, two large-scale datasets for medical entity linking.

\noindent{\bf Results:} Semantic type filtering improves medical entity linking performance across all toolkits and datasets, often by several percentage points of F-1. Further, pretraining \method{} on our novel datasets achieves state-of-the-art performance for semantic type prediction in biomedical text.

\noindent{\bf Conclusions:} Semantic type prediction is a key part of building accurate NLP pipelines for broad-coverage information extraction from biomedical text. We make our source code and novel datasets publicly available to foster reproducible research.

%Medical entity linking is the task of identifying and standardizing medical concepts referred to in an unstructured text. Most of the existing methods adopt a three-step approach of (1) detecting mentions, (2) generating a list of candidate concepts, and finally (3) picking the best concept among them. In this paper, we probe into alleviating the problem of overgeneration of candidate concepts in the candidate generation module, the most under-studied component of medical entity linking. For this, we present \method{}, a fully modular system that prunes out irrelevant candidate concepts based on the predicted \textit{semantic type} of an entity mention. We incorporate \method{} into five off-the-shelf toolkits for medical entity linking and demonstrate that it consistently improves entity linking performance across several benchmark datasets.
%To address the dearth of annotated training data for medical entity linking, we present \wiki{} and \pubmed{}, two large-scale medical entity linking datasets, which can serve as a useful resource for medical entity linking research. 
%We demonstrate that pre-training \method{} on our proposed datasets gives state-of-the-art performance on the task. We make our source code and datasets publicly available to foster reproducible research.
\end{abstract}


\end{frontmatter}

%\linenumbers

\newcommand{\showedits}[1]{\textcolor{red}{#1}}
% UNCOMMENT THIS LINE FOR THE CLEAN VERSION
\renewcommand{\showedits}[1]{#1}

\newcommand{\revtwoedits}[1]{\textcolor{red}{#1}}
% UNCOMMENT THIS LINE FOR THE CLEAN VERSION
\renewcommand{\revtwoedits}[1]{#1}

\section{Introduction}
\label{sec:introduction}

Biomedical natural language processing (NLP) tools are increasingly being applied for a wide variety of purposes, from clinical research \cite{koleck2019natural} to quality improvement \cite{young2019systematic}. One of the key ways in which these tools are used is for broad-coverage information extraction: identifying all of the biomedical concepts, of all types, that are mentioned in a given document. Several well-known biomedical NLP tools have been developed as standalone software packages and are regularly used for broad-coverage extraction in non-NLP research: for example, cTAKES \cite{ctakes} has been explored for ischemic stroke classification \cite{garg2019automating} and studying infection risk \cite{kochar2020pretreatment}; and MetaMap \cite{metamap} is frequently used in pharmacovigilance \cite{luo2017natural} and has even been adapted to health outcomes study in social media \cite{hua2020health}.

One of the central challenges in broad-coverage information extraction is the diversity of concepts in the standardized vocabularies that form the backbone of biomedical text analysis \cite{wajsburt2021jbi}. For example, the Unified Medical Language System, or UMLS \cite{umls}, Metathesaurus contains over 3.5 million unique concepts belonging to 127 different semantic types.\footnote{Counts taken from UMLS 2019AB release.} While much of the prior research on biomedical NLP methods has focused on restricted subsets of concepts, such as diseases and disorders or genes and proteins \cite{Jovanovic2017}, general-purpose tools built for arbitrary use must deal with the full breadth of concept types in reference vocabularies.

In this study, we propose {\it semantic type prediction} as a key component of general-purpose biomedical NLP pipelines. Existing pipelines generally take a multi-stage approach to information extraction that is a natural fit for integrating semantic type prediction. The first stage is {\it mention detection} (also referred to as named entity recognition, or NER), which involves identifying textual mentions corresponding to different medical concepts of interest. The second stage is {\it medical entity linking} (also referred to as medical concept normalization, or MCN \cite{Luo2019}), which can broadly be broken into two phases of {\it candidate generation}---identifying a set of standardized concepts a specific mention may refer to---and {\it disambiguation}---picking the best candidate concept for the observed mention based on the context (includes both word and phrase sense disambiguation, or WSD).

Compared to mention detection and disambiguation, candidate generation is an under-studied component of medical information extraction. Prior methods have 
historically relied on dictionary lookup and string matching \cite{metamap,ctakes} for both NER and candidate
generation, yielding high precision but incomplete coverage
\cite{Travers2006,Reategui2018}.  Recent neural methods have taken an opposite approach to the problem by using entire concept inventories as candidates, providing complete coverage at the cost of large candidate set sizes
\cite{Tutubalina2018,Soysal2018,Zhao2019a,triplet-network}. However, this approach rapidly becomes intractable when
generalizing to wider-coverage vocabularies. Thus, robust strategies to reduce \textit{overgeneration} of candidates are required to leverage the high coverage afforded by neural approaches for a broad-coverage setting.

In addition to cataloguing known surface forms for medical concepts, the UMLS Metathesaurus also assigns each concept one or more semantic types; these types present a significant and under-utilized resource for balancing coverage with candidate set size in medical entity linking. In addition to limiting the set of candidate concepts in full-inventory approaches, semantic type information can reduce problems of ambiguity in text \cite{amb1,amb2,newman-griffis2020jamia}.  For example,  the string \textit{cold} can refer to \textit{common cold} (disease), \textit{cold temperature} (natural phenomena), or \textit{cold brand} (pharmacologic substance) in different contexts. Semantic type prediction can thus inform both full-inventory and dictionary-based approaches to medical entity linking.

Identifying the semantic type of mentions has previously been shown to improve entity linking performance in Wikipedia \cite{deeptype}. 
However, this idea has not yet been systematically explored for medical entity linking, in part due to the dearth of annotated training data for the task. Curation of new biomedical text datasets faces significant barriers in the difficulty and cost of finding expert annotators \cite{med_data_annotation} as well as the confidentiality and privacy issues inherent in sharing medical data \cite{med_data_issues}. These problems are only compounded in the broad-coverage setting, where data must be sufficiently diverse to represent all the kinds of information users of NLP systems may be interested in.

\begin{figure}[t]
	\centering
	\includegraphics[width=\linewidth]{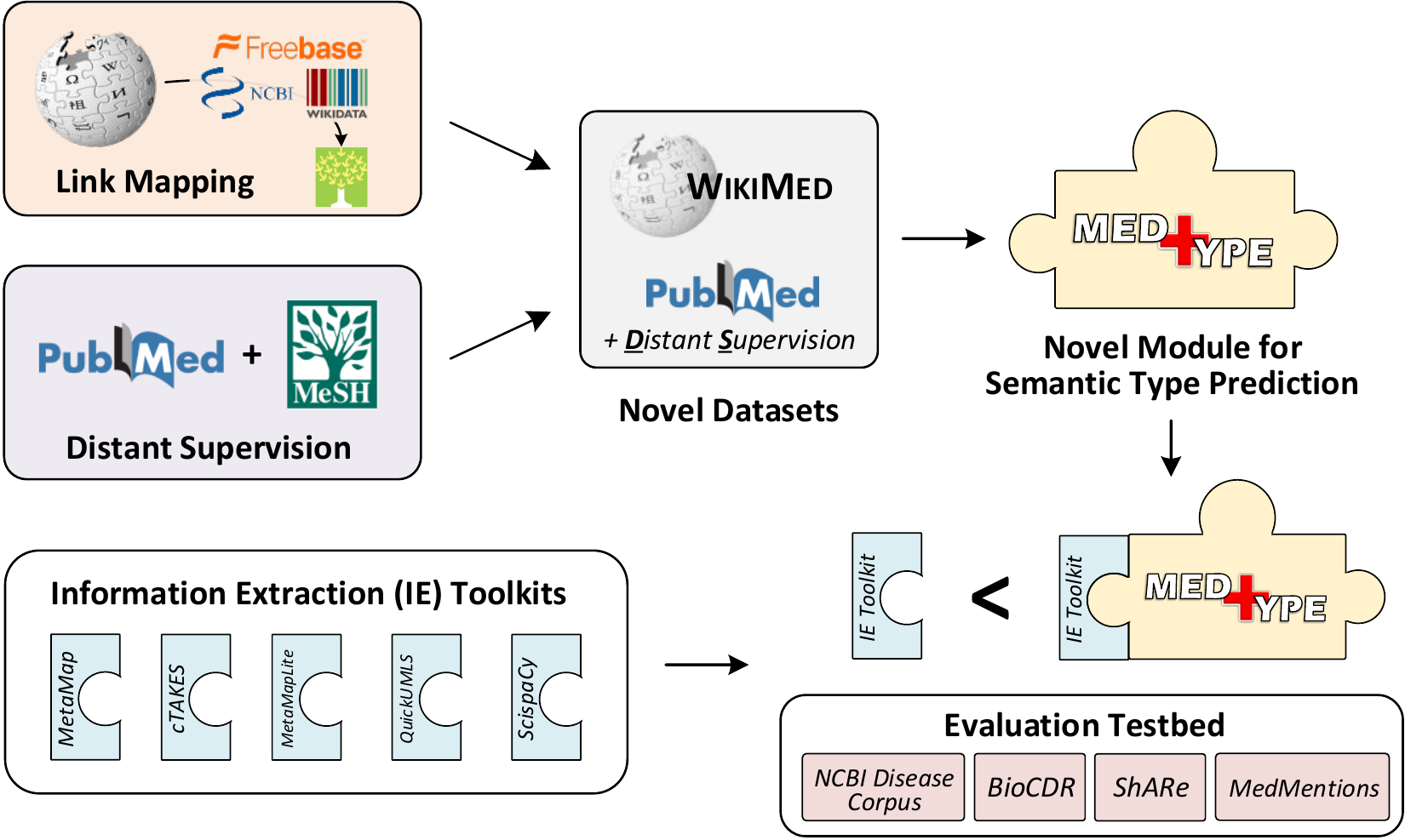}
	\caption{\label{fig:overview} \showedits{Overview of article contributions. We present \method{}, a novel, modular system for biomedical semantic type prediction, together with \wiki{} and \pubmed{}, two large-scale, automatically created datasets for medical concept normalization that we use to pretrain \method{}. We show that integrating \method{} with five commonly used packages for biomedical information extraction improves performance across the board on four benchmark datasets.}}
\end{figure}

%\begin{figure*}[t]
%	\centering
%	\includegraphics[width=\linewidth]{./images/overview_new-crop}
%	\caption{\label{fig:workflow} Overview of article contributions. We present \method{}, a novel, modular system for biomedical semantic type prediction, together with \wiki{} and \pubmed{}, two large-scale, automatically created datasets for medical concept normalization that we use to pretrain \method{}.Overview of \method{}. For a given input text, \method{} takes in the set of identified mentions along with their list of candidate concepts as input. Then, for each mention \method{} predicts its semantic type based on its context in the text. The identified semantic type is used to filter out the irrelevant candidate concepts thus controlling overgeneration of candidates and improving medical entity linking. Please refer to Section \ref{sec:overview} for details.}
%\end{figure*}

This article presents two significant innovations\showedits{, illustrated in Figure~\ref{fig:overview}}: (1) a fully modular approach to alleviating candidate set overgeneration in medical entity linking via semantic type prediction, and (2) two large-scale datasets for medical entity linking research that are freely shareable.  We make the following contributions:
\begin{itemize}[itemsep=1pt,parsep=0pt,partopsep=0pt,leftmargin=10pt,topsep=1pt]
	\item We present \method{}, a deep learning-based modular system for semantic type prediction, and incorporate it into five off-the-shelf toolkits for medical entity linking. We demonstrate that semantic type prediction consistently improves entity linking performance across several benchmark datasets.
	\item To address the dearth of annotated training data for medical entity linking, we present \wiki{} and  \pubmed{}, two automatically-created, large-scale datasets which can serve as a useful resource for medical entity linking research. Our work also demonstrates that pre-training \method{} on our proposed datasets achieves state-of-the-art performance on the semantic type prediction task.
	\item We show that type-based filtering significantly reduces the number of candidates for disambiguation, enabling further improvements in the final step of medical entity linking.
\end{itemize}

%In this paper, we present \method{}, a complete modular system for pruning out over-generated candidates for medical entity linking by predicting the \textit{semantic type} of an entity mention.  Our work makes the following contributions:
%\begin{itemize}[itemsep=1pt,parsep=0pt,partopsep=0pt,leftmargin=10pt,topsep=1pt]
%	\item We incorporate \method{} into five off-the-shelf toolkits for medical entity linking and demonstrate that it consistently improves entity linking performance across several benchmark datasets compared to existing methods.
%	\item To address the dearth of annotated training data for medical entity linking, we present \wiki{} and  \pubmed{}, two large-scale datasets which can serve a useful resource for medical entity linking research. Our work also demonstrates that pre-training \method{} on our proposed datasets gives state-of-the-art performance on the task.
	
%	\item We show that type-based filtering significantly reduces the number of candidates for WSD, enabling further improvements in the final step of medical entity linking.
%\end{itemize}

\noindent \method{}'s source code and the \wiki{} and \pubmed{} datasets proposed in this paper have been made publicly available at \url{http://github.com/svjan5/medtype}.

The remainder of this article is organized as follows. Section~\ref{sec:related_work} highlights related work in the foundational NLP methods and medical NLP literature leading to our work on semantic type filtering. Section~\ref{sec:overview} introduces semantic type filtering as a component of the medical information extraction pipeline, and presents \method{}, our state-of-the-art model for biomedical semantic type prediction. Section~\ref{sec:data_creation} describes our two novel, large-scale corpora, including quality assessments of each corpus. Section~\ref{sec:experiments} describes our experimental protocol, and Section~\ref{sec:results} presents the results of our analysis. Finally, Section~\ref{sec:discussion} discusses implications of our findings for research on broad-coverage information extraction, and Section~\ref{sec:conclusion} concludes the paper.

\section{Related Work}
\label{sec:related_work}

Information extraction is a well-studied task in NLP, and approaches often diverge between the foundational methodologies literature, which typically utilizes news wire or web text, and the medical NLP literature, which reflects adaptations to the unique characteristics of biomedical text and knowledge (e.g., specialized language, rich typologies, etc.).  In this paper, we combine recent insights from foundational methods with the rich expert resources that are central to biomedical information extraction.

%\textbf{Entity Linking:}
%Identifying named entity mentions and linking them to concepts in a knowledge base is an essential task for semantic understanding of natural text and information extraction.
Much of the research in the foundational methods literature focuses on extracting information about real-world entities and concepts (people, places, organizations, products, etc.), drawing on knowledge sources such as Freebase and Wikipedia.
In addition to jointly modeling NER and entity linking as interdependent tasks \cite{end-to-end-EL,joint-learning}, many studies leverage the rich semantics of the target knowledge base to improve linking performance \cite{ELDEN,dynamicgcn-EL}. Knowledge bases often group entities into semantic types, which inform several downstream NLP tasks such as co-reference resolution \cite{durrett2014joint}, relation extraction \cite{yaghoobzadeh-etal-2017-noise}, question answering \cite{das-QA}, and language modeling \cite{ernie}.  Recent studies have shown that fine-grained entity type prediction improves entity linking in Wikipedia text \cite{ling-etal-2015-design,deeptype}, indicating a clear potential for type prediction as a standard component of entity linking pipelines.

%\textbf{Medical Entity Linking:} 
In the biomedical domain, the role of entity type prediction in selecting suitable
candidates for medical concept mentions was recognized in some of the earliest
rule-based medical information extraction tools \cite{Aronson1994}.  However, type
prediction is typically deeply embedded in rule-based NLP tools, hampering generalizability, and discourages their use in deep learning systems.  \cite{semantictype} utilized
neural language modeling frameworks to identify the semantic type of a mention
in a medical text, but did not apply their predictions downstream; in contrast,
\cite{Medlinker} utilized approximate dictionary matching heuristics with
specialized neural language models to improve both medical entity typing and
entity linking in biomedical literature.  However, these works have not explored the efficacy of incorporating the type information within the entity linking task itself. Zhu et al.\ model mention and entity types as latent variable and jointly optimize type learning and entity disambiguation. Our work alleviates the overgeneration problem produced by both
rule-based \cite{Reategui2018} and deep learning systems in practical broad-coverage settings, by using the predicted semantic type to prune irrelevant candidates. We do so in a modular fashion, making it easy to incorporate in any entity linking architecture.

%\section{\method{} Overview}
\section{Semantic Type Prediction with \method{}}
\label{sec:overview}

\begin{figure*}[t]
	\centering
	\includegraphics[width=\linewidth]{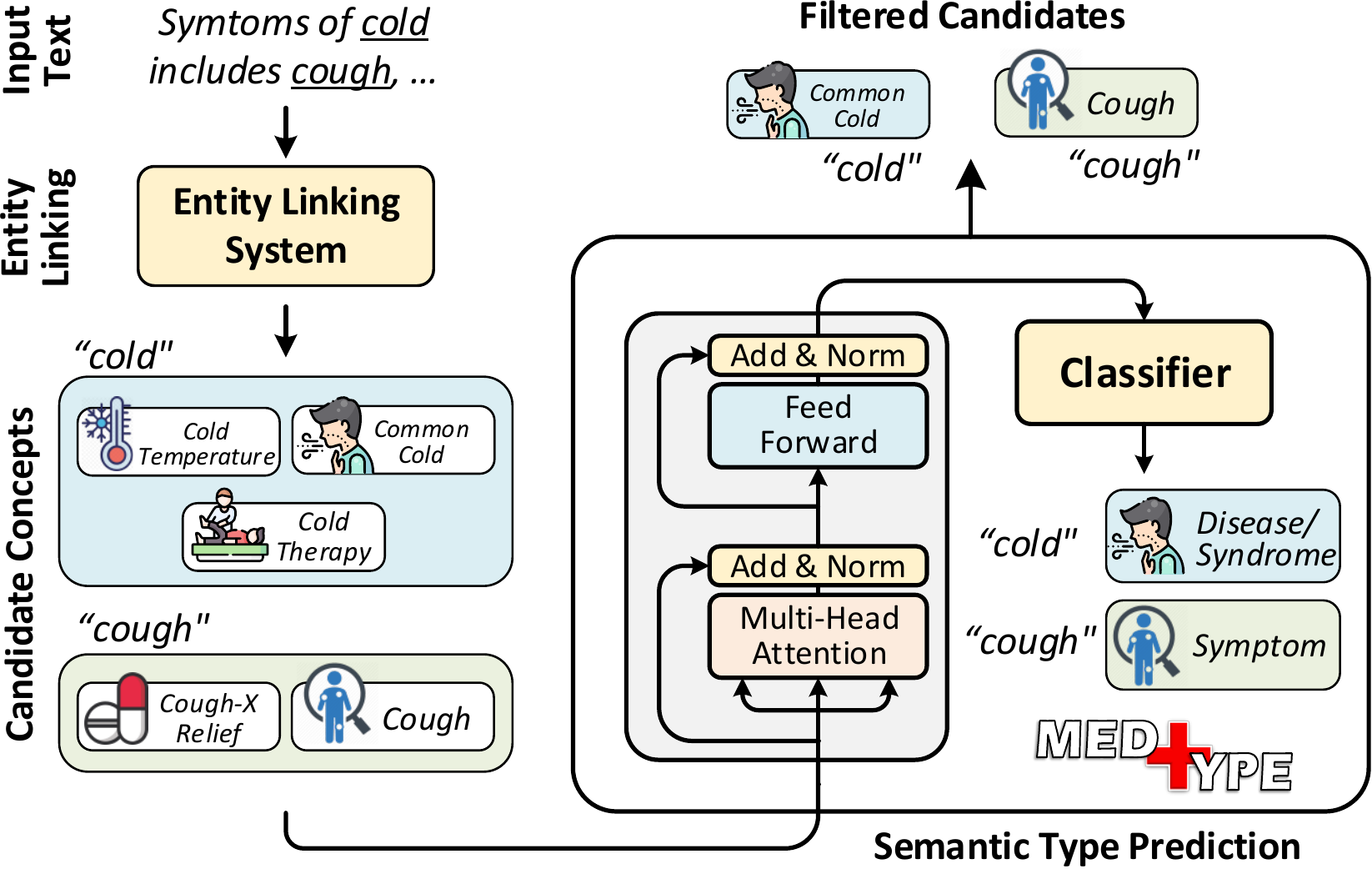}
	\caption{\label{fig:workflow} Overview of \method{}. For a given input text, \method{} takes in the set of identified mentions along with their list of candidate concepts as input. Then, for each mention, \method{} predicts its semantic type based on its context in the text. The identified semantic type is used to filter out the irrelevant candidate concepts thus controlling overgeneration of candidates and improving medical entity linking. Please refer to Section \ref{sec:overview} for details.}
\end{figure*}

Broad-coverage information extraction from biomedical text faces dual challenges of (1) a breadth of dozens of information types and millions of candidate concepts that must be considered; and (2) resolving ambiguity even for known surface forms, long recognized as challenge for off-the-shelf information extraction tools \cite{metamap} even while development of standalone disambiguation and linking models has progressed \cite{Pesaranghader2019,henry2020n2c2}.   
%In the medical domain, resolving ambiguity among possible concepts for a given mention remains a significant weakness of off-the-shelf information extraction tools \cite{metamap}.
For instance, as shown in Figure \ref{fig:workflow}, \textit{`cold'} can refer to several distinct concepts such as \textit{common cold}(disease), \textit{cold temperature} (natural phenomena), or \textit{cold brand of chlorpheniramine-phenylpropanolamine} (pharmacologic substance). This ambiguity arising from \textit{polysemy} and \textit{homonymy} leads to overgeneration of candidate concepts, exacerbated by the breadth of potential information types of interest. Thus, including an additional step to prune irrelevant candidate concepts has the potential to improve entity linking performance by simplifying the final disambiguation step. 

\showedits{In this work, we formulate semantic type prediction and filtering as a standalone module \method{} $: (\m{C},m) \rightarrow \m{C}'$, for integration into biomedical information extraction pipelines. The general type prediction and filtering process is as follows:
\begin{enumerate}
    \item \method{} takes in as input a medical entity mention $m$ and a generated set of candidate concepts $\m{C} = \{c_1, c_2, ..., c_k\}$, each of which has one or more semantic types (here, drawn from the UMLS).
    \item \method{} consists of two steps: $\method{}_{Predict} : m \rightarrow t \in T$, where $T$ is the set of all semantic types, and $\method{}_{Filter} : \m{C} \rightarrow \m{C'}$.
    \item $\method{}_{Predict}$ takes the medical entity mention $m$ and predicts the most likely semantic type $t$ of the mention.
    \item $\method{}_{Filter}$ takes the candidate set $\m{C}$ and outputs a filtered set $\m{C'} = \{c_1', c_2', \dots c_{k'}'\}$, such that $k' \leq k$ and $c_1'\dotsc_{k'}'$ are all of the predicted semantic type $t$.
\end{enumerate}
%, to enable filtering a generated set of candidate concepts $\m{C} = \{c_1, c_2, ..., c_k\}$ down to the concepts of the correct type for a given mention prior, simplifying disambiguation.  
We further present a neural implementation of \method{} as a standalone module} which can be easily integrated into existing biomedical NLP pipelines. In Figure \ref{fig:workflow}, \method{} predicts the given occurrence of \textit{`cold'} as referring to a disease, enabling pruning of the other candidates and resolving the ambiguity without the need of a dedicated disambiguation module. 
%Most of the existing medical entity linkers \cite{ctakes,metamap,scispacy} are non-neural and predominantly rely on substring matching for linking mentions to entities.
%However,
\method{} utilizes recent advances in deep learning-based language modeling techniques \cite{elmo,bert} for encoding context to predict the semantic type of a mention. The overall semantic type filtering workflow and the architecture of \method{} are shown in Figure \ref{fig:workflow}; details of the semantic type prediction task and \method{} architecture are given in the following sections.
%Further details of the method are presented in Section \ref{sec:method_details}. 

%Apart from this, we also propose two large-scale annotated datasets: \wiki{} and \pubmed{}, described in Section \ref{sec:data_creation}.. The proposed datasets are around 3$\times$ and 164$\times$ times larger than the current largest medical entity linking dataset, MedMentions \cite{medmentions}, addressing the need for large-scale training data for medical entity linking research.
%\section{\method{} Details}
%\label{sec:method_details}

\subsection{Information Extraction Problem Definition}
Formally, the task of information extraction is defined as follows. Let $\m{E}  = \{e_1, e_2, ..., e_N\}$ be a predefined set of entities in a knowledge graph and $\m{T} = (w_1, w_2, ..., w_{|\m{T}|})$ be a given unstructured text with $n$ tokens. The information extraction task involves identifying mentions $\{m_1, m_2, ..., m_k\}$ of the form $w_{i...j}$ in $\m{T}$ (mention detection phase) and mapping them to an entity $e \in \m{E}$ (entity linking phase). Following prior work \cite{quickumls,scispacy}, we define $\m{E}$ as the set of entities in the UMLS \cite{umls}. 
%,  a large-scale compendium of multiple source vocabularies which provides broad coverage of millions of biomedical concepts.
Most entity linking methods follow a two-step procedure: (1) Candidate Generation, which involves generating a probable set of candidates $\m{C}_i = \{ e^i_1, e^i_2, ..., e^i_l \ | \ e^i_j \in \m{E}\}$ for each mention $m_i$, and (2) Disambiguation (often referred to as Word/Phrase Sense Disambiguation, or WSD), which involves choosing the highest-likelihood candidate concept  $e^i_j \in \m{C}_i$.

\subsection{Candidate Pruning using Semantic Type}
\label{sec:semantic-type-filtering}

While many non-dictionary-based methods for medical entity linking have been proposed (e.g., \cite{deng-etal-2019-ensemble,Ji2020}), the most frequently-used off-the-shelf tools \cite{metamap,ctakes} for broad-coverage biomedical information extraction (as well as many recent hybrid models \cite{dsouza-ng-2015-sieve,Li2017b,wang2017pdd}) rely heavily on dictionary lookup and sub-string matching.  In the broad-coverage setting, the sheer number of medical concepts and prominence of lexical ambiguity among mentions due to \textit{homonymy} and \textit{polysemy} \cite{amb1,amb2} leads to systematic over-generation of candidate concepts.

To alleviate this problem, we utilize an intermediate step of {\it semantic type filtering}, which takes in a generated candidate set $\m{C}$ for a given mention $m$ and outputs a filtered set $\m{C}' \subseteq \m{C}$ based on the predicted semantic type of $m$. Figure~\ref{fig:workflow} illustrates this process: several irrelevant candidate concepts for the mention \textit{cold} are pruned by identifying its semantic type of \textit{Disease/Syndrome} in the given context.
%\method{}, a fully modular system that takes in a generated candidate set $\m{C}$ for a given mention and outputs a filtered set $\m{C}' \subseteq \m{C}$ as depicted in Figure \ref{fig:workflow}. Here, several irrelevant candidate concepts for the mention \textit{cold} are pruned by identifying its correct semantic type \textit{Disease/Syndrome} through \method{}.  
The semantic type of a mention is identified based on its usage in the text. For instance, in Figure \ref{fig:workflow}, based on its occurrence, the mention \textit{cough} can be interpreted as a \textit{symptom} rather than a \textit{medicine}.

\subsection{Mapping Semantic Types to Groups}
\label{sec:semantic-type-mapping}

The semantic types in the UMLS Metathesaurus present two challenges for type prediction. First, each concept may have more than one semantic type (e.g., C0250873 {\it OX7-SAP} is both a {\it Pharmacologic Substance} and an {\it Immunologic Factor}). Second, type frequencies are strongly right-tailed: for example, 907,398 concepts are of type {\it Eukaryote}, while only two UMLS concepts have type {\it Carbohydrate sequence}; these differences are exacerbated by the sparsity of fine-grained types in entity linking datasets.  To ameliorate both of these issues, we map the 127 semantic types in the UMLS Metathesaurus to 24 groups, as shown in Table~\ref{tab:type_groups}.  These groupings are derived from the UMLS semantic groups defined by \cite{sem_grouping}, with additional use of {\it is-a} relationships to split too broad groups. We use these broader groups as the labels for multi-label semantic type prediction and filtering.
%In \method{}, we use the semantic types defined in UMLS. 
%In the \method{} model introduced in this work, we use the semantic types defined in the UMLS.
%The UMLS Metathesaurus assigns each of its concepts to one or more semantic types such as \textit{Disease or Syndrome}, \textit{Anatomical Structure}, etc., out of a total of 127 types. \method{} thus models entity typing as a multi-label classification problem and utilizes the predicted types of a given mention for pruning the candidate concept set. However, these types are very fine-grained and have sparse coverage in the entity linking corpora. Thus, we use the coarse-grained Semantic Groups developed by \cite{sem_grouping} and \textit{is-a} relationships in the Semantic Network to group these 127 fine types into 24 semantic groups. We present all the semantic type groups with their members in Table \ref{tab:type_groups}.

\begin{table}[!t]
	\centering
	\footnotesize
	\resizebox{\linewidth}{!}{
		\begin{tabular}{p{0.25 \linewidth}p{0.75 \linewidth}}
			\hline
			\toprule
			\textbf{Groups}     & \textbf{Semantic Types}   \\
			\midrule
			Activities \& Behaviors	&	Activity, Behavior, Daily or Recreational Activity, Event, Governmental or Regulatory Activity, Individual Behavior, Machine Activity, Occupational Activity, Social Behavior \\
			Anatomy			& Anatomical Structure, Body Location or Region, Body Part, Organ, or Organ Component, Body Space or Junction, Body Substance, Body System, Cell, Cell Component, Embryonic Structure, Fully Formed Anatomical Structure, Tissue \\
			Chemicals \& Drugs	& Amino Acid, Peptide, or Protein, Antibiotic, Biologically Active Substance, Biomedical or Dental Material, Chemical, Chemical Viewed Functionally, Chemical Viewed Structurally, Element, Ion, or Isotope, Enzyme, Hazardous or Poisonous Substance, Hormone, Immunologic Factor, Indicator, Reagent, or Diagnostic Aid, Inorganic Chemical, Nucleic Acid, Nucleoside, or Nucleotide, Receptor, Vitamin \\
			Concepts \& Ideas	& Classification, Conceptual Entity, Group Attribute, Idea or Concept, Intellectual Product, Language, Quantitative Concept, Regulation or Law, Spatial Concept, Temporal Concept \\
			Devices			& Drug Delivery Device, Medical Device, Research Device \\
			Disease or Syndrome	& Disease or Syndrome \\
			Disorders		& Acquired Abnormality, Anatomical Abnormality, Cell or Molecular Dysfunction, Congenital Abnormality, Experimental Model of Disease, Injury or Poisoning \\
			Finding			& Finding \\
			Functional Concept	& Functional Concept \\
			Genes \& Molecular Sequences & 	Amino Acid Sequence, Carbohydrate Sequence, Gene or Genome, Molecular Sequence, Nucleotide Sequence \\
			Living Beings		& Age Group, Amphibian, Animal, Archaeon, Bacterium, Bird, Eukaryote, Family Group, Fish, Fungus, Group, Human, Mammal, Organism, Patient or Disabled Group, Plant, Population Group, Professional or Occupational Group, Reptile, Vertebrate, Virus \\
			Mental or Behavioral Dysfunction	& Mental or Behavioral Dysfunction \\
			Neoplastic Process	& Neoplastic Process \\
			Objects			& Geographic Area, Entity, Food, Manufactured Object, Physical Object, Substance \\
			Occupations		& Biomedical Occupation or Discipline, Occupation or Discipline \\
			Organic Chemical	& Organic Chemical \\
			Organizations		& Health Care Related Organization, Organization, Professional Society, Self-help or Relief Organization \\
			Pathologic Function	& Pathologic Function \\
			Pharmacologic Substance	& Clinical Drug, Pharmacologic Substance \\
			Phenomena		& Biologic Function, Environmental Effect of Humans, Human-caused Phenomenon or Process, Laboratory or Test Result, Natural Phenomenon or Process, Phenomenon or Process \\
			Physiology		& Cell Function, Clinical Attribute, Genetic Function, Mental Process, Molecular Function, Organ or Tissue Function, Organism Attribute, Organism Function, Physiologic Function \\
			Procedures		& Diagnostic Procedure, Educational Activity, Health Care Activity, Laboratory Procedure, Molecular Biology Research Technique, Research Activity, Therapeutic or Preventive Procedure \\
			Qualitative Concept	& Qualitative Concept \\
			Sign or Symptom		& Sign or Symptom \\
			\bottomrule
		\end{tabular}
	}
	\caption{Grouping of the 127 semantic types in the UMLS Metathesaurus into 24 semantic groups.  The semantic groups were derived from McCray et al.\ \cite{sem_grouping} and \textit{is-a} relationships in the Semantic Network. Refer to Section \ref{sec:semantic-type-mapping} for details.
	\label{tab:type_groups}}
\end{table} 

\subsection{\method{} Architecture} 
\method{} is a neural model for semantic type prediction in biomedical text, which is fully modular and can be included in any biomedical NLP pipeline.
\method{} takes in the input data of the form $\m{D} = [(x_0, y_0), ..., (x_N, y_N)]$ where $x_i$ denotes the mention $m_i$ and its surrounding context. The context comprises of the neighboring tokens in a window of size $k$, i.e., $Con(m_i, k) = (m_i^{-k}, ..., m_i^{-1}, m_i^{1}, ..., m_i^k)$ and $y_i$ is the semantic type. Motivated by the ability to handle polysemous tokens and superior modeling capabilities of long range dependencies of Transformer-based models \cite{transformer}, we utilize a pre-trained BERT \cite{bert} encoder and fine-tune it for our type prediction task. In our experiments, we use BioBERT \cite{biobert}, an adapted BERT model for biomedical corpora. 
We give the mention with its context, i.e.,  $(m_i^{-k}, ..., m_i^{-1}, \texttt{[MEN]}, m, \texttt{[/MEN]}, m_i^{1}, ..., m_i^k)$ as input to the encoder. Here, the special tokens \texttt{[MEN]} and \texttt{[/MEN]} are meant for providing the positional information of the mention to the model. Finally, the embedding corresponding to $\texttt{[MEN]}$ token is passed to a feed-forward classifier for the prediction of semantic types.

\section{Novel Datasets for Medical Entity Linking}
\label{sec:data_creation}
The availability of large scale public datasets helps to drive informatics research forwards \cite{imagenet,snli,Hirschberg2015}. However, curating large-scale biomedical datasets presents significant obstacles, including the expense and scarcity of relevant expertise, which largely precludes crowd-sourcing \cite{med_data_annotation}; this is compounded in the case of medical records by the challenges of maintaining patient confidentiality and privacy \cite{med_data_issues}.
To further medical entity linking research in light of these challenges, we present \wiki{} and \pubmed{}, two large-scale, automatically-created datasets for medical entity linking. We describe both the datasets in detail in the following sections.

\begin{figure}[t]
	\centering
	\includegraphics[width=\linewidth]{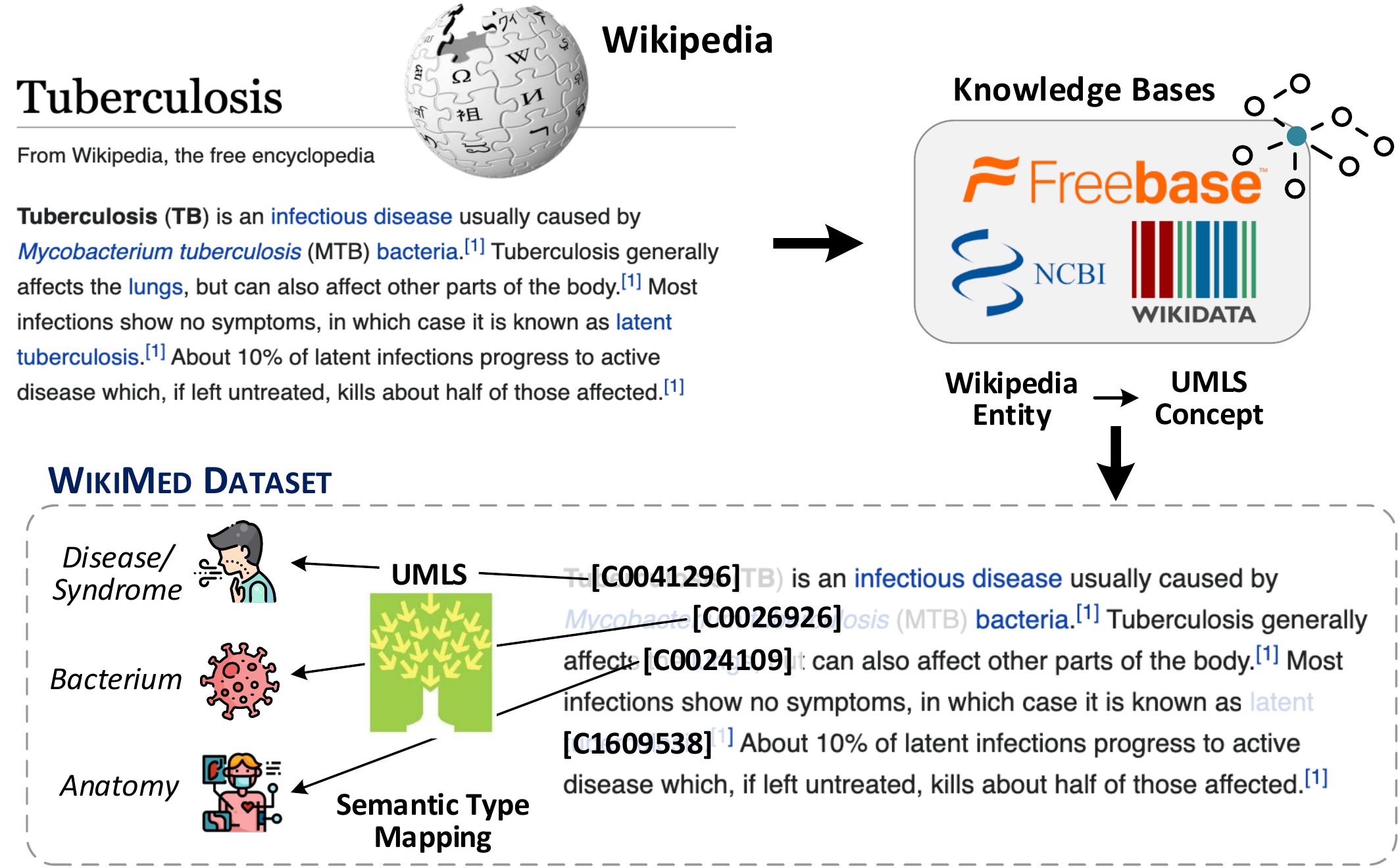}
	\caption{\label{fig:wikimed} Constructing \wiki{} from Wikipedia data. We map each linked mention in Wikipedia articles to a UMLS concept using mappings obtained from Freebase, Wikidata and NCBI knowledge bases. 
}
\end{figure}

\subsection{\wiki{}: Wikipedia-based Medical Entity Linking Corpus}
\label{sec:wiki_creation}
\textbf{\wiki{} Construction:} 
The overall steps for creating \wiki{} dataset are depicted in Figure \ref{fig:wikimed}. Wikipedia, though not restricted to medical information, includes a large number of mentions of medical concepts that can inform entity typing models.  We leverage that for constructing \wiki{} dataset. Firstly, we extract the mapping of Wikipedia pages to UMLS concepts from several existing knowledge bases such as Wikidata \cite{wikidata}, Freebase \cite{freebase}, and the NCBI Taxonomy \cite{ncbi_mapping}. This gives us a one-to-one mapping of approximately 60,500 Wikipedia pages to UMLS concepts. Since UMLS concepts are primarily biomedical in nature, this helps us identify the relevant Wikipedia pages for medical entity linking. Then, for each Wikipedia article, we linked those mentions to UMLS concepts. The Semantic Network (of UMLS) provides semantic types for each UMLS concept which we utilize for further reassigning mentions to semantic types. This results in a high-quality dataset for medical entity typing. Overall, our pipeline extracts around 1 million mentions spanning across 400k Wikipedia articles. More details of the dataset are presented in Table \ref{tab:data_stats}.  Although \wiki{} contains web text on a variety of topics, we find that it helps to improve performance on entity linking in other domains as well as shown in Section \ref{sec:results_type_pred}.

\textbf{\wiki{} Quality:} \showedits{The link structure of Wikipedia, which we utilized for creating the \wiki{} dataset, is normally treated as ground truth in information extraction and natural language processing research \cite{wiki_gtruth4,wiki_gtruth2,wiki_gtruth5,wiki_gtruth3,wiki_gtruth1}. While errors have been found in Wikipedia link structure \cite{wiki_link_error1,wiki_link_error2}, the average error rate of relational statements (including incorrect assertions and incorrect links) has been estimated to be around 2.8\% \cite{wiki_quality}, supporting the use of Wikipedia links as a sufficiently high-quality resource to yield accurate mappings. To assess the correctness of our medically-focused dataset, we randomly sampled 100 links from \wiki{} for manual verification. Three authors (SV, DNG, RJ) reviewed each sample to assess (1) whether the annotated CUI (identified via automated mapping to the UMLS) was appropriate and (2) in cases of an incorrect CUI, whether the annotated semantic type was appropriate. After resolution of disagreements, we found a CUI-level accuracy of 91\%, and a type-level accuracy of 95\% in the 100 reviewed samples. As Wikipedia links are provided a priori in the page hypertext, and not all relevant mentions of an entity are marked with links, we did not assess either precision or recall of mention detection. Thus, while \wiki{} is not appropriate for training or evaluating mention detection models, we find that it provides a high-quality silver standard resource for medical entity linking.}
%\textcolor{blue}{Clarify if this holds. 
%	Thus, we also utilize \wiki{} as one of the evaluation datasets for assessing the performance of medical entity linkers on non-medical domain text.

\wiki{} is significantly larger than previous medical entity linking datasets: $3\times$ larger than MedMentions \cite{medmentions}, and $10\times$ larger than the NCBI Disease Corpus \cite{ncbi_data}.
Moreover, \wiki{} also provides better coverage of entities from different semantic types than existing datasets, as shown in Table \ref{tab:cat_distribution}.

\begin{figure}[t]
	\centering
	\includegraphics[width=0.8\linewidth]{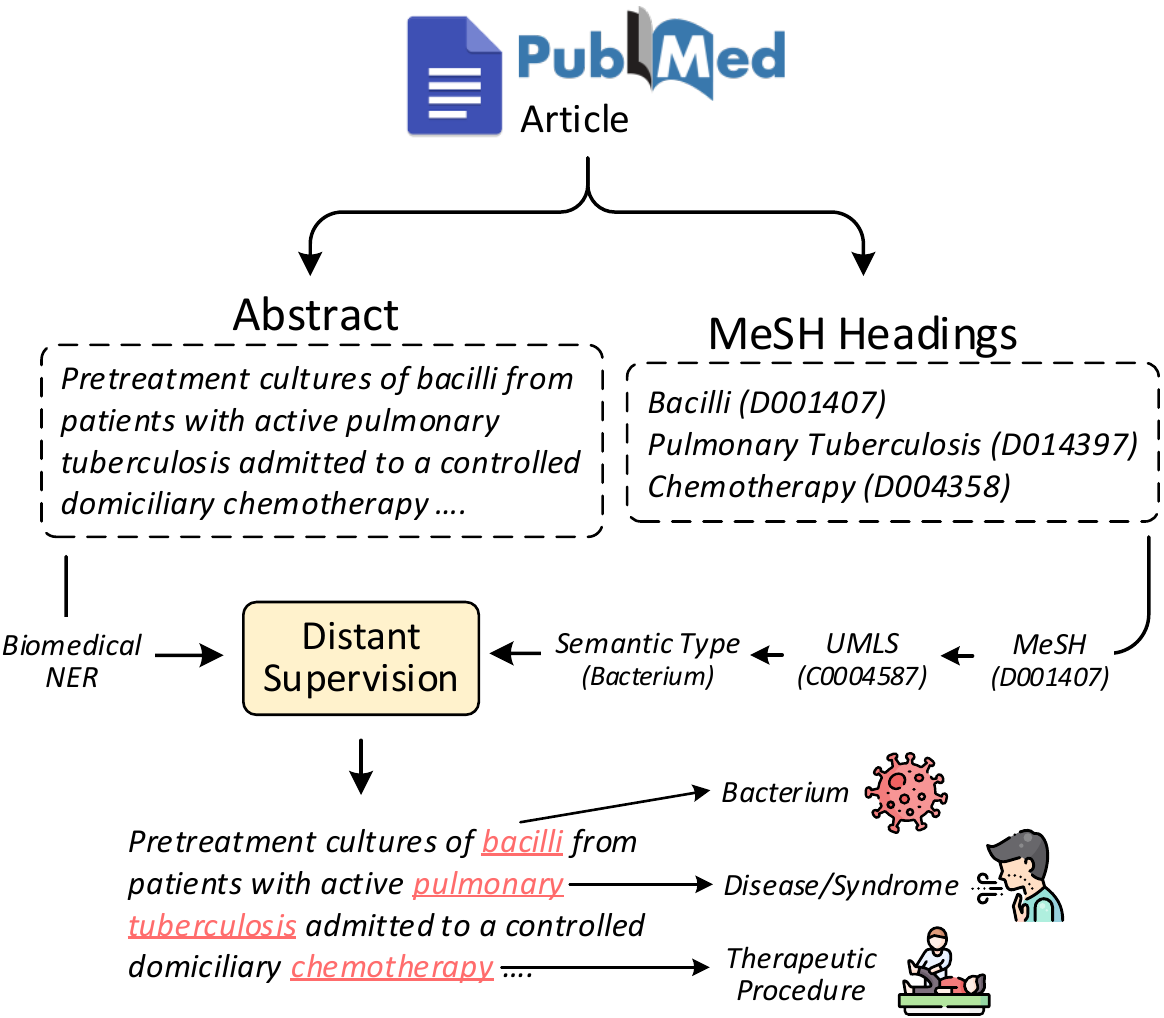}
	\caption{\label{fig:pubmed} Constructing \pubmed{} using distant-supervision on PubMed corpus. For each article, we apply biomedical NER on its abstract for obtaining relevant entity mentions which are then linked using supervision from MeSH headings of the article. Refer to Section \ref{sec:pubmed_creation} for details}
\end{figure}

\begin{table}[t]
	\centering
	%	\small
	\resizebox{\linewidth}{!}{
		\begin{tabular}{lrrrrrr}
			\hline
			\toprule
			&\multicolumn{4}{c}{\it Evaluation datasets}&\multicolumn{2}{c}{\it Novel datasets}\\
			\textbf{Categories}    & \textbf{\ncbi{}}    & \textbf{\cdr{}}    & \textbf{\sharecorpus{}}    & \textbf{\medmen{}}    & \textbf{\wiki{}}    & \textbf{\pubmed{}} \\
			\midrule
			Activities \& Behaviors             &    4      & 7      & 1      & 12,249 & 554     & 2,725,161 \\
			Anatomy                             &    3      & 29     & 4      & 19,098 & 14,366  & 10,688,138 \\
			Chemicals \& Drugs                  &    0      & 32,436 & 1      & 46,420 & 26,809  & 44,476,957 \\
			Concepts \& Ideas                   &    0      & 0      & 1      & 60,475 & 2,562   & 5,274,354 \\
			Devices                             &    0      & 0      & 0      & 2,691  & 483     & 242,599 \\
			Disease or Syndrome                 &    10,760 & 22,603 & 5,895  & 11,709 & 84,706  & 9,846,667 \\
			Disorders                           &    664    & 1,853  & 997    & 3,575  & 8,635   & 1,115,186 \\
			Finding                             &    749    & 2,220  & 500    & 15,666 & 9,285   & 1,778,023 \\
			Functional Concept                  &    0      & 0      & 1      & 23,672 & 117     & 48,553 \\
			Genes \& Molecular Sequences        &    20     & 0      & 0      & 5,582  & 446     & 281,662 \\
			Living Beings                       &    0      & 43     & 7      & 31,691 & 919,694 & 21,339,662 \\
			Mental or Behavioral Dysfunction    &    293    & 3,657  & 410    & 2,463  & 19,196  & 2,353,547 \\
			Neoplastic Process                  &    4,022  & 2,301  & 323    & 4,635  & 16,823  & 1,476,843 \\
			Objects                             &    0      & 129    & 2      & 10,357 & 421     & 5,184,355 \\
			Occupations                         &    0      & 0      & 0      & 1,443  & 1,156   & 654,604 \\
			Organic Chemical                    &    0      & 90,428 & 1      & 10,258 & 17,330  & 50,248,085 \\
			Organizations                       &    0      & 0      & 0      & 2,276  & 0       & 298,119 \\
			Pathologic Function                 &    143    & 3,290  & 2,285  & 4,121  & 4,474   & 1,895,835 \\
			Pharmacologic Substance             &    0      & 90,872 & 1      & 11,935 & 24,878  & 50,696,769 \\
			Phenomena                           &    4      & 163    & 2      & 7,210  & 317     & 1,722,873 \\
			Physiology                          &    15     & 166    & 3      & 24,753 & 2,054   & 10,674,561 \\
			Procedures                          &    5      & 73     & 4      & 37,616 & 4,008   & 7,471,434 \\
			Qualitative Concept                 &    0      & 0      & 7      & 32,564 & 106     & 1,211,747 \\
			Sign or Symptom                     &    211    & 9,844  & 2,687  & 1,809  & 4,212   & 3,750,734 \\
			\bottomrule
		\end{tabular}
	}
	\caption{Frequencies of semantic types in our evaluation datasets and novel training datasets. Overall, we find that our \wiki{} and \pubmed{} datasets give diverse coverage across all semantic types.}
	\label{tab:cat_distribution}
\end{table}

\subsection{\pubmed{}: Distantly-Supervised Biomedical Entity Linking Corpus}
\label{sec:pubmed_creation}

\textbf{\pubmed{} Construction: }
Distant supervision \cite{distant_sup} enables automatic generation of training data and has been exploited for several tasks \cite{ds_event_extraction,ds_entity_linking}, including identifying potential mentions of medical concepts \cite{ds_entity_linking_medical}. To create a large-scale training dataset for medical entity linking drawn from biomedical language, we use distant supervision on PubMed abstracts to generate \pubmed{}.  An overview of the entire process is summarized in Figure \ref{fig:pubmed}. We first run a state-of-the-art biomedical NER model \cite{scispacy} on 20 million PubMed abstracts to extract its medical entity mentions.  We then use the Medical Subject Headings (MeSH) tags assigned to each PubMed article to weakly link the extracted entity mentions to a MeSH concept. A mention is linked only when it exactly matches with the name of one of the provided MeSH headers. The UMLS provides mapping of MeSH headers to UMLS concept identifiers, which we utilize to get the semantic type of each linked mention from Semantic Network as done for mentions in \wiki{}. 
Using this procedure, we created \pubmed{}, a dataset with 58M annotated mentions, which we utilize for pre-training \method{}. The size of \pubmed{} is around 164 times larger than the current largest medical entity linking dataset, MedMentions \cite{medmentions}. Next, we demonstrate that although \pubmed{} is distantly-supervised, it has sufficiently high precision to serve as a valuable resource for medical entity linking research.

\begin{table}[h]
	\centering
	\small
	%    \resizebox{\linewidth}{!}{
	\begin{tabular}{lccc}
		\toprule
		\multicolumn{1}{c}{Documents shared with}            & Precision & Recall  \\
		\midrule
		\ncbi{}              &    86.3  & 6.5   \\
		\cdr{}              &    75.8 & 1.3 \\
		MedMentions              &    90.3 & 5.3 \\
		\bottomrule
	\end{tabular}
	%    }
	\caption{\label{tab:pubmed_prec} Quality assessment of \pubmed{}, based on the subset of documents it shares with the NCBI Disease Corpus, Bio CDR, and MedMentions. Precision and recall are calculated with respect to overlap between our automated annotations in \pubmed{} and the gold standard annotations in the comparison datasets.  We find that although \pubmed{} has low coverage, extracted mentions have high precision across the three datasets.}
\end{table}

\textbf{\pubmed{} Quality Analysis: }

Distant supervision enables large-scale text annotation but can produce noisy data \cite{ds_noise}.
In order to assess the quality of \pubmed{} as a dataset for medical entity linking, we identified the subset of documents overlapping with three manually-annotated datasets using PubMed abstracts: \medmen{} \cite{medmentions}, \ncbi{} \cite{ncbi_data}, and \cdr{} \cite{cdr_data}.  All PubMed documents annotated in these three datasets were included in \pubmed{}. This allowed us to compare the precision and recall of our distantly-supervised mentions to manual annotations.  The results of this analysis are reported in Table~\ref{tab:pubmed_prec}.  Reflecting on the strict requirements for linking a mention in our dataset (identification with a NER tool and exact match to a provided MeSH header), we find that \pubmed{} omits many of the true mentions in these documents, but the vast majority of included mentions are annotated correctly (precision of around 84\%).  Thus, while \pubmed{} would not be appropriate for training medical mention detection (NER) models, its annotations are of high quality for training entity type prediction and disambiguation models.

\section{Experimental Evaluation}
\label{sec:experiments}

Our work makes three distinct contributions to broad-coverage information extraction research: (1) a modular formulation of the semantic type prediction task, which can be easily integrated into any pipelined approach; (2) our \method{} model for semantic type prediction; and (3) our novel datasets for biomedical entity linking research.  We thus performed two types of experimental evaluations leveraging four benchmark datasets for biomedical information extraction (detailed in Section~\ref{sec:datasets}).

{\bf Semantic type prediction:} We first evaluated \method{} as a standalone model for semantic type prediction, comparing it against recent type prediction models (detailed in Section~\ref{sec:baselines}) to measure the specific improvements yielded by our approach.  We used the gold mentions annotated in each dataset directly, without use of a mention detection model.  The label for each mention was identified by mapping its annotated CUI to its semantic type(s) in the UMLS, and from there to one or more of our 24 semantic groups (described in Section~\ref{sec:semantic-type-mapping}).  We trained each type prediction model to predict these classes, using the training portion of each dataset and evaluating on the test set.

In addition, we measured the impact of our novel entity linking datasets: \wiki{} and \pubmed{} by pretraining our best performing model, \method{} on each dataset individually and on both together prior to training on each of the four evaluation datasets, and comparing type prediction performance to using \method{} without pretraining.

{\bf Information extraction:} We then evaluated the impact of using semantic type filtering as part of five widely-used biomedical information extraction pipelines (detailed in Section~\ref{sec:linkers}).  To evaluate the semantic type filtering module and our \method{} implementation separately, we experimented with three approaches for semantic type prediction:
\begin{itemize}
    \item {\bf Oracle (fine):} To evaluate the maximum possible improvement from type-based pruning of candidate concepts, we experimented with an oracle model which always filters the candidate set of entities to entities of the same type as the gold standard CUI. The {\it Fine} oracle filters based on the 127 original types in the UMLS, to control for effects of semantic grouping.
    \item {\bf Oracle (coarse):} Our {\it Coarse} oracle uses the 24 semantic groups defined in Section~\ref{sec:semantic-type-mapping}, to represent an upper bound of what can be achieved using our type prediction models.
    \item {\bf \method{}:} Finally, for a practical evaluation aligned with real-world use, we incorporate both \method{} and its strongest competitor type prediction model into the information extraction pipelines to perform semantic type filtering.
\end{itemize}

Under each of these settings, we integrate semantic type prediction into the information extraction pipeline as follows:

%We experiment with integrating \method{} into widely-used medical entity linking pipelines. In order to evaluate the contributions of \method{} to medical entity linking, we use the following procedure:
\begin{enumerate}
    \item Run biomedical information extraction tools to identify (1) mentions of medical concepts in a document; and (2) a ranked list of candidate CUIs for each mention.
    \item Use one of the above semantic type prediction approaches to predict the type of each mention, and filter the list of candidate CUIs to only CUIs of that type. %to only those of the predicted type.
    \item Return the highest-ranked CUI in the filtered candidates as the final entity linking prediction.
\end{enumerate}

%We evaluate our improved pipeline in two ways: (1) we compare against using the top-ranked CUIs predicted by the medical entity linkers without type prediction; and (2) we compare \method{} against recent methods for type prediction, to measure the specific improvements yielded by our model.

\begin{table}[t]
	\centering
	\small 
	\resizebox{\linewidth}{!}{
		\begin{tabular}{lrrrr}
			\hline
			\toprule
			\textbf{Datasets}     & \textbf{\#Documents}    & \textbf{\#Sentences}    & \textbf{\#Mentions}    & \textbf{\#Unq Concepts} \\
			\midrule
			\ncbi{}    & 792    & 7,645    & 6,817    & 1,638 \\
			\cdr{}    & 1,500    & 14,166    & 28,559    & 9,149 \\
			\sharecorpus{}    & 431    & 27,246    & 17,809    & 1,719 \\
			\medmen{}    & 4,392    & 42,602    & 352,496    & 34,724 \\
			\midrule
			\textbf{\wiki{}}    & 393,618    & 11,331,321    & 1,067,083   & 57,739 \\
			\textbf{\pubmed{} }   & 13,197,430   &   127,670,590  &   57,943,354    &  44,881 \\
			\bottomrule
		\end{tabular}
	}
	\caption{Details of the medical entity linking datasets used in our experiments; \#Unq Con refers to the number of unique CUIs in each dataset. \wiki{} is our novel automatically-annotated Wikipedia dataset, and \pubmed{} is our novel distantly supervised dataset.}% Please refer to Section \ref{sec:datasets} for more details.}
	\label{tab:data_stats}
\end{table} 

\subsection{Datasets}
\label{sec:datasets}

In our experiments, we evaluate the models on four benchmark datasets: the NCBI Disease Corpus \cite{ncbi_data}, {\cdr{}} \cite{cdr_data}, {\sharecorpus{}} \cite{sharecorpus_data}, and {MedMentions} \cite{medmentions} for medical entity linking.
%, as well as our novel \wiki{} dataset. Please note that we avoid evaluation on distantly-supervised \pubmed{} as some of its mentions might be wrongly annotated (refer to Section \ref{sec:pubmed_creation}). 
These datasets span across different text genres, such as biomedical research articles and Electronic Health Records (EHR), and information domains, allowing us to evaluate the generality of \method{} across diverse domains.
The dataset statistics and the semantic type distribution are presented in Table~\ref{tab:data_stats} and Table~\ref{tab:cat_distribution} respectively. Below, we provide a short description of each dataset.
%In our experiments, we evaluate on the following four benchmark datasets. The dataset statistics are presented in Table~\ref{tab:data_stats}.

\begin{itemize}[itemsep=1pt,parsep=0pt,partopsep=0pt,leftmargin=10pt,topsep=1pt]
	\item \textbf{\ncbi{}:}  The NCBI Disease Corpus \cite{ncbi_data}, which we refer to as \ncbi{} for brevity, consists of 793 PubMed abstracts annotated with disease mentions and their corresponding concepts in the MEDIC vocabulary \cite{MEDIC}. 
	%91\% of the mentions were linked to a single  concept, while the rest were linked to a combination.
	
	% The MEDIC vocabulary comprises two standard disease controlled vocabularies, MeSH \footnote{\url{http://www.nlm.nih.gov/mesh/}} and OMIM \footnote{\url{http://www.ncbi.nlm.nih. gov/omim}} identifiers.

	\item \textbf{\cdr{}:} The CDR corpus \cite{cdr_data} consists of 1,500 PubMed abstracts annotated with mentions of chemicals, diseases, and relations between them. These mentions were normalized to their unique concept identifiers, using MeSH as the controlled vocabulary.
	
	\item \textbf{\sharecorpus{}:} The ShARe corpus \cite{pradhan2015evaluating} is a collection of de-identified clinical notes, which was used for a series of NLP shared tasks.  We use the subset used in a 2014 shared task \cite{share-2014}, consisting of 431 documents annotated for disorder mentions and grounded to SNOMED CT.
	
	\item \textbf{\medmen{}:} The MedMentions data of  \cite{medmentions} consists of 4,392 PubMed abstracts annotated with several biomedical mentions. Each mention is labeled with a unique concept identifier and a semantic type using the UMLS as the target ontology.
	% 91.7\% of the concepts were linked to a single semantic type, 8\% to 2 types, and the rest to more than 2.
	
	%\item \textbf{\wiki{}} and \textbf{\pubmed{}} are the datasets proposed in this paper. Please refer to Section \ref{sec:data_creation} for details.
\end{itemize}

\subsection{Type Prediction Baselines}
\label{sec:baselines}
We compare \method{} against four recent neural entity typing methods. \textbf{\baseattn{}} \cite{attention_ner} utilizes attention mechanism for extracting relevant information from the context of a mention for type prediction. \textbf{\basefc{}} and \textbf{\basernn{}} are two neural network based models proposed by \cite{deeptype} for entity typing. \textbf{\basecnn{}} \cite{murty-etal-2018-hierarchical} is another neural approach which utilizes CNNs for modeling the global context of a mention for type prediction. \textbf{MedNER} \cite{Medlinker} uses NLM and dictionary mapping to predict semantic type of medical mentions.

\subsection{Biomedical Information Extraction Tools}
\label{sec:linkers}
We integrate \method{} into five widely-used tools for biomedical information extraction, each of which performs mention detection (NER) and produces a ranked list of candidate CUIs for each mention. Below, we describe each of them in brief. 

\begin{itemize}[itemsep=1pt,parsep=0pt,partopsep=0pt,leftmargin=10pt,topsep=1pt]
	\item \textbf{\metamap{}} \cite{metamap}  leverages a knowledge-intensive approach based on symbolic NLP and linguistic techniques to map biomedical mentions in text to UMLS concepts. \metamap{} was developed for indexing scientific literature.
	\item \textbf{\ctakes{}} \cite{ctakes} uses a terminology-agnostic dictionary look-up algorithm for mapping named entities to UMLS concepts. We utilize the Clinical Pipeline of \ctakes{} augmented with LVG Annotator\footnote{https://cwiki.apache.org/confluence/display/\ctakes{}/ \ctakes{}+4.0+-+LVG}. 
	\ctakes{} was developed for analyzing clinical text.
	\item \textbf{\metamaplite{}} \cite{metamaplite} re-implements the basic functionalities of \metamap{} with an additional emphasis on real-time  processing and competitive performance. 
	\item \textbf{\quickumls{}} \cite{quickumls}  is a fast, unsupervised algorithm that leverages approximate, dictionary-matching techniques for mapping biomedical entities in text. \quickumls{} was developed as a general-purpose tool and evaluated on consumer-generated texts \cite{quickumls}.
	\item \textbf{\scispacy{}} \cite{scispacy} builds upon the robust spaCy library \cite{spacy} for several biomedical and scientific text-processing applications such as parsing, named entity recognition, and entity linking. \scispacy{} was developed primarily for analyzing scientific literature.
%	\item \textbf{MedLinker} \cite{Medlinker} utilize approximate dictionary matching heuristics with specialized neural language models for medical entity typing and entity linking in biomedical literature.
\end{itemize}

We do not use the recent CLAMP \cite{Soysal2018} system in our experiments, as it does not provide access to a generated list of candidates for a mention prior to the disambiguation step.

%\reminder{DNG: These aren't really baselines; they're the tools we're using to do NER and initial candidate set generation.  The baseline is just not using MedType.} 
%We integrate \method{} into five widely-used entity linking models, each of which performs NER and produces a ranked list of candidate CUIs for each mention: \textbf{\metamap{}} \cite{metamap}, \textbf{\ctakes{}} \cite{ctakes}, \textbf{\metamaplite{}} \cite{metamaplite}, \textbf{\quickumls{}}  \cite{quickumls}, and  \textbf{\scispacy{}} \cite{scispacy}, and \textbf{MedLinker} \cite{Medlinker}. A brief description of each method is included in Appendix \ref*{sec:linker_desc}. 

\subsection{Evaluation Metrics}
\label{sec:evaluation}

For semantic type prediction, which we model as a multi-label classification problem, following \cite{herer,reside}, we use the area under the Precision-Recall curve (AUC) as our evaluation metric. 

For entity linking, we evaluate the performance using F1-score for two metrics. In (1) \textit{Exact\_mention\_id\_match} (\textbf{Exact}),  \showedits{true positives are only those samples where both the predicted mention bounds and entity concept identifier exactly match the annotation}. This is directly adopted from TAC KBP 2013\footnote{https://tac.nist.gov/2013/KBP/}. In (2) \textit{Partial\_mention\_id\_match} (\textbf{Partial}), \showedits{a weighted score is assigned to predicted mentions based on the amount of overlap with annotated mention bounds and entity id match. Following \cite{partial_metric}, for mention matching, the number of overlapped characters between system generated mention and a ground-truth mention is considered}. All the scores are computed using an open-source entity linking evaluation toolkit\footnote{https://github.com/wikilinks/neleval}.

\subsection{Implementation Details}

\textbf{Online Demo \& medtype-as-service:} Along with providing a step-by-step guide for reproducing all the results reported in the paper, we also provide code for running an online demo of \method. We also provide a scalable implementation of \method{} called medtype-as-service which is based on \textit{bert-as-service} \cite{bert_as_service} for processing thousands of documents simultaneously. 

\textbf{Hyperparameters:} We use pre-trained weights of BioBERT \cite{biobert} for initializing BERT component of \method{}. \method{} is implemented using HuggingFace Transformers library \cite{huggingface}. For training, we utilize Adam optimizer \cite{adam} with a learning rate in range $(10^{-3}, 10^{-5})$. The window size of context ($k$) is chosen from $\{48,64,128\}$. The best hyperparameters were selected based on the performance on the validation split of the datasets. We use the default hyperparameters for all the entity linkers and components of \method{}. A grid search over the validation split was performed for deciding a threshold for each semantic type from the range of (0.001, 1). The area under the Precision-Recall curve (AUC) was used for choosing the best threshold. 

\textbf{Training Details:} All training was performed on NVIDIA-GTX 1080Ti GPUs. Each training epoch of \method{} takes from 5 mins to 2 days depending on the size of the dataset. The models are trained for multiple epochs until the validation performance starts to degrade. In terms of number of parameters, \method{} has around 110 million parameters (same as BERT-base model).

\section{Results}
\label{sec:results}
%\reminder{DNG: More effectively phrased as the major finding from each (suggestions: "Incorporating \method{} in existing EL models improves performance", "", and "\wiki{} and \pubmed{} contribute to specific semantic types", or something like that}

%In this section, we provide details for the following major findings:
Medical information extraction is a complex process, with multiple points of evaluation and multiple types of impact from any new contribution.  We present results for four specific questions that examine the impact of semantic type filtering with \method{}:
%In this section, we answer the following questions:   
 \begin{itemize}[itemsep=2pt,topsep=2pt,parsep=1pt,partopsep=0pt,leftmargin=20pt]
 	\item[Q1.] How effective is \method{} for semantic type prediction, and what is the impact of our novel datasets? (Section \ref{sec:results_type_pred})
	\item[Q2.] Does incorporating \method{} in existing entity linking systems help the overall pipeline? (Section \ref{sec:results_main})    
	%\item[Q3.] Where do performance gains from using \wiki{} and \pubmed{} come from? (Section \ref{sec:type_analysis})
	\item[Q3.] What specific successes do we see from combining \method{}, \wiki{}, and \pubmed{}, and what are remaining challenges? (Section~\ref{sec:type_analysis})
	\item[Q4.] How much does semantic type-based filtering help prune irrelevant candidates? (Section \ref{sec:results_cand_prun})
\end{itemize}

%\subsection{Evaluation of Semantic Type Prediction}
\subsection{\method{} is State-of-the-Art for Medical Semantic Type Prediction}
\label{sec:results_type_pred}
%For evaluating the effectiveness of our proposed model, \method{}, for semantic-type prediction, we compare it against the baselines stated in Section \ref{sec:baselines}. Also, for analyzing the impact of using our proposed, we report the performance of \method{} pre-trained on \pubmed{}.  
The first step in our evaluation is a modular investigation of the semantic type prediction task on its own. In this section, we compare \method{} against the baseline methods detailed in Section \ref{sec:baselines} for semantic type prediction. We also evaluate the effectiveness of utilizing \wiki{} and \pubmed{} datasets for the task. For quantifying the benefit of our proposed method and datasets, we report the performance of \method{} trained under different settings, as defined below.

\begin{itemize}[itemsep=1pt,parsep=0pt,partopsep=0pt,leftmargin=10pt,topsep=1pt]
	\item \textbf{\method{} (MT)} denotes \method{} trained on the training split of the corresponding datasets. 
	\item \textbf{MT $\leftarrow$ \wiki{}} refers to the model first trained on \wiki{} and then fine-tuned using the training data. 
	\item \textbf{MT $\leftarrow$ \pubmed{}} similar to T $\leftarrow$ \wiki{}, indicates \method{} first trained on \pubmed{} and then fine-tuned on the training data. 
	\item \textbf{MT $\leftarrow$ Both} denotes the combined model which utilizes both the proposed datasets. It concatenates BERT encoding from T $\leftarrow$ \wiki{} and T $\leftarrow$ \pubmed{} models and passes it to a classifier which is trained using the training dataset.
\end{itemize}

%As we model type prediction as a multi-label classification problem, we use the area under the Precision-Recall curve (AUC) as our evaluation metric. 
Semantic type prediction results are presented in Table \ref{tab:type_pred}. We find that \method{} outperforms all the baselines on three of the four evaluation datasets when trained only on the training split. Compared to the best performing baseline, we obtain a gain of 0.2, 0.7, and 9.1 AUC on \cdr{}, \sharecorpus{}, and \medmen{} respectively. \medmen{} contains a much greater diversity of semantic types than other datasets (as shown in Table \ref{tab:cat_distribution}). Thus, obtaining a large improvement on it indicates that \method{} is more suited for handling large set of types compared to the baseline methods.

Further, we find that utilizing our novel datasets \wiki{} and \pubmed{} yields considerable gain in performance. On average, we obtain an increase in AUC of 1.7 from \wiki{} alone, 3.9 from \pubmed{} alone, and 4.5 from using both, across all datasets. The combined model which allows to incorporate the benefits from both the corpora gives the best performance. This shows that both the datasets contain complementary high-value information for semantic type prediction.

%The best results on all 4 datasets are achieved by pre-training \method{} on our novel \pubmed{} dataset. Pre-training provides an average gain of 3.6 AUC across all datasets compared to the vanilla \method{} model, and achieves between 1.6-7\% gain in AUC over the second-best method. These results clearly demonstrate that \pubmed{} contains high-value information for semantic type prediction across data types. 

\begin{table}[!t]
	\centering
	\small
	%	\resizebox{\linewidth}{!}{
	\begin{tabular}{lx{1.6cm}x{1.6cm}x{1.6cm}c}%cccc}
		\toprule
		%\multicolumn{1}{c}{}            & \multicolumn{1}{c}{\textbf{\ncbi{}}}         & \multicolumn{1}{c}{\textbf{\cdr{}}}         & \multicolumn{1}{c}{\textbf{\sharecorpus{}}}             & \multicolumn{1}{c}{\textbf{MedMent}} \\
		&{\bf\ncbi{}}&{\bf\cdr{}}&{\bf\sharecorpus{}}&{\bf\medmen{}}\\
		\midrule
		\baseattn{} \cite{attention_ner}   &  94.5 & 89.1 &  88.7 & 72.0\\
		\basefc{} \cite{deeptype}    &  95.1 & 82.9 & 89.3 & 72.9 \\
		\basernn{} \cite{deeptype} & 92.8 & 86.9 & 86.1 & 74.1 \\
		\basecnn{}  \cite{murty-etal-2018-hierarchical} & 95.2 & 88.9 & 89.8 & 74.4 \\
		MedNER \cite{Medlinker} & 95.6 & 90.2 &84.4 &67.5 \\
		\midrule
		\method{} (MT) & 94.5 & 90.4 & 90.5 & 83.5 \\
		MT $\leftarrow$ \wiki{} & 94.9 & 93.5 & 93.2& 84.0 \\
		MT $\leftarrow$ \pubmed{} & 96.8 & \textbf{97.3} & 93.6 & 86.8 \\
		MT $\leftarrow$ Both & \textbf{97.2}  & \textbf{97.3} & \textbf{95.1} & \textbf{87.3}\\
		%			T  $\leftarrow$ Both           &  \textbf{96.5}  &  95.6  & \textbf{93.1} & \textbf{81.7}\\
		\bottomrule
	\end{tabular}
	%	}
	\caption{\label{tab:type_pred} Semantic type prediction results, comparing \method{} (with and without additional corpora) to our four baselines; we report the area under the precision-recall curve as our evaluation metric. MT $\leftarrow$ X denotes \method{} first trained on X dataset then fine-tuned using T. We find that \method{} outperforms other methods on 3 out of 4 datasets. Also, pre-training on \wiki{} and \pubmed{} gives substantial boost in the performance. More details are provided in Section \ref{sec:results_type_pred}.}
	
\end{table}

%The maximum gain is obtained in MedMentions dataset which comprises of several mentions from various semantic type 
%For quantifying the benefit of utilizing our proposed datasets, we report the performance of \method{} trained under different settings, as defined below.
%\begin{itemize}[itemsep=1pt,parsep=0pt,partopsep=0pt,leftmargin=10pt,topsep=1pt]
%	\item \textbf{Training data (T)} denotes \method{} trained on the training split of the corresponding datasets. 
%	\item \textbf{T $\leftarrow$ \pubmed{}} similar to T $\leftarrow$ \wiki{}, indicates \method{} first trained on \pubmed{} and then fine-tuned on the training data. 
%\end{itemize}

%The overall results are presented in Table \ref{tab:data_effect}. We find that there is a substantial gain in performance on using \wiki{} and \pubmed{} along with the training data. Further,  we find that the combined model which allows to incorporate the benefits from both the corpora gives the best performance on 3 out of 4 datasets. This shows that both the datasets contain complementary information relevant for semantic type prediction. Overall, we get an average absolute increase of 9.8, 12.7, and 13.4 AUC from \wiki{}, \pubmed{}, and the combined corpora respectively. \\

\begin{table}[!t]
	\centering
	\small
	\resizebox{\textwidth}{!}{
		\begin{tabular}{lcccccccc}
			\toprule
			\multicolumn{1}{c}{}  & \multicolumn{2}{c}{\textbf{\ncbi{}}}         & \multicolumn{2}{c}{\textbf{\cdr{}}}         & \multicolumn{2}{c}{\textbf{\sharecorpus{}}}             & \multicolumn{2}{c}{\textbf{\medmen{}}} \\
			\cmidrule(r){2-3}  \cmidrule(r){4-5} \cmidrule(r){6-7} \cmidrule(r){8-9}
			& Exact & Partial & Exact & Partial & Exact & Partial & Exact & Partial \\
			\midrule
			\textbf{MetaMap} & 39.6 & 45.0  &  54.2 & 56.3  &  33.8 & 34.6  &  36.7 & 39.8 \\
			\midrule
			Oracle (Fine) & +0.8 & +1.0  &  +0.3 & +0.4  &  +0.5 & +0.6  &  +6.4 & +6.9 \\
			Oracle (Coarse) & +0.8 & +1.0  &  +0.2 & +0.3  &  +0.5 & +0.6  &  +5.7 & +6.1 \\
			\basecnn{} & +0.7 & +0.8  &  +0.2 & +0.3  &  +0.2 & +0.3  &  +3.6 & +3.8 \\
			\midrule
			\method{}  & \textbf{+0.8} & \textbf{+1.0}  &  \textbf{+0.2 }& \textbf{+0.3}  &  +0.3 & +0.4  &  +4.0 & +4.3 \\
			\midrule \midrule
			\textbf{cTakes} & 39.2 & 45.9  &  54.5 & 57.0  &  32.3 & 33.3  &  16.9 & 18.3 \\
			\midrule
			Oracle (Fine) & +0.3 & +0.3  &  +0.1 & +0.2  &  +0.1 & +0.2  &  +0.2 & +0.2 \\
			Oracle (Coarse) & +0.3 & +0.3  &  +0.1 & +0.2  &  +0.1 & +0.2  &  +0.2 & +0.2 \\
			\basecnn{} & +0.3 & +0.3  &  +0.1 & +0.2  &  +0.0 & +0.1  &  +0.1 & +0.1 \\
			\midrule
			\method{}  & \textbf{+0.3} & \textbf{+0.3}  &  \textbf{+0.1} & \textbf{+0.1}  &  +0.1 & +0.1  &  \textbf{+0.2 }& \textbf{+0.2} \\
			\midrule \midrule
			\textbf{MetaMapLite} & 35.4 & 39.4  &  50.3 & 51.5  &  27.1 & 27.5  &  32.6 & 35.2 \\
			\midrule
			Oracle (Fine) & +5.9 & +5.9  &  +2.7 & +2.8  &  +4.7 & +4.8  &  +7.2 & +7.8 \\
			Oracle (Coarse) & +5.9 & +5.9  &  +2.6 & +2.7  &  +4.7 & +4.7  &  +6.0 & +6.5 \\
			\basecnn{} & +5.7 & +5.7  &  +2.3 & +2.4  &  +4.1 & +4.1  &  +3.9 & +4.0 \\
			\midrule
			\method{}  & \textbf{+5.9} & \textbf{+5.9}  &  +2.5 & +2.6  &  +4.3 & +4.4  &  +4.4 & +4.6 \\
			\midrule \midrule
			\textbf{QuickUMLS} & 27.0 & 31.7  &  36.5 & 39.1  &  17.3 & 19.2  &  28.7 & 31.4 \\
			\midrule
			Oracle (Fine) & +0.2 & +0.6  &  +5.0 & +5.2  &  +5.2 & +5.5  &  +9.8 & +10.7 \\
			Oracle (Coarse) & +0.2 & +0.6  &  +4.5 & +4.6  &  +5.1 & +5.4  &  +7.7 & +8.5 \\
			\basecnn{} & +0.0 & +0.2  &  +4.0 & +4.1  &  +4.0 & +4.2  &  +4.9 & +5.2 \\
			\midrule
			\method{}  & +0.1 & +0.5  &  +4.3 & +4.4  &  +4.8 & +5.0  &  +5.9 & +6.4 \\
			\midrule \midrule
			\textbf{ScispaCy} & 43.1 & 47.5  &  49.4 & 53.7  &  25.4 & 29.0  &  37.2 & 40.6 \\
			\midrule
			Oracle (Fine) & +2.2 & +4.1  &  +1.7 & +2.6  &  +3.5 & +5.1  &  +8.2 & +9.4 \\
			Oracle (Coarse) & +2.2 & +4.1  &  +1.7 & +2.5  &  +3.4 & +5.0  &  +6.8 & +7.8 \\
			\basecnn{} & +1.7 & +3.6  &  +0.5 & +1.2  &  +2.9 & +4.0  &  +3.5 & +3.9 \\
			\midrule
			\method{}  & +1.9 & +3.8  &  +1.3 & +2.2  &  +3.1 & +4.5  &  +4.1 & +4.6\\
			\bottomrule
		\end{tabular}
	}
	\caption{\label{tab:main_results} 
		%		For quantifying the impact of incorporating a type-based candidate concepts filtering step, 
			For quantifying the impact of semantic type prediction on medical entity linking, 
		we report the F1-score for five medical entity linking methods on multiple datasets.	
		%The change in performance of existing medical entity linkers with different mention disambiguation methods on multiple datasets. We report the F1-score for strict mention and entity identifier match. 
		For each method, the first row is its base performance, and the following rows indicate the change in F1-score on incorporating a type-based candidate concepts filtering step. \textbf{Bold} indicates the case when \method{} performance matches with an oracle. We report the results with the oracle type predictors (fine-grained and coarse-grained) and \method{}. Overall, we find that \method{} gives performance comparable to an oracle and improves medical entity linking across all settings. Please refer to Section \ref{sec:results_main} for details.} 
\end{table}

\subsection{\method{} Consistently Improves Overall Information Extraction Performance}
\label{sec:results_main}
The primary goal of our study is to investigate the impact of adding a semantic type prediction module to the medical information extraction pipeline.  In this section, we evaluate the impact of \method{} on biomedical information extraction when integrated with the tools detailed in Section \ref{sec:linkers}. Table~\ref{tab:main_results} reports the results for the \textit{Exact\_mention\_id\_match} and \textit{Partial\_mention\_id\_match} metrics, as described in Section \ref{sec:evaluation}. 
%For \wiki{} dataset, we evaluate only on identified mentions which overlap with the ground truth. We make this exception since Wikipedia editors are encouraged to link only the early occurrences of an entity, as opposed to exhaustive annotations. Moreover, in \wiki{}, we only include the mentions for which mapping to UMLS is available, which is not comprehensive.

As discussed in Section \ref{sec:datasets}, the \ncbi{}, \cdr{}, and \sharecorpus{} datasets were annotated for specific categories of medical concept mentions (e.g., diseases and disorders only); concept mentions outside of these categories were excluded from annotation. By contrast, the information extraction tools we experimented with were all preconfigured for broad-coverage extraction of all types of medical information.  Thus, the set of predicted medical concept mentions output by any one of our toolkits could include concepts of a type excluded from dataset annotation---predictions which we are therefore unable to evaluate.  To avoid including these mentions in our evaluation, we filtered the output of each toolkit for a given dataset to the semantic types included in that dataset's annotation (e.g., disease mentions only for the NCBI Disease Corpus). We determined the semantic type of predicted concept mentions using the final CUI produced as the top-ranked candidate after processing with the full information extraction pipeline (including semantic type prediction, when used).  Thus, if the top-ranked candidate for a given mention was of an excluded type when using an unmodified entity linker, that mention would be excluded from evaluation (informing both mention detection and entity linking evaluation); however, if the introduction of semantic type filtering removed that top-ranked candidate in favor of a lower-ranked candidate of a type {\it included} in dataset annotation, the mention would be included in evaluation.  %Implications for evaluation of medical information extraction arising from this mismatch between annotation goals and application goals are further discussed in Section~\ref{sec:evaluation-mismatch}.

%To evaluate the maximum possible improvement from type-based pruning of candidate concepts, we report entity linking results using an \textit{Oracle} model which always filters the candidate set to entities of the same type as the gold standard CUI.
We compare \method{} against the two oracle approaches described in Section~\ref{sec:experiments}, as well as against the best-performing baseline from Section~\ref{sec:results_type_pred}.
%As described in Section \ref{sec:method_details}, we grouped the 127 types provided by the UMLS into 24 coarse-grained semantic type groups for training \method{}; we therefore present the results using both a fine-grained oracle (\textit{Oracle (Fine)}, which uses all 127 original types) and a coarse-grained (\textit{Oracle (Coarse)}, using the 24 grouped types).
For each information extraction system, we report its default performance along with the change in scores when adding different type-based candidate filtering methods. The results for \method{} are obtained after pre-training on \wiki{} and \pubmed{} datasets, based on our findings in Section~\ref{sec:results_type_pred}.

Across most information extraction tools and datasets, \method{} yields a substantial improvement in performance, and it consistently matches or outperforms \basecnn{}, the best prior method for type prediction. Notably, in no situation does \method{} degrade performance; thus, the results indicate that including a type-based filtering step enhances information extraction systems in most cases. (See Section~\ref{sec:toolkit-comparison} for a discussion of the differences between performance of individual information extraction tools.) The gain with \method{} is comparable to improvement with using an oracle, indicating that \method{} is reliable enough to use off-the-shelf. The results also show that there is not much difference in performance of Oracle (Fine) and Oracle (Coarse). This justifies our choice of working with 24 semantic groups rather than the 127 semantic types defined in the UMLS Metathesaurus.
 
 We used paired bootstrap significance testing \cite{bootstrap_stat_sig} for validating statistical significance $(p < 0.01)$ of improvements from \method{} compared to the default pipeline and the top performing baseline performance. Our results clearly support the central thesis of this work, that pruning irrelevant candidate concepts based on semantic type helps improve medical entity linking.

\subsection{\method{} Improves Entity Linking Performance}
\label{sec:results_el_only}

The evaluations described in Section~\ref{sec:results_main} account for both mention detection---which semantic type filtering can affect by removing all candidates for a mention, leading to its exclusion---and entity linking.  We therefore isolated the effect of \method{} on the entity linking portion of the information extraction pipeline alone by restricting our analysis to only predicted concept mentions overlapping with gold annotated mentions, and calculating the {\it Partial\_mention\_id\_match} F-1 metric (detailed in Section~\ref{sec:evaluation}) on this subset.  Table~\ref{tab:el_only_results} reports results for \scispacy{} (the best-performing information extraction tool) on all four evaluation datasets.

Baseline performance with \scispacy{} is 7-10 points higher in this more restricted evaluation, as compared to Table~\ref{tab:main_results}, reflecting the additional challenges of mention detection which go into the overall evaluation.  Semantic type filtering leads to similar improvements for \ncbi{} and \cdr{} in this setting, but noticeably larger improvements on \sharecorpus{} and \medmen{}, demonstrating that overall information extraction improvements from semantic type filtering are coming primarily from the entity linking portion of the pipeline.

\begin{table}[!t]
	\centering
	\small
	\resizebox{\textwidth}{!}{
		\begin{tabular}{lx{1.7cm}x{1.7cm}x{1.7cm}c}%cccc}
			\toprule
			%\multicolumn{1}{c}{}  & \multicolumn{1}{c}{\textbf{\ncbi{}} (1042)}         & \multicolumn{1}{c}{\textbf{\cdr{}} (9243)}         & \multicolumn{1}{c}{\textbf{\sharecorpus{}} (6691)}             & \multicolumn{1}{c}{\textbf{MedMent} (61367)} \\
			&\textbf{\ncbi{}}& \textbf{\cdr{}}& \textbf{\sharecorpus{}}& \textbf{\medmen{}}\\
			&(1,042)&(9,243)&(6,691)&(61,367)\\
% 			\cmidrule(r){2-3}  \cmidrule(r){4-5} \cmidrule(r){6-7} \cmidrule(r){8-9}
% 			& Exact & Partial & Exact & Partial & Exact & Partial & Exact & Partial \\
			\midrule
			\textbf{ScispaCy} &56.0  &  60.9   &  30.9  &   42.8\\
			\midrule
			Oracle (Fine) &  +4.2  &   +2.7   &  +5.3  &    +9.9\\
			Oracle (Coarse) &  +4.2  &   +2.6   &  +5.3  &    +8.1\\
			\basecnn{} &  +3.5  &   +1.2   &  +4.2  &    +4.1\\
			\midrule
			\method{}  &  +3.8  &   +2.2   &  +4.7  &    +4.9\\
			\bottomrule
		\end{tabular}
	}
	\caption{\label{tab:el_only_results} 
		%		For quantifying the impact of incorporating a type-based candidate concepts filtering step, 
		Results of {\it Partial\_mention\_id\_match} evaluation of \scispacy{} on all four evaluation datasets. Evaluation is restricted to only predicted samples that overlap with gold annotations, to control for the effects of mention detection errors. The number of samples in this restricted subset of each dataset is given in the column headers.
		}\vspace{-4mm}
\end{table}

\subsection{Gains and Challenges of \method{}, \wiki{}, and \pubmed{}}
%Type-Level Analysis of Improvement from \pubmed{}} 
\label{sec:type_analysis}
{\bf \pubmed{} and \wiki{} yield large improvements for rare types:}
As observed in Section \ref{sec:results_type_pred}, pretraining \method{} on \wiki{} and \pubmed{} led to substantial increases in semantic type prediction performance. In this section, we investigate which types of medical concept mentions were improved the most from this pretraining step.
%what led to the improvement on semantic type prediction from . 
%Here, we report F1-score  on prominent semantic types for $\text{T}$, $\text{T} \leftarrow \text{\wiki{}}$, and $\text{T} \leftarrow \text{\pubmed{}}$ models. 
For this, we report the F1 score of \method{}, MT $\leftarrow$ \wiki{}, MT $\leftarrow$ \pubmed{} and MT $\leftarrow$ Both models (as defined in Section \ref{sec:results_type_pred}) across all semantic types on all the datasets. The overall results are summarized in Table \ref{tab:typewise_perf}. In general, we find that performance improves across all semantic types as we utilize additional corpora, but the maximum gain is obtained on types which have less coverage in the training split. For instance, on types such as \textit{Pathological Function} and \textit{Sign or Symptom} in the NCBI Disease Corpus, the F1 score jumps from 0 to 80 and 83.3 respectively.
Thus, the broad coverage of medical concept types in \wiki{} and \pubmed{}, combined with their large scale, helps to fill in the gaps of semantic types that are not well-represented in the evaluation datasets directly.
%Thus, the results give insight behind the overall gain which we obtain on utilizing \wiki{} and \pubmed{} for semantic type prediction. 

%\begin{figure}[h]
%	\centering
%	\includegraphics[width=1.\linewidth]{./images/confusion_matrix-crop}
%	\caption{\label{fig:confusion_matrix} Confusion matrix on validation split of MedMentions dataset of \method{} pre-trained on \wiki{} and \pubmed{} datasets. There matrix shows that the model correctly identifies the correct semantic type in most cases. However, it does get confused for similar classes such \textit{Disorders}, \textit{Diseases} and \textit{Mental Dysfunction}. Please refer to Section \ref{sec:type_analysis} for more details. }
%\end{figure}

\begin{landscape}
\begin{table}
	\centering
	\footnotesize
	\begin{tabular}{lcccccccccccccccc}
	\toprule
	\multicolumn{1}{c}{}  & \multicolumn{4}{c}{\textbf{\ncbi{}}}         & \multicolumn{4}{c}{\textbf{\cdr{}}}         & \multicolumn{4}{c}{\textbf{\sharecorpus{}}}             & \multicolumn{4}{c}{\textbf{MedMentions}} \\
	\cmidrule(r){2-5}  \cmidrule(r){6-9} \cmidrule(r){10-13} \cmidrule(r){14-17}
	\multicolumn{1}{c}{}  & MT & $\leftarrow$ W & $\leftarrow$ P & $\leftarrow$ B         &  MT & $\leftarrow$ W & $\leftarrow$ P & $\leftarrow$ B & MT & $\leftarrow$ W & $\leftarrow$ P & $\leftarrow$ B & MT & $\leftarrow$ W & $\leftarrow$ P & $\leftarrow$ B \\
	\midrule
Activities \& Beh.    & -    &   -     &  -     &  -      &   -     &  -     &  -     &  -      &    -     &  -     &  -     &  -      &   71.9  &  71.7  &  74.4  & \textbf{74.9} \\
Anatomy               & -    &   -     &  -     &  -      &   -     &  -     &  -     &  -      &    -     &  -     &  -     &  -      &   81.3  &  82.7  &  \textbf{86.5}  & 86 \\
Chemicals \& Drugs    & -    &   -     &  -     &  -      &   83    &  83    &  91.5  &  \textbf{91.8}   &    -     &  -     &  -     &  -      &   77.8  &  78.1  &  \textbf{82.2}  & \textbf{82.2} \\
Concepts \& Ideas     & -    &   -     &  -     &  -      &   -     &  -     &  -     &  -      &    -     &  -     &  -     &  -      &   80.5  &  81.2  &  \textbf{82.8}  & \textbf{82.8} \\
Devices               & -    &   -     &  -     &  -      &   -     &  -     &  -     &  -      &    -     &  -     &  -     &  -      &   52.2  &  46.4  &  \textbf{55.5}  & 54.1 \\
Disease or Syn.       & 94.5 &   95.5  &  97.2  &  \textbf{97.6}   &   87.8  &  90.5  &  93.2  &  \textbf{93.7}   &    84.6  &  91.3  &  92.3  &  \textbf{92.8}   &   79    &  81    &  84.4  & \textbf{84.9} \\
Disorders             & 58.9 &   68.7  &  69    &  \textbf{69.2}   &   82.4  &  79.4  &  \textbf{85.8}  &  85.7   &    50.7  &  78    &  79.9  &  \textbf{82.1}   &   62.1  &  64.4  &  67.9  & \textbf{68.6} \\
Finding               & 0    &   45    &  46.8  &  \textbf{51.2}   &   59.6  &  77.1  &  86.1  &  \textbf{87.6}   &    47.5  &  79.5  &  82.5  &  \textbf{83.3}   &   54.8  &  57.5  &  58.5  & \textbf{59.8} \\
Functional Concept    & -    &   -     &  -     &  -      &   -     &  -     &  -     &  -      &    -     &  -     &  -     &  -      &   76.7  &  76.4  &  77.2  & \textbf{77.4} \\
Genes \& Mol. Seq.    & -    &   -     &  -     &  -      &   -     &  -     &  -     &  -      &    -     &  -     &  -     &  -      &   67.8  &  67    &  \textbf{72}    & \textbf{72} \\
Living Beings         & -    &   -     &  -     &  -      &   0     &  0     &  \textbf{57.1}  &  40     &    -     &  -     &  -     &  -      &   88.1  &  88.6  &  90.1  & \textbf{90.1} \\
Mental/Beh. Dys.      & 17.4 &   81.1  &  \textbf{83.3}  &  \textbf{83.3}   &   58.8  &  90.1  &  92.6  &  \textbf{92.9}   &    48.4  &  83.2  &  78.8  &  \textbf{85.4}   &   76.7  &  79    &  80.7  & \textbf{82.2} \\
Neoplastic Process    & 91.7 &   93.1  &  \textbf{94.2}  &  92.7   &   90.9  &  90.8  &  \textbf{94.6}  &  92.2   &    71.5  &  89.2  &  90.9  &  \textbf{91.4}   &   85.6  &  86    &  87.4  & \textbf{88.1} \\
Objects               & -    &   -     &  -     &  -      &   0     &  20.8  &  \textbf{46.4}  &  29.2   &    -     &  -     &  -     &  -      &   72.3  &  71.6  &  75.7  & \textbf{76.1} \\
Occupations           & -    &   -     &  -     &  -      &   -     &  -     &  -     &  -      &    -     &  -     &  -     &  -      &   46.7  &  47.1  &  \textbf{58.4}  & 55.5 \\
Organic Chemical      & -    &   -     &  -     &  -      &   91.9  &  91.3  &  \textbf{94.3}  &  94.1   &    -     &  -     &  -     &  -      &   71.9  &  73.6  &  \textbf{80.6}  & 80.2 \\
Organizations         & -    &   -     &  -     &  -      &   -     &  -     &  -     &  -      &    -     &  -     &  -     &  -      &   73    &  74    &  75.6  & \textbf{77.3} \\
Pathologic Function   & 0    &   76.2  &  \textbf{82.4}  &  80     &   59.6  &  86.2  &  90.2  &  \textbf{91}     &    74.6  &  85.1  &  85.9  &  \textbf{86.5}   &   65.6  &  69.9  &  70.1  & \textbf{72.7} \\
Pharm. Substance      & -    &   -     &  -     &  -      &   92    &  91.8  &  \textbf{93.3}  &  93.1   &    -     &  -     &  -     &  -      &   63.6  &  64.3  &  \textbf{70.8}  & 70.3 \\
Phenomena             & -    &   -     &  -     &  -      &   33.3  &  74.3  &  \textbf{93.8}  &  92.3   &    -     &  -     &  -     &  -      &   51.1  &  54.3  &  \textbf{61.5} & 60.7 \\
Physiology            & -    &   -     &  -     &  -      &   0     &  60.8  &  \textbf{63.7}  &  60.8   &    -     &  -     &  -     &  -      &   72.7  &  74.6  &  77.3  & \textbf{77.8} \\
Procedures            & -    &   -     &  -     &  -      &   0     &  0     &  44.4  &  \textbf{53.3}   &    -     &  -     &  -     &  -      &   77.1  &  78.3  &  \textbf{80.3}  & 80.2 \\
Qualitative Concept   & -    &   -     &  -     &  -      &   -     &  -     &  -     &  -      &    -     &  -     &  -     &  -      &   82.8  &  83.5  &  84.1  & \textbf{84.4} \\
Sign or Symptom       & 0    &   81.8  &  \textbf{83.3}  &  \textbf{83.3}   &   46.4  &  89.5  &  89.9  &  \textbf{91.7}   &    80.6  &  92.8  &  \textbf{94.7}  &  94.4   &   72.1  &  75.4  &  75.1  & \textbf{78.9} \\
\bottomrule

\end{tabular}
	\caption{\label{tab:typewise_perf}
		Type-wise analysis of the impact on using \method{} with \pubmed{} on \ncbi{}, \cdr{}, \sharecorpus{}, and \medmen{} datasets. We report F1-score for each semantic type. MT denotes \method{}, $\leftarrow$ W and $\leftarrow$ P indicate \method{} first pre-trained on \wiki{} and \pubmed{} dataset, and $\leftarrow$ B denotes \method{} pre-trained on both the datasets. '-' mean that the semantic type was not part of the dataset.}
\end{table}

\end{landscape}

\textbf{Error analysis of \method{}:} To gain insight into further opportunities for improvement in semantic type prediction, we analyzed \method{} errors in the validation split of the MedMentions dataset when using our best performing model, which is pre-trained on both \wiki{} and \pubmed{} datasets.  As reflected by Table~\ref{tab:type_pred}, \method{} is able to identify the correct semantic type in the majority of cases.  However, as Table~\ref{tab:typewise_perf} shows, performance is not uniform across semantic types; e.g., {\it Devices}, {\it Finding}, {\it Occupations}, and {\it Phenomena} (all involving fairly common words) remain particularly challenging in these data. 
%The results are summarized in Figure \ref{fig:confusion_matrix}. 
Table~\ref{tab:medtype-confusion} shows the semantic types most commonly confused with one another, in many cases, we see mispredictions of more abstract types such as {\it Objects}, {\it Concepts \& Ideas}, and {\it Functional Concepts}, regardless of gold semantic type.
%However, as shown in Table~\ref{tab:medtype-confusion}, semantically similar types such as \textit{Diseases} and \textit{Disorders} or \textit{Organical Chemicals} and \textit{Chemical \& drugs} \method{} fails to recognize the correct semantic type.
Thus, there is still significant scope for improvement on this problem.

\begin{table}[h]
	\centering
	\small
	\renewcommand{\arraystretch}{1.2}
	\begin{tabular}{p{0.35\linewidth}p{0.58 \linewidth}}
		\toprule
		\textbf{Target Semantic Type}      & \multicolumn{1}{c}{\textbf{Top Confused Semantic Types}}\\
		\midrule
		Devices & Concepts \& Ideas, Objects, Procedures, \\
		Disorders & Disease or Syndrome, Finding		\\
		Finding  & Concept \& Ideas, Physiology, Functional Concept\\
		Functional Concept & Procedures, Concepts \& Ideas \\
		Genes \& Mol. Sequences & Chemicals \& Drugs \\
		Mental and Behavioral Dys. & Disease or Syndrome, Finding\\
		Objects & Concepts \& Ideas, Chemicals \& Drugs \\ 
		Occupations & Procedures, Concepts \& Ideas, Functional Concepts \\ 
		Organic Chemicals & Chemicals \& Drugs, Pharmacological Substances\\
		Organizations & Concepts \& Ideas, Procedures, Living Beings \\
		Pathologic Functions & Disease or Syndrome, Finding, Functional Concepts\\  
		Pharmacological Substance & Chemical \& Drugs, Organic Chemicals\\
		
		%		EHR documents & \\ 
		
		\bottomrule
	\end{tabular}
	\caption{\label{tab:medtype-confusion} Most frequent confusions in semantic type predictions on the MedMentions validation set, using \method{} pretrained on \wiki{} and \pubmed{}.
	}
\end{table}

%\begin{figure}[t]
%	\centering
%	\includegraphics[width=\linewidth]{./images/analysis_plot-crop}
%	\caption{\label{fig:data_analysis} Analysis of the impact on using \pubmed{} dataset on two semantic types: \textit{Mental Dysfunction} and \textit{Sign or Symptoms} on \cdr{} and \sharecorpus{} datasets. We find that training \method{} on our proposed datasets gives considerable improvement on semantic type prediction. Refer to Section \ref{sec:type_analysis} for details.} 
%\end{figure}
\begin{figure}[h]
	\centering
	\includegraphics[width=\linewidth]{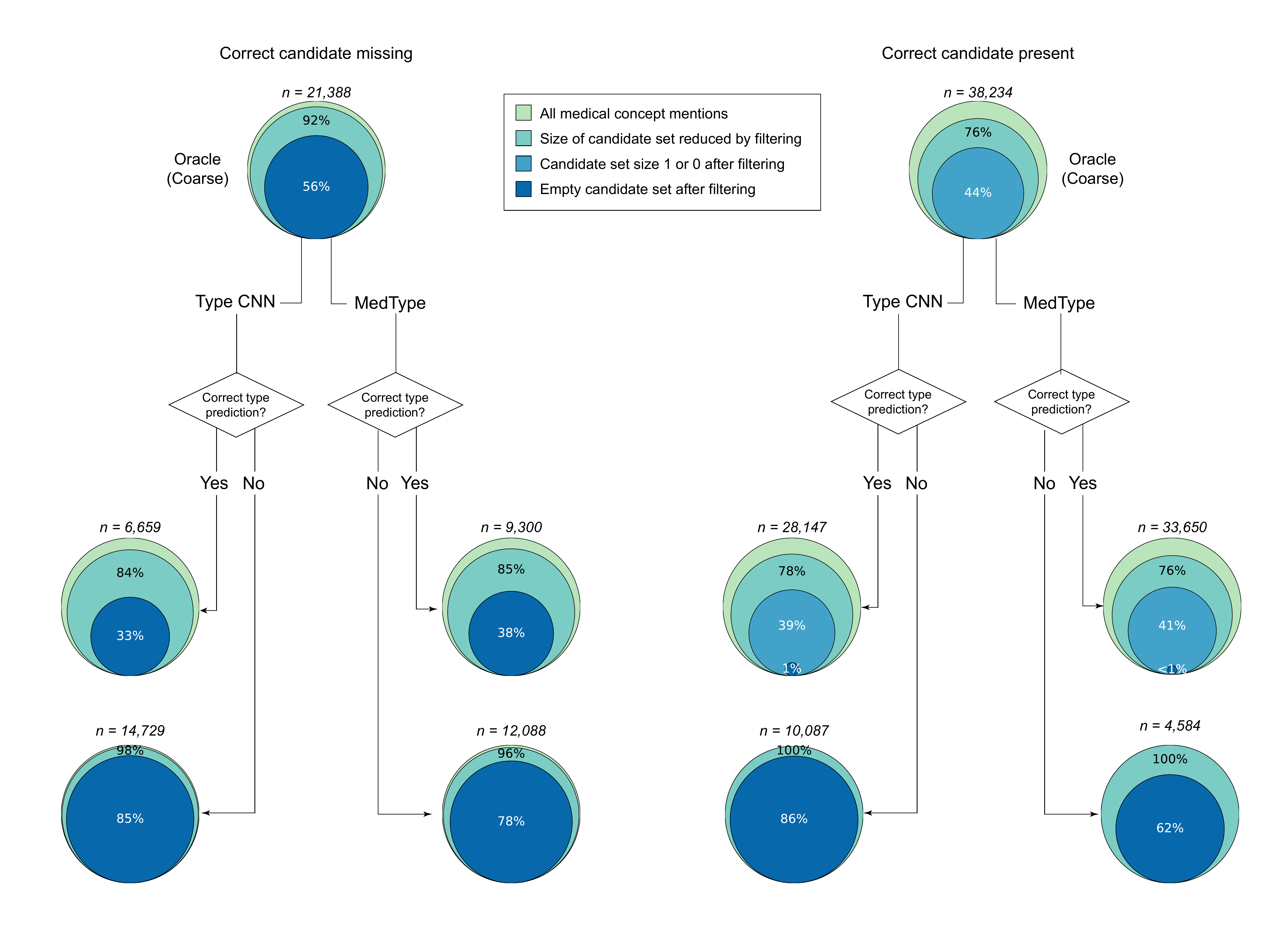}
	\caption{Outcomes of semantic type filtering in MedMentions data, in terms of reduction in candidate set size. All results are reported using the best-performing information extraction model (ScispaCy). Top graphs display candidate set reduction using oracle type filtering, broken down into whether the correct candidate was included in the list generated by ScispaCy. Bottom graphs illustrate corresponding outcomes from \method{} and the strongest type prediction baseline (Type CNN), broken down by whether the predicted type was correct. The number of samples each graph displays is provided, along with the percentage of these samples included in each reduction category.
	    %Outcomes of using \method{} to prune candidate concept sets in
		%MedMentions, using \scispacy{}. Mentions for which \scispacy{} did not generate one or more correct candidate are omitted. The results are thus for 38,234 annotated mentions in the dataset. For more details please refer to Section \ref{sec:results_cand_prun}.
	}
	\label{fig:coarse_medtype_outcomes}
\end{figure}

\subsection{Impact of Semantic Type Prediction on Candidate Generation}
\label{sec:results_cand_prun}
%\subsection{On Quality of \pubmed{} Corpus}
%\label{sec:results_pubmed_quality}

The preceding sections have shown that semantic type filtering consistently improves
entity linking performance when using the candidate scoring methods provided in
each of our evaluated information extraction tools. However, candidate ranking and
disambiguation are active areas of research
\cite{triplet-network,Pesaranghader2019}, and the modular nature of both our \method{} model and the semantic type filtering task makes
it easy to incorporate type filtering into any entity extraction pipeline. We therefore
investigated the impact of semantic type prediction in filtering out
over-generated candidate concepts, in order to understand how type filtering simplifies the final disambiguation task.

\noindent \textbf{Semantic type-based pruning consistently reduces the candidate
	set size.} Figure~\ref{fig:coarse_medtype_outcomes} illustrates the outcomes of
type-based pruning on the candidate set sizes for both the 38,234 samples in the
MedMentions test set where \scispacy{} included the correct CUI in its candidate
set and the 21,388 where it did not.  Oracle type information, representing the upper bound of what type-based
pruning can achieve, reduces the candidate set size in over 75\% of ``Correct candidate present'' cases at the
coarse level, and directly solves the sense disambiguation problem in 44\% of
cases.  Fine-grained typing, not shown in Figure~\ref{fig:coarse_medtype_outcomes}, only slightly improves these results---candidate set size reduction in 81\% of cases, full disambiguation in 54\%---while
significantly complicating the type prediction problem, further supporting our
choice of coarse labels for \method{}.  \method{}, in turn, achieves most of the
reductions in candidate set size yielded by oracle information, and the
performance improvements shown in Table~\ref{tab:main_results} clearly demonstrate the practical gains from this filtering.  \method{} further considerably
reduces the number of type mispredictions over the best baseline, as seen also in
Table~\ref{tab:type_pred}.
%Appendix Figure~\ref*{fig:coarse_filtering_impact} provides further details on the degree of reduction in candidate set size.

\begin{figure}[t]
	\centering
	\includegraphics[width=0.9\linewidth]{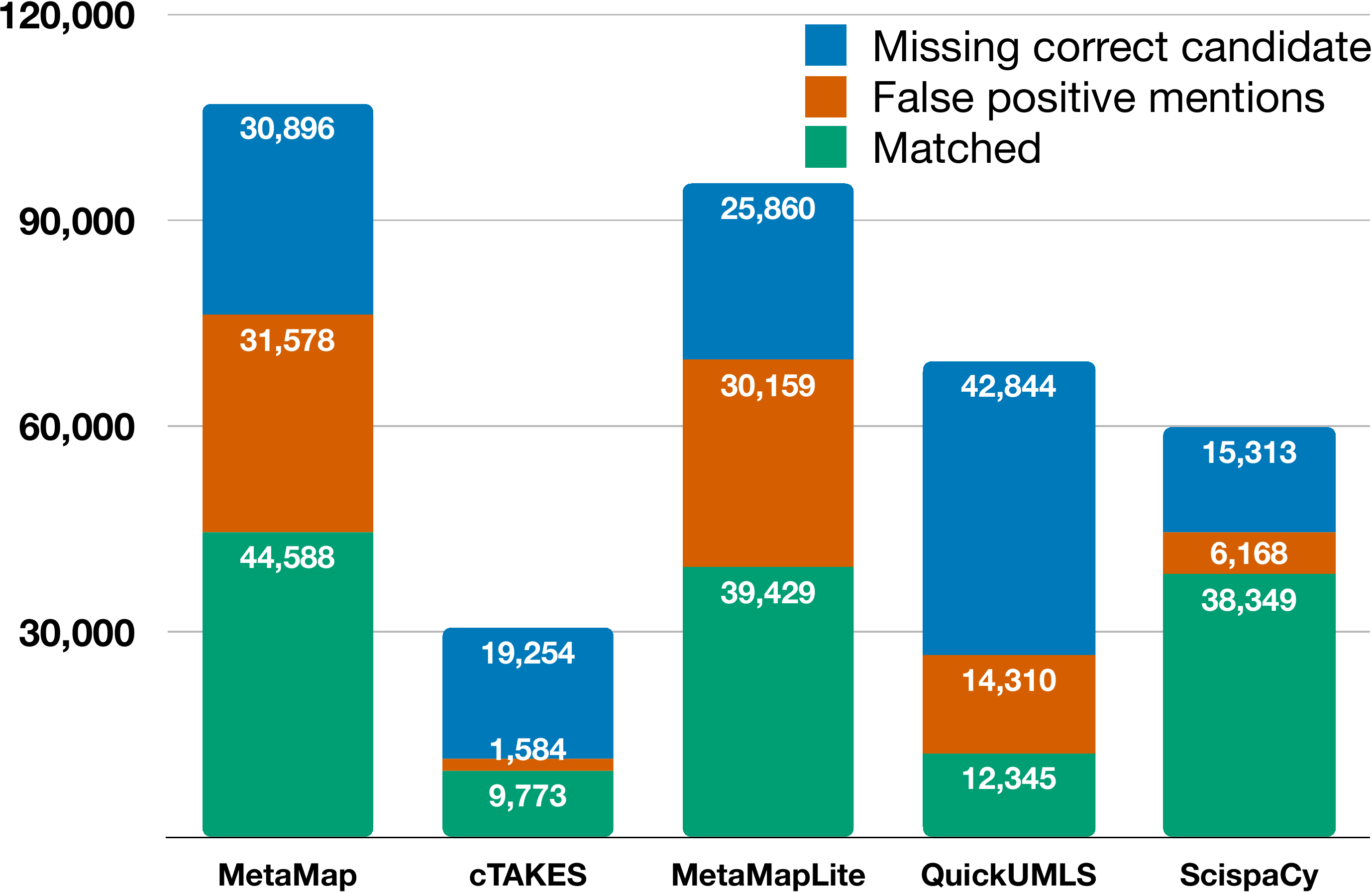}
	\caption{Error analysis of output predictions from all information extraction tools on
		the MedMentions test set (annotated set size: 70,405 mentions). False positive mentions are spurious entity spans extracted by the tools; Missing correct candidate cases indicate exclusion of the correct entity from the returned candidate list.
		Matched indicates that neither of these errors were present. Refer to Section \ref{sec:results_cand_prun} for details.}
	\label{fig:NER_CG_Errors}
\end{figure}
\begin{figure*}
	\centering
	\includegraphics[width=0.9\textwidth]{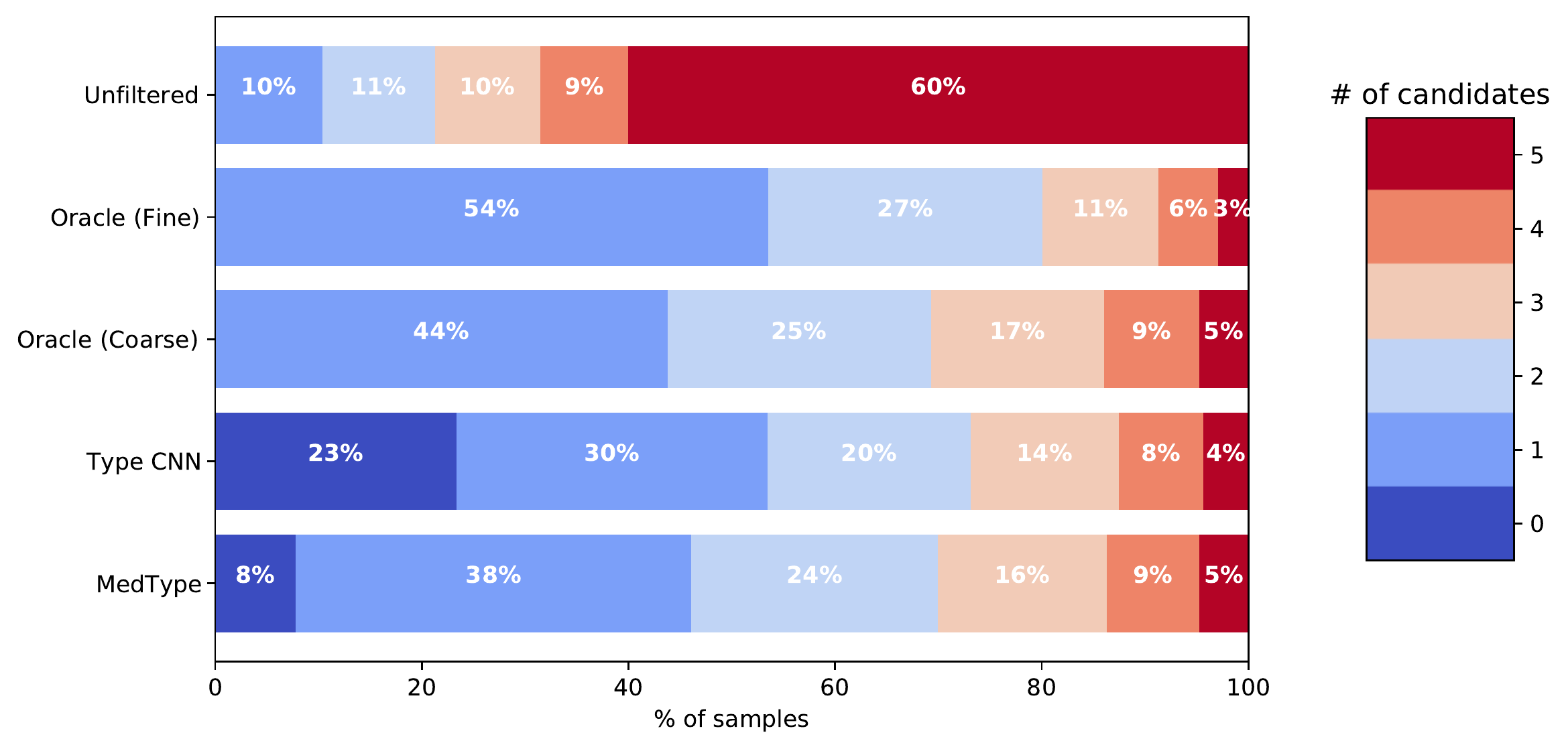}
	\caption{Distribution of candidate set sizes in \medmen{} using \scispacy{},
		comparing unfiltered concepts to candidate sets filtered using
		semantic type prediction strategies. Only mentions predicted by \scispacy{} that included the correct CUI in the candidate set are included. Larger bars to the left-hand side of the figure indicate greater reductions in candidate set size.}
	\label{fig:coarse_filtering_impact}
\end{figure*}

%\vspace{-4mm}
\ \\
\noindent \textbf{\method{} can help improve the full extraction pipeline.} 
Failures can occur at all three stages of entity extraction: mention detection (NER), candidate
generation, and disambiguation. Figure~\ref{fig:NER_CG_Errors} illustrates
the number of medical concepts extracted by the information extraction tools we used in the
MedMentions test set, broken down into (1) false positive mentions, where the mention detection stage of the pipeline produced
a false positive entity span; (2) missing correct candidates, where the candidate generation phase of the pipeline
did not include the correct entity in the candidate list; and (3) matches, where
the tool found a valid span and included the correct entity in the candidate set. The
five tools evaluated varied widely in the number of entities output, but in all
cases include a significant number of both mention detection and candidate generation errors.
In addition to \method{}'s utility in reducing candidate set sizes, which allows
for broader-coverage candidate generation methods, we also observe that in all
cases where a false positive mention was produced, \method{} classified it as a
\textit{None} type; this indicates clear utility in incorporating \method{} as a
component of any system to filter out false positives in NER.

\ \\
\noindent \textbf{Degree of candidate set size reduction from semantic type filtering.}
Figure~\ref{fig:coarse_filtering_impact} expands the analyses presented in
Figure~\ref{fig:coarse_medtype_outcomes} to show the detailed distribution of
the candidate set sizes within the predicted samples of \medmen{} that included the correct candidate, comparing oracle type filtering strategies to \method{} and the best type prediction baseline.   \scispacy{}, presented here as the best-performing information extraction tool on MedMentions, limits its
output candidate set to 5 by default; however, all tools used displayed similar
behavior in our experiments.

\section{Discussion}
\label{sec:discussion}

We have demonstrated that semantic type filtering is a valuable addition to NLP pipelines for broad-coverage biomedical information extraction.  \showedits{We discuss broader impacts of \method{} in biomedical NLP in Section~\ref{sec:broader-impacts}, and other approaches to semantic type filtering in Section~\ref{sec:other-approaches}. We further highlight the contributions of our novel \wiki{} and \pubmed{} datasets for biomedical concept normalization research in Section~\ref{sec:dataset-impacts}, and note potential effects of biased data in Section~\ref{sec:bias}. Finally, we discuss two further implications of our findings for continued research on this important use case: the choice of information extraction tool for a given setting (Section~\ref{sec:toolkit-comparison}), and opportunities for further research synthesizing semantic type prediction and disambiguation (Section~\ref{sec:disambiguation-research})}.

\subsection{Broader applicability of \method{} in biomedical NLP}
\label{sec:broader-impacts}
\showedits{Identifying mentions of biomedical concepts in text is one of the fundamental building blocks of biomedical NLP. As a result, a wide variety of highly heterogeneous methods have been developed to perform concept identification \cite{Kersloot2020}. As a fully modular component which takes as input a set of candidates and returns a set of candidate as output, \method{} can be easily incorporated into any type of medical concept recognition system that uses a set of candidate concepts. Such systems are key elements of NLP pipelines for diverse applications, such as adverse drug event detection \cite{Chen2020c}, biosurveillance \cite{Oliveira2020}, and patient phenotyping \cite{Afzal2018}. Morever, many biomedical NLP applications that do not use concept-level mapping nevertheless make use of coarse-grained type information \cite{DeBruijn2011,Wei2020}, which the modular type prediction component of \method{} is well positioned to enhance. \method{}'s role in refining and organizing medical information in text thus makes it a valuable addition to a wide variety of biomedical NLP pipelines, and its fine-tuning process can be easily used to adapt it to any dataset.}

\subsection{Generalizability and other approaches to semantic type filtering}
\label{sec:other-approaches}
\showedits{Beyond alignment to the UMLS and other controlled vocabularies, biomedical NLP systems often employ custom typologies for specific applications, such as in analyzing radiology notes \cite{Hassanpour2016} or functional status information \cite{Thieu2021}. As seen in our experiments without pretraining, \method{} can be trained to predict the semantic types of a dataset using a relatively small amount of data (i.e., hundreds of documents). Thus, \method{} could be deployed as an element of NLP pipelines with custom typologies as well, via an intermediate step of training the type prediction model on the task-specific dataset.}

\showedits{More broadly, semantic type filtering as presented here is not specific to our \method{} implementation; a variety of approaches could be used within the general framework described in Section~\ref{sec:overview}. Past work has leveraged rule-based and lexical approaches for semantic type prediction \cite{metamap,Hassanpour2016}, or incorporated semantic type prediction as one element of a larger joint neural system \cite{Medlinker}. \method{} serves as a strong baseline for additional research in this area.}

\subsection{\wiki{} and \pubmed{} are valuable resources for biomedical concept normalization research}
\label{sec:dataset-impacts}
\showedits{The expense and difficulty of producing large-scale datasets is a major limiting factor in biomedical NLP research. This is particularly the case for the labor-intensive task of annotating datasets for biomedical concept normalization, where information density is high and there are thousands of candidate concepts to choose from in the annotation process. The \wiki{} and \pubmed{} datasets introduced in this work are a step towards alleviating this problem, presenting millions of annotated concept mentions with a high diversity in semantic type coverage. While these datasets were automatically created and therefore subject to noise from the link mapping process (\wiki{}) and from distant supervision (\pubmed{}), our evaluation of them shows the annotations to be a high-quality silver standard, which can serve as a valuable resource for further research on semantic type prediction and biomedical concept normalization.\footnote{
    We note that \wiki{} and \pubmed{} should not, however, be used to train biomedical mention detection (NER) systems, as the automated annotation process emphasized precision over recall and many potentially valid concept mentions were not included due to missing links (wiki{}) or MeSH headers (\pubmed{}).
}
}

\subsection{Potential effects of biased data on \method{} and novel datasets}
\label{sec:bias}
\showedits{The effects of biased data and algorithms in producing biased AI systems (including medical AI systems) is an important and rapidly-growing area of inquiry \cite{bender2021on,Obermeyer2019}. While \method{} is not directly predicting sensitive information related to patients, or decisions about their treatment, it is nonetheless worth noting potential sources of bias that may be reflected in the outcomes of this study. Two interrelated types of bias are important to discuss: demographic bias (e.g., racial or gender bias) and statistical bias (in the sense of modeling the characteristics of one dataset over another). One major contributing factor to demographic bias in NLP systems is a lack of representatively diverse data; by learning the characteristics of data produced by a subset of the population, the resulting models are less effective in more diverse settings \cite{blodgett2017racial,Koenecke2020}. A significant portion of biomedical NLP research (including many of the datasets used in this article) relies on PubMed---which reflects racial disparities in scientific funding and publication \cite{Ginther2011}---and Wikipedia---which exhibits both racial and gender biases in the presentation of information \cite{Adams2019,Ezell2021}. These biases thus have the potential to be propagated in terms of the different sets of language in which NLP models will be most effective. From a more statistical sense, models trained on one genre of text (such as Wikipedia) generally show some performance degradation when applied to text from other genres (such as PubMed). Investigating potential biases in biomedical NLP systems for information extraction is an important direction to continue in future work.}
%\subsection{Evaluation mismatch}
%\label{sec:evaluation-mismatch}

%Blah blah
%\begin{itemize}
%    \item Things that are out of scope
%    \item Even in-scope things may change in performance between R\&D (focused) and application (built for broad)
%\end{itemize}

%\begin{table}[!t]
%	\centering
%	\small
%	\begin{tabular}{lcccccccc}
%		\toprule
%		\multicolumn{1}{c}{}  & \multicolumn{1}{c}{\textbf{\ncbi{}}}         & \multicolumn{1}{c}{\textbf{\cdr{}}}         & \multicolumn{1}{c}{\textbf{\sharecorpus{}}}\\
%		\midrule
%		\metamap{}      &   9008/10140  &  40473/51992   &  79738/83532 \\
%		\ctakes{}       &   2690/3891   &  9712/22109    &  28024/32240 \\
%		\metamaplite{}  &   8680/9573   &  37061/46650   &  65248/68434 \\
%		\quickumls{}    &   7917/8797   &  36159/45433   &  58747/61751 \\
%		\scispacy{}     &   5527/6505   &  24057/32511   &  42486/45206 \\
%		\midrule
%	\end{tabular}
%	\caption{\label{tab:false_positive} False positive frequency of different entity linkers.} \vspace{-4mm}
%\end{table}

%\subsection{Pipeline evaluation}
%\label{sec:pipeline-evaluation}

%Blah blah

%\begin{itemize}
%    \item Internal validity vs external validity
%\end{itemize}

\subsection{Contribution of semantic type filtering by information extraction toolkit}
\label{sec:toolkit-comparison}

While our results show consistent improvements in information extraction performance from integrating semantic type prediction, the effect size varies from toolkit to toolkit and genre to genre.  For example, improvements in \metamap{} performance are 1\% or less for \ncbi{}, \cdr{}, and \sharecorpus{}, while \quickumls{} performance noticeably improves on all datasets but \ncbi{}, and \scispacy{} and \metamaplite{} show large improvements from semantic type filtering across the board. These differences are in large part attributable to differences in the size of the candidate set produced by each toolkit; for example, \ctakes{}, which sees small relative improvements from type filtering, produces the fewest average candidates per mention of the tools we used, while \scispacy{} (as illustrated in Figure~\ref{fig:coarse_filtering_impact}) produces its built-in maximum of 5 candidates for the majority of samples analyzed. This indicates that revisiting candidate generation strategies, using semantic type filtering to balance out more permissive candidate generation, is a worthwhile direction for improving coverage in biomedical information extraction.

\subsection{Opportunities for disambiguation research using semantic type filtering}
\label{sec:disambiguation-research}

Disambiguating the candidate concepts produced by medical entity extraction pipelines has been a long-standing area of research, with several tools developed to integrate with existing pipelines.  The YTEX suite of algorithms \cite{Garla2011,Garla2012} extends both \metamap{} and \ctakes{} with a disambiguation module that helps to reduce noise considerably, although \cite{Osborne2013} found that it often over-filtered correct concepts.  There has also been significant research in recent years on developing standalone models for disambiguation, using co-occurrence and feature-based approaches \cite{henry-etal-2017-evaluating,antunes2017supervised,DUQUE20189} as well as neural models \cite{Pesaranghader2019,Zhang2019a}.  Medical concept normalization more broadly has also become an increasing research focus \cite{henry2020n2c2,Tutubalina2018}, with significant opportunities for disambiguation research \cite{newman-griffis2020jamia}.

\method{}, and the semantic type filtering task more broadly, can be easily combined with any of these approaches to create a multi-stage filtering strategy for the disambiguation stage of the information extraction pipeline. \method{} performs coarse filtering to a high-confidence set based on predicted type, a key step for narrowing down over-generated candidate sets in both open-ended deep learning systems and dictionary-based pipelines built for broad coverage; disambiguation methods can then perform a fine-grained selection of the correct candidate to further improve entity linking performance.  We highlight this as an important direction for future work on medical entity linking.

\subsection{Limitations of this study}
\label{sec:limitations}

\revtwoedits{\method{} consistently improves the performance of the medical entity linking systems we evaluated. However, this study has some limitations that can help to guide further research on medical entity linking methods. While our use of coarse-grained semantic types simplified the type prediction task and removed the issue of multiple valid types for UMLS concepts, these semantic groups can be overly broad in practice (e.g., combining symptoms and diagnoses into a single category) and may be qualitatively undesirable. Our fine-grained oracle results in Table~\ref{tab:main_results} also showed frequent improvement over the coarse-grained oracle, particularly in the heterogeneous \medmen{} dataset, suggesting further potential improvement from a more granular type prediction system.}

\revtwoedits{In addition, while \method{} helps to correct for candidate generation errors by pruning out all candidate concepts of the wrong type, it cannot identify a candidate that was not generated in the first place. Similarly, a candidate selection algorithm that improperly scores candidate concepts within a single semantic type will not be affected by \method{}. Future research can leverage the value of semantic type filtering to take advantage of broad-coverage candidate generation approaches to improve recall, and fine-grained candidate scoring algorithms focusing on specific semantic types to improve precision.}

\revtwoedits{For application purposes in biomedical settings, explainability and system accountability are often of high importance. Providing explanations for the opaque outputs of deep neural network models in medical settings remains a significant challenge \cite{feng-etal-2020-explainable}, and there is an active debate over how explainable such models can be \cite{Rudin2019}. Providing insight into \method{} successes and failures, and options for users to adjust system parameters for their specific settings, will be an important part of supporting broader adoption of biomedical NLP technologies like \method{}.}

\revtwoedits{Finally, our results are necessarily limited by the homogeneity of some of our datasets. Of the evaluation sets, only \medmen{} includes samples of all semantic types; our picture of \method{}'s impact is thus incomplete for other PubMed data or for clinical language.}
\section{Conclusion}
\label{sec:conclusion}

Broad-coverage information extraction from biomedical text is an important application area for biomedical NLP tools, and one which poses significant challenges in the scale and diversity of information to extract.  To help address these challenges, we introduced semantic type prediction as a modular component of biomedical information extraction pipelines, and presented \method{}, a state-of-the-art neural model for semantic type prediction.  We demonstrated that semantic type prediction measurably improves information extraction performance on four benchmark datasets from different genres of text and types of information, and that these improvements are observed consistently when integrating type prediction into five commonly-used tools for biomedical information extraction.  We further presented two new, automatically-created datasets, \wiki{} and \pubmed{}, which are significantly larger than any previous resources for medical entity linking research. \showedits{While the automated annotation processes to create these datasets introduced some noise, they retained high fidelity in their annotations (over 84\% precision for \pubmed{}, and 91\% CUI-level accuracy in \wiki{}) and our results demonstrate their utility in training semantic type prediction models}. We make the source code for our experiments and our two novel datasets available to the community from \url{http://github.com/svjan5/medtype}, as a resource for further research on biomedical information extraction.
%We presented \method{}, a fully modular system for improving medical entity linking by filtering out irrelevant candidate concepts based on the predicted semantic type of an entity mention.  \method{} improves entity linking performance for a variety of popular medical entity extraction toolkits across several benchmark datasets, clearly demonstrating the utility of a type-based filtering step in medical entity linking.  We also presented \pubmed{}, a novel large-scale distantly-supervised dataset of medical entity mentions constructed from biomedical abstracts in PubMed.  Pre-training on our proposed dataset substantively improves \method{} performance, and we share the source code and proposed dataset with the community as a resource for medical entity linking research.

\section*{Contributions}
Shikhar Vashishth: Conceptualization, Methodology, Investigation, Writing. Denis Newman-Griffis: Conceptualization, Visualization, Writing. Rishabh Joshi: Methodology, Investigation. Ritam Dutt: Methodology, Investigation. Carolyn Rosé: Supervision, Funding Acquisition.

%\section*{Declaration of Competing Interest}
%TBD

\section*{Acknowledgments}
The research reported in this publication was supported in part by National Science Foundation Grants (IIS 1917668 and IIS 1822831), Dow Chemical, and the National Library of Medicine of the National Institutes of Health under award number T15 LM007059.

%\section*{References}
\bibliography{references}

\begin{thebibliography}{100}
\expandafter\ifx\csname url\endcsname\relax
  \def\url#1{\texttt{#1}}\fi
\expandafter\ifx\csname urlprefix\endcsname\relax\def\urlprefix{URL }\fi
\expandafter\ifx\csname href\endcsname\relax
  \def\href#1#2{#2} \def\path#1{#1}\fi

\bibitem{koleck2019natural}
T.~A. Koleck, C.~Dreisbach, P.~E. Bourne, S.~Bakken,
  \href{https://doi.org/10.1093/jamia/ocy173}{{Natural language processing of
  symptoms documented in free-text narratives of electronic health records: a
  systematic review}}, Journal of the American Medical Informatics Association
  26~(4) (2019) 364--379.
\newblock \href
  {http://arxiv.org/abs/https://academic.oup.com/jamia/article-pdf/26/4/364/34151341/ocy173.pdf}
  {\path{arXiv:https://academic.oup.com/jamia/article-pdf/26/4/364/34151341/ocy173.pdf}},
  \href {http://dx.doi.org/10.1093/jamia/ocy173}
  {\path{doi:10.1093/jamia/ocy173}}.
\newline\urlprefix\url{https://doi.org/10.1093/jamia/ocy173}

\bibitem{young2019systematic}
I.~J.~B. Young, S.~Luz, N.~Lone,
  \href{http://www.sciencedirect.com/science/article/pii/S1386505619302370}{{A
  systematic review of natural language processing for classification tasks in
  the field of incident reporting and adverse event analysis}}, International
  Journal of Medical Informatics 132 (2019) 103971.
\newblock \href
  {http://dx.doi.org/https://doi.org/10.1016/j.ijmedinf.2019.103971}
  {\path{doi:https://doi.org/10.1016/j.ijmedinf.2019.103971}}.
\newline\urlprefix\url{http://www.sciencedirect.com/science/article/pii/S1386505619302370}

\bibitem{ctakes}
G.~K. Savova, J.~J. Masanz, P.~V. Ogren, J.~Zheng, S.~Sohn, K.~C.
  Kipper-Schuler, C.~G. Chute,
  \href{https://doi.org/10.1136/jamia.2009.001560}{{Mayo clinical Text Analysis
  and Knowledge Extraction System (cTAKES): architecture, component evaluation
  and applications.}}, Journal of the American Medical Informatics Association
  17~(5) (2010) 507--513.
\newblock \href {http://dx.doi.org/10.1136/jamia.2009.001560}
  {\path{doi:10.1136/jamia.2009.001560}}.
\newline\urlprefix\url{https://doi.org/10.1136/jamia.2009.001560}

\bibitem{garg2019automating}
R.~Garg, E.~Oh, A.~Naidech, K.~Kording, S.~Prabhakaran,
  \href{http://www.sciencedirect.com/science/article/pii/S1052305719300485}{{Automating
  Ischemic Stroke Subtype Classification Using Machine Learning and Natural
  Language Processing}}, Journal of Stroke and Cerebrovascular Diseases 28~(7)
  (2019) 2045--2051.
\newblock \href
  {http://dx.doi.org/https://doi.org/10.1016/j.jstrokecerebrovasdis.2019.02.004}
  {\path{doi:https://doi.org/10.1016/j.jstrokecerebrovasdis.2019.02.004}}.
\newline\urlprefix\url{http://www.sciencedirect.com/science/article/pii/S1052305719300485}

\bibitem{kochar2020pretreatment}
B.~Kochar, W.~Cai, A.~Cagan, A.~N. Ananthakrishnan,
  \href{http://www.sciencedirect.com/science/article/pii/S0016508520302432}{Pretreatment
  frailty is independently associated with increased risk of infections after
  immunosuppression in patients with inflammatory bowel diseases},
  Gastroenterology 158~(8) (2020) 2104 -- 2111.e2.
\newblock \href
  {http://dx.doi.org/https://doi.org/10.1053/j.gastro.2020.02.032}
  {\path{doi:https://doi.org/10.1053/j.gastro.2020.02.032}}.
\newline\urlprefix\url{http://www.sciencedirect.com/science/article/pii/S0016508520302432}

\bibitem{metamap}
A.~R. Aronson, F.-M. Lang,
  \href{http://www.pubmedcentral.nih.gov/articlerender.fcgi?artid=2995713{\&}tool=pmcentrez{\&}rendertype=abstract}{{An
  overview of MetaMap: historical perspective and recent advances.}}, Journal
  of the American Medical Informatics Association : JAMIA 17~(3) (2010)
  229--36.
\newblock \href {http://dx.doi.org/10.1136/jamia.2009.002733}
  {\path{doi:10.1136/jamia.2009.002733}}.
\newline\urlprefix\url{http://www.pubmedcentral.nih.gov/articlerender.fcgi?artid=2995713{\&}tool=pmcentrez{\&}rendertype=abstract}

\bibitem{luo2017natural}
Y.~Luo, W.~K. Thompson, T.~M. Herr, Z.~Zeng, M.~A. Berendsen, S.~R.
  Jonnalagadda, M.~B. Carson, J.~Starren,
  \href{https://doi.org/10.1007/s40264-017-0558-6}{{Natural Language Processing
  for EHR-Based Pharmacovigilance: A Structured Review}}, Drug Safety 40~(11)
  (2017) 1075--1089.
\newblock \href {http://dx.doi.org/10.1007/s40264-017-0558-6}
  {\path{doi:10.1007/s40264-017-0558-6}}.
\newline\urlprefix\url{https://doi.org/10.1007/s40264-017-0558-6}

\bibitem{hua2020health}
M.~Hua, S.~Sadah, V.~Hristidis, P.~Talbot,
  \href{https://doi.org/10.2196/15684}{Health effects associated with
  electronic cigarette use: Automated mining of online forums}, J Med Internet
  Res 22~(1) (2020) e15684.
\newblock \href {http://dx.doi.org/10.2196/15684} {\path{doi:10.2196/15684}}.
\newline\urlprefix\url{https://doi.org/10.2196/15684}

\bibitem{wajsburt2021jbi}
P.~Wajsbürt, A.~Sarfati, X.~Tannier,
  \href{http://www.sciencedirect.com/science/article/pii/S1532046421000137}{Medical
  concept normalization in french using multilingual terminologies and
  contextual embeddings}, Journal of Biomedical Informatics (2021) 103684\href
  {http://dx.doi.org/https://doi.org/10.1016/j.jbi.2021.103684}
  {\path{doi:https://doi.org/10.1016/j.jbi.2021.103684}}.
\newline\urlprefix\url{http://www.sciencedirect.com/science/article/pii/S1532046421000137}

\bibitem{umls}
O.~Bodenreider,
  \href{http://nar.oxfordjournals.org/lookup/doi/10.1093/nar/gkh061}{{The
  Unified Medical Language System (UMLS): integrating biomedical terminology}},
  Nucleic Acids Research 32~(90001) (2004) D267--D270.
\newblock \href {http://dx.doi.org/10.1093/nar/gkh061}
  {\path{doi:10.1093/nar/gkh061}}.
\newline\urlprefix\url{http://nar.oxfordjournals.org/lookup/doi/10.1093/nar/gkh061}

\bibitem{Jovanovic2017}
J.~Jovanovi{\'{c}}, E.~Bagheri, {Semantic annotation in biomedicine: The
  current landscape}, Journal of Biomedical Semantics 8~(1) (2017) 1--18.
\newblock \href {http://dx.doi.org/10.1186/s13326-017-0153-x}
  {\path{doi:10.1186/s13326-017-0153-x}}.

\bibitem{Luo2019}
Y.-F. Luo, W.~Sun, A.~Rumshisky,
  \href{http://www.sciencedirect.com/science/article/pii/S1532046419300504}{{MCN:
  A comprehensive corpus for medical concept normalization}}, Journal of
  Biomedical Informatics 92 (2019) 103132.
\newblock \href {http://dx.doi.org/https://doi.org/10.1016/j.jbi.2019.103132}
  {\path{doi:https://doi.org/10.1016/j.jbi.2019.103132}}.
\newline\urlprefix\url{http://www.sciencedirect.com/science/article/pii/S1532046419300504}

\bibitem{Travers2006}
D.~A. Travers, S.~W. Haas,
  \href{https://doi.org/10.1197/j.aem.2006.06.054}{{Unified Medical Language
  System Coverage of Emergency-medicine Chief Complaints}}, Academic Emergency
  Medicine 13~(12) (2006) 1319--1323.
\newblock \href {http://dx.doi.org/10.1197/j.aem.2006.06.054}
  {\path{doi:10.1197/j.aem.2006.06.054}}.
\newline\urlprefix\url{https://doi.org/10.1197/j.aem.2006.06.054}

\bibitem{Reategui2018}
R.~Re{\'{a}}tegui, S.~Ratt{\'{e}},
  \href{https://doi.org/10.1186/s12911-018-0654-2}{{Comparison of MetaMap and
  cTAKES for entity extraction in clinical notes}}, BMC Medical Informatics and
  Decision Making 18~(3) (2018) 74.
\newblock \href {http://dx.doi.org/10.1186/s12911-018-0654-2}
  {\path{doi:10.1186/s12911-018-0654-2}}.
\newline\urlprefix\url{https://doi.org/10.1186/s12911-018-0654-2}

\bibitem{Tutubalina2018}
E.~Tutubalina, Z.~Miftahutdinov, S.~Nikolenko, V.~Malykh,
  \href{http://www.sciencedirect.com/science/article/pii/S1532046418301126}{{Medical
  concept normalization in social media posts with recurrent neural networks}},
  Journal of Biomedical Informatics 84 (2018) 93--102.
\newblock \href {http://dx.doi.org/https://doi.org/10.1016/j.jbi.2018.06.006}
  {\path{doi:https://doi.org/10.1016/j.jbi.2018.06.006}}.
\newline\urlprefix\url{http://www.sciencedirect.com/science/article/pii/S1532046418301126}

\bibitem{Soysal2018}
E.~Soysal, J.~Wang, M.~Jiang, Y.~Wu, S.~Pakhomov, H.~Liu, H.~Xu,
  \href{http://dx.doi.org/10.1093/jamia/ocx132}{{CLAMP – a toolkit for
  efficiently building customized clinical natural language processing
  pipelines}}, Journal of the American Medical Informatics Association 25~(3)
  (2018) 331--336.
\newblock \href {http://dx.doi.org/10.1093/jamia/ocx132}
  {\path{doi:10.1093/jamia/ocx132}}.
\newline\urlprefix\url{http://dx.doi.org/10.1093/jamia/ocx132}

\bibitem{Zhao2019a}
S.~Zhao, T.~Liu, S.~Zhao, F.~Wang,
  \href{https://doi.org/10.1609/aaai.v33i01.3301817}{{A Neural Multi-Task
  Learning Framework to Jointly Model Medical Named Entity Recognition and
  Normalization}}, in: Proceedings of the Thirty-Third AAAI Conference on
  Artificial Intelligence, AAAI Press, 2019, pp. 817--824.
\newblock \href {http://dx.doi.org/10.1609/aaai.v33i01.3301817}
  {\path{doi:10.1609/aaai.v33i01.3301817}}.
\newline\urlprefix\url{https://doi.org/10.1609/aaai.v33i01.3301817}

\bibitem{triplet-network}
I.~Mondal, S.~Purkayastha, S.~Sarkar, P.~Goyal, J.~Pillai, A.~Bhattacharyya,
  M.~Gattu, \href{https://www.aclweb.org/anthology/W19-1912}{Medical entity
  linking using triplet network}, in: Proceedings of the 2nd Clinical Natural
  Language Processing Workshop, Association for Computational Linguistics,
  Minneapolis, Minnesota, USA, 2019, pp. 95--100.
\newblock \href {http://dx.doi.org/10.18653/v1/W19-1912}
  {\path{doi:10.18653/v1/W19-1912}}.
\newline\urlprefix\url{https://www.aclweb.org/anthology/W19-1912}

\bibitem{amb1}
M.~Weeber, J.~G. Mork, A.~R. Aronson,
  \href{https://pubmed.ncbi.nlm.nih.gov/11825285}{Developing a test collection
  for biomedical word sense disambiguation}, Proceedings. AMIA Symposium (2001)
  746--750.
\newline\urlprefix\url{https://pubmed.ncbi.nlm.nih.gov/11825285}

\bibitem{amb2}
G.~K. Savova, A.~R. Coden, I.~L. Sominsky, R.~Johnson, P.~V. Ogren, P.~C.
  de~Groen, C.~G. Chute,
  \href{http://www.sciencedirect.com/science/article/pii/S1532046408000245}{Word
  sense disambiguation across two domains: Biomedical literature and clinical
  notes}, Journal of Biomedical Informatics 41~(6) (2008) 1088 -- 1100.
\newblock \href {http://dx.doi.org/https://doi.org/10.1016/j.jbi.2008.02.003}
  {\path{doi:https://doi.org/10.1016/j.jbi.2008.02.003}}.
\newline\urlprefix\url{http://www.sciencedirect.com/science/article/pii/S1532046408000245}

\bibitem{newman-griffis2020jamia}
D.~Newman-Griffis, G.~Divita, B.~Desmet, A.~Zirikly, C.~P. Rosé,
  E.~Fosler-Lussier, \href{https://doi.org/10.1093/jamia/ocaa269}{{Ambiguity in
  medical concept normalization: An analysis of types and coverage in
  electronic health record datasets}}, Journal of the American Medical
  Informatics Association 28 (2021) 516--532.
\newblock \href
  {http://arxiv.org/abs/https://academic.oup.com/jamia/advance-article-pdf/doi/10.1093/jamia/ocaa269/34908292/ocaa269.pdf}
  {\path{arXiv:https://academic.oup.com/jamia/advance-article-pdf/doi/10.1093/jamia/ocaa269/34908292/ocaa269.pdf}},
  \href {http://dx.doi.org/10.1093/jamia/ocaa269}
  {\path{doi:10.1093/jamia/ocaa269}}.
\newline\urlprefix\url{https://doi.org/10.1093/jamia/ocaa269}

\bibitem{deeptype}
J.~R. Raiman, O.~M. Raiman, Deeptype: multilingual entity linking by neural
  type system evolution, in: Thirty-Second AAAI Conference on Artificial
  Intelligence, 2018.

\bibitem{med_data_annotation}
J.~A. Fries, P.~Varma, V.~S. Chen, K.~Xiao, H.~Tejeda, P.~Saha, J.~Dunnmon,
  H.~Chubb, S.~Maskatia, M.~Fiterau, S.~Delp, E.~Ashley, C.~R{\'e}, J.~R.
  Priest, Weakly supervised classification of aortic valve malformations using
  unlabeled cardiac mri sequences, Nature Communications 10~(1) (2019) 3111.

\bibitem{med_data_issues}
K.~J. Cios, G.~W. Moore,
  \href{http://www.sciencedirect.com/science/article/pii/S0933365702000490}{Uniqueness
  of medical data mining}, Artificial Intelligence in Medicine 26~(1) (2002) 1
  -- 24, medical Data Mining and Knowledge Discovery.
\newblock \href
  {http://dx.doi.org/https://doi.org/10.1016/S0933-3657(02)00049-0}
  {\path{doi:https://doi.org/10.1016/S0933-3657(02)00049-0}}.
\newline\urlprefix\url{http://www.sciencedirect.com/science/article/pii/S0933365702000490}

\bibitem{end-to-end-EL}
N.~Kolitsas, O.-E. Ganea, T.~Hofmann,
  \href{https://www.aclweb.org/anthology/K18-1050}{End-to-end neural entity
  linking}, in: Proceedings of the 22nd Conference on Computational Natural
  Language Learning, Association for Computational Linguistics, Brussels,
  Belgium, 2018, pp. 519--529.
\newblock \href {http://dx.doi.org/10.18653/v1/K18-1050}
  {\path{doi:10.18653/v1/K18-1050}}.
\newline\urlprefix\url{https://www.aclweb.org/anthology/K18-1050}

\bibitem{joint-learning}
P.~H. Martins, Z.~Marinho, A.~F.~T. Martins,
  \href{https://www.aclweb.org/anthology/P19-2026}{Joint learning of named
  entity recognition and entity linking}, in: Proceedings of the 57th Annual
  Meeting of the Association for Computational Linguistics: Student Research
  Workshop, Association for Computational Linguistics, Florence, Italy, 2019,
  pp. 190--196.
\newblock \href {http://dx.doi.org/10.18653/v1/P19-2026}
  {\path{doi:10.18653/v1/P19-2026}}.
\newline\urlprefix\url{https://www.aclweb.org/anthology/P19-2026}

\bibitem{ELDEN}
P.~Radhakrishnan, P.~Talukdar, V.~Varma,
  \href{https://www.aclweb.org/anthology/N18-1167}{{ELDEN}: Improved entity
  linking using densified knowledge graphs}, in: Proceedings of the 2018
  Conference of the North {A}merican Chapter of the Association for
  Computational Linguistics: Human Language Technologies, Volume 1 (Long
  Papers), Association for Computational Linguistics, New Orleans, Louisiana,
  2018, pp. 1844--1853.
\newblock \href {http://dx.doi.org/10.18653/v1/N18-1167}
  {\path{doi:10.18653/v1/N18-1167}}.
\newline\urlprefix\url{https://www.aclweb.org/anthology/N18-1167}

\bibitem{dynamicgcn-EL}
J.~Wu, R.~Zhang, Y.~Mao, H.~Guo, M.~Soflaei, J.~Huai, Dynamic graph
  convolutional networks for entity linking, in: Proceedings of The Web
  Conference 2020, 2020, pp. 1149--1159.

\bibitem{durrett2014joint}
G.~Durrett, D.~Klein, A joint model for entity analysis: Coreference, typing,
  and linking, Transactions of the association for computational linguistics 2
  (2014) 477--490.

\bibitem{yaghoobzadeh-etal-2017-noise}
Y.~Yaghoobzadeh, H.~Adel, H.~Sch{\"u}tze,
  \href{https://www.aclweb.org/anthology/E17-1111}{Noise mitigation for neural
  entity typing and relation extraction}, in: Proceedings of the 15th
  Conference of the {E}uropean Chapter of the Association for Computational
  Linguistics: Volume 1, Long Papers, Association for Computational
  Linguistics, Valencia, Spain, 2017, pp. 1183--1194.
\newline\urlprefix\url{https://www.aclweb.org/anthology/E17-1111}

\bibitem{das-QA}
R.~Das, M.~Zaheer, S.~Reddy, A.~McCallum,
  \href{https://www.aclweb.org/anthology/P17-2057}{Question answering on
  knowledge bases and text using universal schema and memory networks}, in:
  Proceedings of the 55th Annual Meeting of the Association for Computational
  Linguistics (Volume 2: Short Papers), Association for Computational
  Linguistics, Vancouver, Canada, 2017, pp. 358--365.
\newblock \href {http://dx.doi.org/10.18653/v1/P17-2057}
  {\path{doi:10.18653/v1/P17-2057}}.
\newline\urlprefix\url{https://www.aclweb.org/anthology/P17-2057}

\bibitem{ernie}
Z.~Zhang, X.~Han, Z.~Liu, X.~Jiang, M.~Sun, Q.~Liu,
  \href{https://www.aclweb.org/anthology/P19-1139}{{ERNIE}: Enhanced language
  representation with informative entities}, in: Proceedings of the 57th Annual
  Meeting of the Association for Computational Linguistics, Association for
  Computational Linguistics, Florence, Italy, 2019, pp. 1441--1451.
\newblock \href {http://dx.doi.org/10.18653/v1/P19-1139}
  {\path{doi:10.18653/v1/P19-1139}}.
\newline\urlprefix\url{https://www.aclweb.org/anthology/P19-1139}

\bibitem{ling-etal-2015-design}
X.~Ling, S.~Singh, D.~S. Weld,
  \href{https://www.aclweb.org/anthology/Q15-1023}{Design challenges for entity
  linking}, Transactions of the Association for Computational Linguistics 3
  (2015) 315--328.
\newblock \href {http://dx.doi.org/10.1162/tacl_a_00141}
  {\path{doi:10.1162/tacl_a_00141}}.
\newline\urlprefix\url{https://www.aclweb.org/anthology/Q15-1023}

\bibitem{Aronson1994}
A.~R. Aronson, T.~C. Rindflesch, A.~C. Browne, {Exploiting a Large Thesaurus
  for Information Retrieval}, in: Intelligent Multimedia Information Retrieval
  Systems and Management - Volume 1, RIAO '94, LE CENTRE DE HAUTES ETUDES
  INTERNATIONALES D'INFORMATIQUE DOCUMENTAIRE, Paris, FRA, 1994, pp. 197--216.

\bibitem{semantictype}
I.~Nejadgholi, K.~C. Fraser, B.~De~Bruijn, M.~Li, A.~LaPlante, K.~Zine
  El~Abidine, \href{https://www.aclweb.org/anthology/D19-6219}{Recognizing
  {UMLS} semantic types with deep learning}, in: Proceedings of the Tenth
  International Workshop on Health Text Mining and Information Analysis (LOUHI
  2019), Association for Computational Linguistics, Hong Kong, 2019, pp.
  157--167.
\newblock \href {http://dx.doi.org/10.18653/v1/D19-6219}
  {\path{doi:10.18653/v1/D19-6219}}.
\newline\urlprefix\url{https://www.aclweb.org/anthology/D19-6219}

\bibitem{Medlinker}
D.~Loureiro, A.~M. Jorge, Medlinker: Medical entity linking with neural
  representations and dictionary matching, in: J.~M. Jose, E.~Yilmaz,
  J.~Magalh{\~a}es, P.~Castells, N.~Ferro, M.~J. Silva, F.~Martins (Eds.),
  Advances in Information Retrieval, Springer International Publishing, Cham,
  2020, pp. 230--237.

\bibitem{Pesaranghader2019}
A.~Pesaranghader, S.~Matwin, M.~Sokolova, A.~Pesaranghader,
  \href{https://doi.org/10.1093/jamia/ocy189}{{deepBioWSD: effective deep
  neural word sense disambiguation of biomedical text data}}, Journal of the
  American Medical Informatics Association 26~(5) (2019) 438--446.
\newblock \href {http://dx.doi.org/10.1093/jamia/ocy189}
  {\path{doi:10.1093/jamia/ocy189}}.
\newline\urlprefix\url{https://doi.org/10.1093/jamia/ocy189}

\bibitem{henry2020n2c2}
S.~Henry, Y.~Wang, F.~Shen, O.~Uzuner,
  \href{https://doi.org/10.1093/jamia/ocaa106}{{The 2019 National Natural
  language processing (NLP) Clinical Challenges (n2c2)/Open Health NLP (OHNLP)
  shared task on clinical concept normalization for clinical records}}, Journal
  of the American Medical Informatics Association 27~(10) (2020) 1529--1537.
\newblock \href
  {http://arxiv.org/abs/https://academic.oup.com/jamia/article-pdf/27/10/1529/34153509/ocaa106.pdf}
  {\path{arXiv:https://academic.oup.com/jamia/article-pdf/27/10/1529/34153509/ocaa106.pdf}},
  \href {http://dx.doi.org/10.1093/jamia/ocaa106}
  {\path{doi:10.1093/jamia/ocaa106}}.
\newline\urlprefix\url{https://doi.org/10.1093/jamia/ocaa106}

\bibitem{elmo}
M.~E. Peters, M.~Neumann, M.~Iyyer, M.~Gardner, C.~Clark, K.~Lee,
  L.~Zettlemoyer, Deep contextualized word representations, in: Proc. of NAACL,
  2018.

\bibitem{bert}
J.~Devlin, M.-W. Chang, K.~Lee, K.~Toutanova, Bert: Pre-training of deep
  bidirectional transformers for language understanding, arXiv preprint
  arXiv:1810.04805.

\bibitem{quickumls}
L.~Soldaini, N.~Goharian, Quickumls: a fast, unsupervised approach for medical
  concept extraction, in: MedIR workshop, sigir, 2016, pp. 1--4.

\bibitem{scispacy}
M.~{Neumann}, D.~{King}, I.~{Beltagy}, W.~{Ammar}, {ScispaCy: Fast and Robust
  Models for Biomedical Natural Language Processing}, arXiv e-prints (2019)
  arXiv:1902.07669\href {http://arxiv.org/abs/1902.07669}
  {\path{arXiv:1902.07669}}.

\bibitem{deng-etal-2019-ensemble}
P.~Deng, H.~Chen, M.~Huang, X.~Ruan, L.~Xu,
  \href{https://www.aclweb.org/anthology/D19-5721}{An ensemble {CNN} method for
  biomedical entity normalization}, in: Proceedings of The 5th Workshop on
  BioNLP Open Shared Tasks, Association for Computational Linguistics, Hong
  Kong, China, 2019, pp. 143--149.
\newblock \href {http://dx.doi.org/10.18653/v1/D19-5721}
  {\path{doi:10.18653/v1/D19-5721}}.
\newline\urlprefix\url{https://www.aclweb.org/anthology/D19-5721}

\bibitem{Ji2020}
Z.~Ji, Q.~Wei, H.~Xu, \href{https://pubmed.ncbi.nlm.nih.gov/32477646
  https://www.ncbi.nlm.nih.gov/pmc/articles/PMC7233044/}{{BERT-based Ranking
  for Biomedical Entity Normalization}}, AMIA Joint Summits on Translational
  Science proceedings. AMIA Joint Summits on Translational Science 2020 (2020)
  269--277.
\newline\urlprefix\url{https://pubmed.ncbi.nlm.nih.gov/32477646
  https://www.ncbi.nlm.nih.gov/pmc/articles/PMC7233044/}

\bibitem{dsouza-ng-2015-sieve}
J.~D{'}Souza, V.~Ng,
  \href{https://www.aclweb.org/anthology/P15-2049}{Sieve-based entity linking
  for the biomedical domain}, in: Proceedings of the 53rd Annual Meeting of the
  Association for Computational Linguistics and the 7th International Joint
  Conference on Natural Language Processing (Volume 2: Short Papers),
  Association for Computational Linguistics, Beijing, China, 2015, pp.
  297--302.
\newblock \href {http://dx.doi.org/10.3115/v1/P15-2049}
  {\path{doi:10.3115/v1/P15-2049}}.
\newline\urlprefix\url{https://www.aclweb.org/anthology/P15-2049}

\bibitem{Li2017b}
H.~Li, Q.~Chen, B.~Tang, X.~Wang, H.~Xu, B.~Wang, D.~Huang,
  \href{https://doi.org/10.1186/s12859-017-1805-7}{{CNN-based ranking for
  biomedical entity normalization}}, BMC Bioinformatics 18~(11) (2017) 385.
\newblock \href {http://dx.doi.org/10.1186/s12859-017-1805-7}
  {\path{doi:10.1186/s12859-017-1805-7}}.
\newline\urlprefix\url{https://doi.org/10.1186/s12859-017-1805-7}

\bibitem{wang2017pdd}
M.~Wang, J.~Zhang, J.~Liu, W.~Hu, S.~Wang, X.~Li, W.~Liu, Pdd graph: Bridging
  electronic medical records and biomedical knowledge graphs via entity
  linking, in: C.~d'Amato, M.~Fernandez, V.~Tamma, F.~Lecue,
  P.~Cudr{\'e}-Mauroux, J.~Sequeda, C.~Lange, J.~Heflin (Eds.), The Semantic
  Web -- ISWC 2017, Springer International Publishing, Cham, 2017, pp.
  219--227.

\bibitem{sem_grouping}
A.~McCray, A.~Burgun, O.~Bodenreider,
  \href{http://semanticnetwork.nlm.nih.gov/SemGroups/}{Aggregating umls
  semantic types for reducing conceptual complexity.}, Proceedings of Medinfo
  10~(pt 1) (2001) 216--20.
\newline\urlprefix\url{http://semanticnetwork.nlm.nih.gov/SemGroups/}

\bibitem{transformer}
A.~Vaswani, N.~Shazeer, N.~Parmar, J.~Uszkoreit, L.~Jones, A.~N. Gomez, L.~u.
  Kaiser, I.~Polosukhin,
  \href{http://papers.nips.cc/paper/7181-attention-is-all-you-need.pdf}{Attention
  is all you need}, in: I.~Guyon, U.~V. Luxburg, S.~Bengio, H.~Wallach,
  R.~Fergus, S.~Vishwanathan, R.~Garnett (Eds.), Advances in Neural Information
  Processing Systems 30, Curran Associates, Inc., 2017, pp. 5998--6008.
\newline\urlprefix\url{http://papers.nips.cc/paper/7181-attention-is-all-you-need.pdf}

\bibitem{biobert}
J.~Lee, W.~Yoon, S.~Kim, D.~Kim, S.~Kim, C.~H. So, J.~Kang,
  \href{https://doi.org/10.1093/bioinformatics/btz682}{{BioBERT: a pre-trained
  biomedical language representation model for biomedical text mining}},
  Bioinformatics\href {http://dx.doi.org/10.1093/bioinformatics/btz682}
  {\path{doi:10.1093/bioinformatics/btz682}}.
\newline\urlprefix\url{https://doi.org/10.1093/bioinformatics/btz682}

\bibitem{imagenet}
O.~Russakovsky, J.~Deng, H.~Su, J.~Krause, S.~Satheesh, S.~Ma, Z.~Huang,
  A.~Karpathy, A.~Khosla, M.~Bernstein, A.~C. Berg, L.~Fei-Fei, {ImageNet Large
  Scale Visual Recognition Challenge}, International Journal of Computer Vision
  (IJCV) 115~(3) (2015) 211--252.
\newblock \href {http://dx.doi.org/10.1007/s11263-015-0816-y}
  {\path{doi:10.1007/s11263-015-0816-y}}.

\bibitem{snli}
S.~R. Bowman, G.~Angeli, C.~Potts, C.~D. Manning, A large annotated corpus for
  learning natural language inference, in: Proceedings of the 2015 Conference
  on Empirical Methods in Natural Language Processing (EMNLP), Association for
  Computational Linguistics, 2015.

\bibitem{Hirschberg2015}
J.~Hirschberg, C.~D. Manning, {Advances in natural language processing},
  Science 349~(6245) (2015) 261--266.
\newblock \href {http://dx.doi.org/10.1126/science.aaa8685}
  {\path{doi:10.1126/science.aaa8685}}.

\bibitem{wikidata}
D.~Vrande\v{c}i\'{c}, M.~Kr\"{o}tzsch,
  \href{http://doi.acm.org/10.1145/2629489}{Wikidata: A free collaborative
  knowledgebase}, Commun. ACM 57~(10) (2014) 78--85.
\newblock \href {http://dx.doi.org/10.1145/2629489}
  {\path{doi:10.1145/2629489}}.
\newline\urlprefix\url{http://doi.acm.org/10.1145/2629489}

\bibitem{freebase}
K.~Bollacker, C.~Evans, P.~Paritosh, T.~Sturge, J.~Taylor,
  \href{http://doi.acm.org/10.1145/1376616.1376746}{Freebase: A collaboratively
  created graph database for structuring human knowledge}, in: Proceedings of
  the 2008 ACM SIGMOD International Conference on Management of Data, SIGMOD
  '08, ACM, New York, NY, USA, 2008, pp. 1247--1250.
\newblock \href {http://dx.doi.org/10.1145/1376616.1376746}
  {\path{doi:10.1145/1376616.1376746}}.
\newline\urlprefix\url{http://doi.acm.org/10.1145/1376616.1376746}

\bibitem{ncbi_mapping}
R.~D. Page, Linking {NCBI} to wikipedia: a wiki-based approach, PLoS currents 3
  (2011) RRN1228.
\newblock \href {http://dx.doi.org/10.1371/currents.RRN1228}
  {\path{doi:10.1371/currents.RRN1228}}.

\bibitem{wiki_gtruth4}
A.-M. Vercoustre, J.~Pehcevski, J.~A. Thom, Using wikipedia categories and
  links in entity ranking, in: International Workshop of the Initiative for the
  Evaluation of XML Retrieval, Springer, 2007, pp. 321--335.

\bibitem{wiki_gtruth2}
K.~Nakayama, T.~Hara, S.~Nishio, Wikipedia link structure and text mining for
  semantic relation extraction., in: SemSearch, 2008, pp. 59--73.

\bibitem{wiki_gtruth5}
D.~Gillick, S.~Kulkarni, L.~Lansing, A.~Presta, J.~Baldridge, E.~Ie,
  D.~Garcia-Olano, \href{https://www.aclweb.org/anthology/K19-1049}{Learning
  dense representations for entity retrieval}, in: Proceedings of the 23rd
  Conference on Computational Natural Language Learning (CoNLL), Association
  for Computational Linguistics, Hong Kong, China, 2019, pp. 528--537.
\newblock \href {http://dx.doi.org/10.18653/v1/K19-1049}
  {\path{doi:10.18653/v1/K19-1049}}.
\newline\urlprefix\url{https://www.aclweb.org/anthology/K19-1049}

\bibitem{wiki_gtruth3}
A.~Fogarolli, Word sense disambiguation based on wikipedia link structure, in:
  2009 IEEE International Conference on Semantic Computing, 2009, pp. 77--82.
\newblock \href {http://dx.doi.org/10.1109/ICSC.2009.7}
  {\path{doi:10.1109/ICSC.2009.7}}.

\bibitem{wiki_gtruth1}
F.~Wu, D.~S. Weld, \href{https://www.aclweb.org/anthology/P10-1013}{Open
  information extraction using {W}ikipedia}, in: Proceedings of the 48th Annual
  Meeting of the Association for Computational Linguistics, Association for
  Computational Linguistics, Uppsala, Sweden, 2010, pp. 118--127.
\newline\urlprefix\url{https://www.aclweb.org/anthology/P10-1013}

\bibitem{wiki_link_error1}
C.~Wang, R.~Zhang, X.~He, A.~Zhou, Error link detection and correction in
  wikipedia, in: Proceedings of the 25th ACM International on Conference on
  Information and Knowledge Management, 2016, pp. 307--316.

\bibitem{wiki_link_error2}
B.~M. Pateman, C.~G. Johnson, Using the wikipedia link structure to correct the
  wikipedia link structure, Association for Computational Linguistics, 2010.

\bibitem{wiki_quality}
G.~Weaver, B.~Strickland, G.~Crane, Quantifying the accuracy of relational
  statements in wikipedia: a methodology, in: JCDL, Vol.~6, Citeseer, 2006, pp.
  358--358.

\bibitem{medmentions}
S.~Mohan, D.~Li, \href{https://openreview.net/forum?id=SylxCx5pTQ}{Medmentions:
  A large biomedical corpus annotated with {UMLS} concepts}, in: Automated
  Knowledge Base Construction (AKBC), 2019.
\newline\urlprefix\url{https://openreview.net/forum?id=SylxCx5pTQ}

\bibitem{ncbi_data}
R.~I. Do{\u g}an, R.~Leaman, Z.~Lu,
  \href{https://pubmed.ncbi.nlm.nih.gov/24393765}{Ncbi disease corpus: a
  resource for disease name recognition and concept normalization}, Journal of
  biomedical informatics 47 (2014) 1--10.
\newblock \href {http://dx.doi.org/10.1016/j.jbi.2013.12.006}
  {\path{doi:10.1016/j.jbi.2013.12.006}}.
\newline\urlprefix\url{https://pubmed.ncbi.nlm.nih.gov/24393765}

\bibitem{distant_sup}
M.~Mintz, S.~Bills, R.~Snow, D.~Jurafsky, Distant supervision for relation
  extraction without labeled data, in: Proceedings of the Joint Conference of
  the 47th Annual Meeting of the ACL and the 4th International Joint Conference
  on Natural Language Processing of the AFNLP: Volume 2-Volume 2, Association
  for Computational Linguistics, 2009, pp. 1003--1011.

\bibitem{ds_event_extraction}
K.~Reschke, M.~Jankowiak, M.~Surdeanu, C.~D. Manning, D.~Jurafsky, Event
  extraction using distant supervision, in: Language Resources and Evaluation
  Conference (LREC), 2014.

\bibitem{ds_entity_linking}
M.~Fan, Q.~Zhou, T.~F. Zheng, Distant supervision for entity linking, arXiv
  preprint arXiv:1505.03823.

\bibitem{ds_entity_linking_medical}
D.~Newman-Griffis, A.~M. Lai, E.~Fosler-Lussier,
  \href{https://www.aclweb.org/anthology/W18-3026}{Jointly embedding entities
  and text with distant supervision}, in: Proceedings of The Third Workshop on
  Representation Learning for {NLP}, Association for Computational Linguistics,
  Melbourne, Australia, 2018, pp. 195--206.
\newblock \href {http://dx.doi.org/10.18653/v1/W18-3026}
  {\path{doi:10.18653/v1/W18-3026}}.
\newline\urlprefix\url{https://www.aclweb.org/anthology/W18-3026}

\bibitem{ds_noise}
S.~Riedel, L.~Yao, A.~McCallum, Modeling relations and their mentions without
  labeled text, in: Proceedings of the 2010 European Conference on Machine
  Learning and Knowledge Discovery in Databases: Part III, Springer-Verlag,
  Berlin, Heidelberg, 2010.

\bibitem{cdr_data}
J.~Li, Y.~Sun, R.~J. Johnson, D.~Sciaky, C.-H. Wei, R.~Leaman, A.~P. Davis,
  C.~J. Mattingly, T.~C. Wiegers, Z.~Lu,
  \href{https://pubmed.ncbi.nlm.nih.gov/27161011}{Biocreative v cdr task
  corpus: a resource for chemical disease relation extraction}, Database : the
  journal of biological databases and curation 2016 (2016) baw068.
\newblock \href {http://dx.doi.org/10.1093/database/baw068}
  {\path{doi:10.1093/database/baw068}}.
\newline\urlprefix\url{https://pubmed.ncbi.nlm.nih.gov/27161011}

\bibitem{sharecorpus_data}
N.~Elhadad, S.~Pradhan, S.~Gorman, S.~Manandhar, W.~Chapman, G.~Savova,
  \href{https://www.aclweb.org/anthology/S15-2051}{{S}em{E}val-2015 task 14:
  Analysis of clinical text}, in: Proceedings of the 9th International Workshop
  on Semantic Evaluation ({S}em{E}val 2015), Association for Computational
  Linguistics, Denver, Colorado, 2015, pp. 303--310.
\newblock \href {http://dx.doi.org/10.18653/v1/S15-2051}
  {\path{doi:10.18653/v1/S15-2051}}.
\newline\urlprefix\url{https://www.aclweb.org/anthology/S15-2051}

\bibitem{MEDIC}
A.~P. Davis, T.~C. Wiegers, M.~C. Rosenstein, C.~J. Mattingly, Medic: a
  practical disease vocabulary used at the comparative toxicogenomics database,
  Database 2012.

\bibitem{pradhan2015evaluating}
S.~Pradhan, N.~Elhadad, B.~R. South, D.~Martinez, L.~Christensen, A.~Vogel,
  H.~Suominen, W.~W. Chapman, G.~Savova, Evaluating the state of the art in
  disorder recognition and normalization of the clinical narrative, Journal of
  the American Medical Informatics Association 22~(1) (2015) 143--154.

\bibitem{share-2014}
S.~Pradhan, N.~Elhadad, W.~Chapman, S.~Manandhar, G.~Savova,
  \href{https://www.aclweb.org/anthology/S14-2007}{{S}em{E}val-2014 task 7:
  Analysis of clinical text}, in: Proceedings of the 8th International Workshop
  on Semantic Evaluation ({S}em{E}val 2014), Association for Computational
  Linguistics, Dublin, Ireland, 2014, pp. 54--62.
\newblock \href {http://dx.doi.org/10.3115/v1/S14-2007}
  {\path{doi:10.3115/v1/S14-2007}}.
\newline\urlprefix\url{https://www.aclweb.org/anthology/S14-2007}

\bibitem{attention_ner}
S.~Shimaoka, P.~Stenetorp, K.~Inui, S.~Riedel, Neural architectures for
  fine-grained entity type classification, in: Proceedings of the 15th
  Conference of the {E}uropean Chapter of the Association for Computational
  Linguistics: Volume 1, Long Papers, Valencia, Spain, 2017, pp. 1271--1280.

\bibitem{murty-etal-2018-hierarchical}
S.~Murty, P.~Verga, L.~Vilnis, I.~Radovanovic, A.~McCallum,
  \href{https://www.aclweb.org/anthology/P18-1010}{Hierarchical losses and new
  resources for fine-grained entity typing and linking}, in: Proceedings of the
  56th Annual Meeting of the Association for Computational Linguistics (Volume
  1: Long Papers), Association for Computational Linguistics, Melbourne,
  Australia, 2018, pp. 97--109.
\newblock \href {http://dx.doi.org/10.18653/v1/P18-1010}
  {\path{doi:10.18653/v1/P18-1010}}.
\newline\urlprefix\url{https://www.aclweb.org/anthology/P18-1010}

\bibitem{metamaplite}
D.~Demner-Fushman, W.~J. Rogers, A.~R. Aronson, Metamap lite: an evaluation of
  a new java implementation of metamap, Journal of the American Medical
  Informatics Association 24~(4) (2017) 841--844.

\bibitem{spacy}
M.~Honnibal, I.~Montani, {spaCy 2}: Natural language understanding with {B}loom
  embeddings, convolutional neural networks and incremental parsing, to appear
  (2017).

\bibitem{herer}
P.~Xu, D.~Barbosa, \href{https://www.aclweb.org/anthology/N19-1323}{Connecting
  language and knowledge with heterogeneous representations for neural relation
  extraction}, in: Proceedings of the 2019 Conference of the North {A}merican
  Chapter of the Association for Computational Linguistics: Human Language
  Technologies, Volume 1 (Long and Short Papers), Association for Computational
  Linguistics, Minneapolis, Minnesota, 2019, pp. 3201--3206.
\newblock \href {http://dx.doi.org/10.18653/v1/N19-1323}
  {\path{doi:10.18653/v1/N19-1323}}.
\newline\urlprefix\url{https://www.aclweb.org/anthology/N19-1323}

\bibitem{reside}
S.~Vashishth, R.~Joshi, S.~S. Prayaga, C.~Bhattacharyya, P.~Talukdar,
  \href{http://aclweb.org/anthology/D18-1157}{{RESIDE}: Improving
  distantly-supervised neural relation extraction using side information}, in:
  Proceedings of the 2018 Conference on Empirical Methods in Natural Language
  Processing, Association for Computational Linguistics, Brussels, Belgium,
  2018, pp. 1257--1266.
\newline\urlprefix\url{http://aclweb.org/anthology/D18-1157}

\bibitem{partial_metric}
A.~Tong, L.~Diduch, J.~Fiscus, Y.~Haghpanah, S.~Huang, D.~Joy, K.~Peterson,
  I.~Soboroff, \href{https://doi.org/10.1007/s10590-017-9200-8}{Overview of the
  nist 2016 lorehlt evaluation}, Machine Translation 32~(1–2) (2018) 11–30.
\newblock \href {http://dx.doi.org/10.1007/s10590-017-9200-8}
  {\path{doi:10.1007/s10590-017-9200-8}}.
\newline\urlprefix\url{https://doi.org/10.1007/s10590-017-9200-8}

\bibitem{bert_as_service}
H.~Xiao, bert-as-service, \url{https://github.com/hanxiao/bert-as-service}
  (2018).

\bibitem{huggingface}
T.~{Wolf}, L.~{Debut}, V.~{Sanh}, J.~{Chaumond}, C.~{Delangue}, A.~{Moi},
  P.~{Cistac}, T.~{Rault}, R.~{Louf}, M.~{Funtowicz}, J.~{Brew}, {HuggingFace's
  Transformers: State-of-the-art Natural Language Processing}, arXiv e-prints
  (2019) arXiv:1910.03771\href {http://arxiv.org/abs/1910.03771}
  {\path{arXiv:1910.03771}}.

\bibitem{adam}
D.~P. {Kingma}, J.~{Ba}, {Adam: A Method for Stochastic Optimization}, arXiv
  e-prints (2014) arXiv:1412.6980\href {http://arxiv.org/abs/1412.6980}
  {\path{arXiv:1412.6980}}.

\bibitem{bootstrap_stat_sig}
T.~Berg-Kirkpatrick, D.~Burkett, D.~Klein,
  \href{https://www.aclweb.org/anthology/D12-1091}{An empirical investigation
  of statistical significance in {NLP}}, in: Proceedings of the 2012 Joint
  Conference on Empirical Methods in Natural Language Processing and
  Computational Natural Language Learning, Association for Computational
  Linguistics, Jeju Island, Korea, 2012, pp. 995--1005.
\newline\urlprefix\url{https://www.aclweb.org/anthology/D12-1091}

\bibitem{Kersloot2020}
M.~G. Kersloot, F.~J.~P. van Putten, A.~Abu-Hanna, R.~Cornet, D.~L. Arts,
  \href{https://doi.org/10.1186/s13326-020-00231-z}{{Natural language
  processing algorithms for mapping clinical text fragments onto ontology
  concepts: a systematic review and recommendations for future studies}},
  Journal of Biomedical Semantics 11~(1) (2020) 14.
\newblock \href {http://dx.doi.org/10.1186/s13326-020-00231-z}
  {\path{doi:10.1186/s13326-020-00231-z}}.
\newline\urlprefix\url{https://doi.org/10.1186/s13326-020-00231-z}

\bibitem{Chen2020c}
L.~Chen, Y.~Gu, X.~Ji, Z.~Sun, H.~Li, Y.~Gao, Y.~Huang,
  \href{https://doi.org/10.1093/jamia/ocz141}{{Extracting medications and
  associated adverse drug events using a natural language processing system
  combining knowledge base and deep learning}}, Journal of the American Medical
  Informatics Association 27~(1) (2020) 56--64.
\newblock \href {http://dx.doi.org/10.1093/jamia/ocz141}
  {\path{doi:10.1093/jamia/ocz141}}.
\newline\urlprefix\url{https://doi.org/10.1093/jamia/ocz141}

\bibitem{Oliveira2020}
C.~R. Oliveira, P.~Niccolai, A.~M. Ortiz, S.~S. Sheth, E.~D. Shapiro, L.~M.
  Niccolai, C.~A. Brandt, \href{https://medinform.jmir.org/2020/11/e20826
  https://doi.org/10.2196/20826
  http://www.ncbi.nlm.nih.gov/pubmed/32469840}{{Natural Language Processing for
  Surveillance of Cervical and Anal Cancer and Precancer: Algorithm Development
  and Split-Validation Study}}, JMIR Med Inform 8~(11) (2020) e20826.
\newblock \href {http://dx.doi.org/10.2196/20826} {\path{doi:10.2196/20826}}.
\newline\urlprefix\url{https://medinform.jmir.org/2020/11/e20826
  https://doi.org/10.2196/20826 http://www.ncbi.nlm.nih.gov/pubmed/32469840}

\bibitem{Afzal2018}
N.~Afzal, V.~P. Mallipeddi, S.~Sohn, H.~Liu, R.~Chaudhry, C.~G. Scott, I.~J.
  Kullo, A.~M. Arruda-Olson,
  \href{http://www.sciencedirect.com/science/article/pii/S1386505617304756}{{Natural
  language processing of clinical notes for identification of critical limb
  ischemia}}, International Journal of Medical Informatics 111 (2018) 83--89.
\newblock \href
  {http://dx.doi.org/https://doi.org/10.1016/j.ijmedinf.2017.12.024}
  {\path{doi:https://doi.org/10.1016/j.ijmedinf.2017.12.024}}.
\newline\urlprefix\url{http://www.sciencedirect.com/science/article/pii/S1386505617304756}

\bibitem{DeBruijn2011}
B.~de~Bruijn, C.~Cherry, S.~Kiritchenko, J.~Martin, X.~Zhu,
  \href{https://doi.org/10.1136/amiajnl-2011-000150}{{Machine-learned solutions
  for three stages of clinical information extraction: the state of the art at
  i2b2 2010}}, Journal of the American Medical Informatics Association 18~(5)
  (2011) 557--562.
\newblock \href {http://dx.doi.org/10.1136/amiajnl-2011-000150}
  {\path{doi:10.1136/amiajnl-2011-000150}}.
\newline\urlprefix\url{https://doi.org/10.1136/amiajnl-2011-000150}

\bibitem{Wei2020}
Q.~Wei, Z.~Ji, Z.~Li, J.~Du, J.~Wang, J.~Xu, Y.~Xiang, F.~Tiryaki, S.~Wu,
  Y.~Zhang, C.~Tao, H.~Xu, \href{https://doi.org/10.1093/jamia/ocz063}{{A study
  of deep learning approaches for medication and adverse drug event extraction
  from clinical text}}, Journal of the American Medical Informatics Association
  27~(1) (2020) 13--21.
\newblock \href {http://dx.doi.org/10.1093/jamia/ocz063}
  {\path{doi:10.1093/jamia/ocz063}}.
\newline\urlprefix\url{https://doi.org/10.1093/jamia/ocz063}

\bibitem{Hassanpour2016}
S.~Hassanpour, C.~P. Langlotz,
  \href{https://www.sciencedirect.com/science/article/pii/S0933365715001244}{{Information
  extraction from multi-institutional radiology reports}}, Artificial
  Intelligence in Medicine 66 (2016) 29--39.
\newblock \href
  {http://dx.doi.org/https://doi.org/10.1016/j.artmed.2015.09.007}
  {\path{doi:https://doi.org/10.1016/j.artmed.2015.09.007}}.
\newline\urlprefix\url{https://www.sciencedirect.com/science/article/pii/S0933365715001244}

\bibitem{Thieu2021}
T.~Thieu, J.~C. Maldonado, P.-S. Ho, M.~Ding, A.~Marr, D.~Brandt,
  D.~Newman-Griffis, A.~Zirikly, L.~Chan, E.~Rasch,
  \href{https://www.sciencedirect.com/science/article/pii/S1386505620318876}{{A
  comprehensive study of mobility functioning information in clinical notes:
  Entity hierarchy, corpus annotation, and sequence labeling}}, International
  Journal of Medical Informatics 147 (2021) 104351.
\newblock \href
  {http://dx.doi.org/https://doi.org/10.1016/j.ijmedinf.2020.104351}
  {\path{doi:https://doi.org/10.1016/j.ijmedinf.2020.104351}}.
\newline\urlprefix\url{https://www.sciencedirect.com/science/article/pii/S1386505620318876}

\bibitem{bender2021on}
E.~M. Bender, T.~Gebru, A.~McMillan-Major, S.~Shmitchell,
  \href{https://doi.org/10.1145/3442188.3445922}{{On the Dangers of Stochastic
  Parrots: Can Language Models Be Too Big?}}, in: Proceedings of the 2021 ACM
  Conference on Fairness, Accountability, and Transparency, FAccT '21,
  Association for Computing Machinery, New York, NY, USA, 2021, pp. 610--623.
\newblock \href {http://dx.doi.org/10.1145/3442188.3445922}
  {\path{doi:10.1145/3442188.3445922}}.
\newline\urlprefix\url{https://doi.org/10.1145/3442188.3445922}

\bibitem{Obermeyer2019}
Z.~Obermeyer, B.~Powers, C.~Vogeli, S.~Mullainathan,
  \href{http://science.sciencemag.org/content/366/6464/447.abstract}{{Dissecting
  racial bias in an algorithm used to manage the health of populations}},
  Science 366~(6464) (2019) 447 LP -- 453.
\newblock \href {http://dx.doi.org/10.1126/science.aax2342}
  {\path{doi:10.1126/science.aax2342}}.
\newline\urlprefix\url{http://science.sciencemag.org/content/366/6464/447.abstract}

\bibitem{blodgett2017racial}
S.~L. Blodgett, B.~O'Connor, {Racial disparity in natural language processing:
  A case study of social media african-american english}, arXiv preprint
  arXiv:1707.00061.

\bibitem{Koenecke2020}
A.~Koenecke, A.~Nam, E.~Lake, J.~Nudell, M.~Quartey, Z.~Mengesha, C.~Toups,
  J.~R. Rickford, D.~Jurafsky, S.~Goel,
  \href{http://www.pnas.org/content/117/14/7684.abstract}{{Racial disparities
  in automated speech recognition}}, Proceedings of the National Academy of
  Sciences 117~(14) (2020) 7684 LP -- 7689.
\newblock \href {http://dx.doi.org/10.1073/pnas.1915768117}
  {\path{doi:10.1073/pnas.1915768117}}.
\newline\urlprefix\url{http://www.pnas.org/content/117/14/7684.abstract}

\bibitem{Ginther2011}
D.~K. Ginther, W.~T. Schaffer, J.~Schnell, B.~Masimore, F.~Liu, L.~L. Haak,
  R.~Kington, \href{https://pubmed.ncbi.nlm.nih.gov/21852498
  https://www.ncbi.nlm.nih.gov/pmc/articles/PMC3412416/}{{Race, ethnicity, and
  NIH research awards}}, Science 333~(6045) (2011) 1015--1019.
\newblock \href {http://dx.doi.org/10.1126/science.1196783}
  {\path{doi:10.1126/science.1196783}}.
\newline\urlprefix\url{https://pubmed.ncbi.nlm.nih.gov/21852498
  https://www.ncbi.nlm.nih.gov/pmc/articles/PMC3412416/}

\bibitem{Adams2019}
J.~Adams, H.~Br{\"{u}}ckner, C.~Naslund,
  \href{https://doi.org/10.1177/2378023118823946}{{Who Counts as a Notable
  Sociologist on Wikipedia? Gender, Race, and the “Professor Test”}},
  Socius 5 (2019) 2378023118823946.
\newblock \href {http://dx.doi.org/10.1177/2378023118823946}
  {\path{doi:10.1177/2378023118823946}}.
\newline\urlprefix\url{https://doi.org/10.1177/2378023118823946}

\bibitem{Ezell2021}
J.~M. Ezell, \href{https://doi.org/10.1080/01419870.2020.1851383}{{Empathy
  plasticity: decolonizing and reorganizing Wikipedia and other online spaces
  to address racial equity}}, Ethnic and Racial Studies 44~(8) (2021)
  1324--1336.
\newblock \href {http://dx.doi.org/10.1080/01419870.2020.1851383}
  {\path{doi:10.1080/01419870.2020.1851383}}.
\newline\urlprefix\url{https://doi.org/10.1080/01419870.2020.1851383}

\bibitem{Garla2011}
V.~Garla, V.~L. Re, Z.~Dorey-Stein, F.~Kidwai, M.~Scotch, J.~Womack,
  A.~Justice, C.~Brandt, {The Yale cTAKES extensions for document
  classification: Architecture and application}, Journal of the American
  Medical Informatics Association 18~(5) (2011) 614--620.
\newblock \href {http://dx.doi.org/10.1136/amiajnl-2011-000093}
  {\path{doi:10.1136/amiajnl-2011-000093}}.

\bibitem{Garla2012}
V.~N. Garla, C.~Brandt, {Knowledge-based biomedical word sense disambiguation:
  an evaluation and application to clinical document classification}, Journal
  of the American Medical Informatics Association (2012) 882--886\href
  {http://dx.doi.org/10.1136/amiajnl-2012-001350}
  {\path{doi:10.1136/amiajnl-2012-001350}}.

\bibitem{Osborne2013}
J.~D. Osborne, B.~Gyawali, T.~Solorio, {Evaluation of YTEX and MetaMap for
  clinical concept recognition}, CEUR Workshop Proceedings 1179.
\newblock \href {http://arxiv.org/abs/1402.1668} {\path{arXiv:1402.1668}}.

\bibitem{henry-etal-2017-evaluating}
S.~Henry, C.~Cuffy, B.~McInnes,
  \href{https://www.aclweb.org/anthology/W17-2334}{Evaluating feature
  extraction methods for knowledge-based biomedical word sense disambiguation},
  in: {B}io{NLP} 2017, Association for Computational Linguistics, Vancouver,
  Canada,, 2017, pp. 272--281.
\newblock \href {http://dx.doi.org/10.18653/v1/W17-2334}
  {\path{doi:10.18653/v1/W17-2334}}.
\newline\urlprefix\url{https://www.aclweb.org/anthology/W17-2334}

\bibitem{antunes2017supervised}
R.~Antunes, S.~Matos,
  \href{https://www.degruyter.com/view/journals/jib/14/4/article-20170051.xml}{Supervised
  learning and knowledge-based approaches applied to biomedical word sense
  disambiguation}, Journal of Integrative Bioinformatics 14~(4) (01 Dec. 2017)
  20170051.
\newblock \href {http://dx.doi.org/https://doi.org/10.1515/jib-2017-0051}
  {\path{doi:https://doi.org/10.1515/jib-2017-0051}}.
\newline\urlprefix\url{https://www.degruyter.com/view/journals/jib/14/4/article-20170051.xml}

\bibitem{DUQUE20189}
A.~Duque, M.~Stevenson, J.~Martinez-Romo, L.~Araujo,
  \href{http://www.sciencedirect.com/science/article/pii/S0933365717303627}{Co-occurrence
  graphs for word sense disambiguation in the biomedical domain}, Artificial
  Intelligence in Medicine 87 (2018) 9 -- 19.
\newblock \href
  {http://dx.doi.org/https://doi.org/10.1016/j.artmed.2018.03.002}
  {\path{doi:https://doi.org/10.1016/j.artmed.2018.03.002}}.
\newline\urlprefix\url{http://www.sciencedirect.com/science/article/pii/S0933365717303627}

\bibitem{Zhang2019a}
C.~Zhang, D.~Bi{\'{s}}, X.~Liu, Z.~He,
  \href{https://doi.org/10.1186/s12859-019-3079-8}{{Biomedical word sense
  disambiguation with bidirectional long short-term memory and attention-based
  neural networks}}, BMC Bioinformatics 20~(16) (2019) 502.
\newblock \href {http://dx.doi.org/10.1186/s12859-019-3079-8}
  {\path{doi:10.1186/s12859-019-3079-8}}.
\newline\urlprefix\url{https://doi.org/10.1186/s12859-019-3079-8}

\bibitem{feng-etal-2020-explainable}
J.~Feng, C.~Shaib, F.~Rudzicz,
  \href{https://aclanthology.org/2020.emnlp-main.115}{Explainable clinical
  decision support from text}, in: Proceedings of the 2020 Conference on
  Empirical Methods in Natural Language Processing (EMNLP), Association for
  Computational Linguistics, Online, 2020, pp. 1478--1489.
\newblock \href {http://dx.doi.org/10.18653/v1/2020.emnlp-main.115}
  {\path{doi:10.18653/v1/2020.emnlp-main.115}}.
\newline\urlprefix\url{https://aclanthology.org/2020.emnlp-main.115}

\bibitem{Rudin2019}
C.~Rudin, \href{https://doi.org/10.1038/s42256-019-0048-x}{{Stop explaining
  black box machine learning models for high stakes decisions and use
  interpretable models instead}}, Nature Machine Intelligence 1~(5) (2019)
  206--215.
\newblock \href {http://dx.doi.org/10.1038/s42256-019-0048-x}
  {\path{doi:10.1038/s42256-019-0048-x}}.
\newline\urlprefix\url{https://doi.org/10.1038/s42256-019-0048-x}

\end{thebibliography}

\end{document}

\newcommand{\refalg}[1]{Algorithm \ref{#1}}
\newcommand{\refeqn}[1]{Equation \ref{#1}}
\newcommand{\reffig}[1]{Figure \ref{#1}}
\newcommand{\reftbl}[1]{Table \ref{#1}}
\newcommand{\refsec}[1]{Section \ref{#1}}

\newcommand{\reminder}[1]{\textcolor{red}{[[ #1 ]]}\typeout{#1}}
\newcommand{\reminderR}[1]{\textcolor{gray}{[[ #1 ]]}\typeout{#1}}

\newcommand{\add}[1]{\textcolor{red}{#1}\typeout{#1}}
\newcommand{\remove}[1]{\sout{#1}\typeout{#1}}

\newcommand{\m}[1]{\mathcal{#1}}
\newcommand{\bmm}[1]{\bm{\mathcal{#1}}}
\newcommand{\real}[1]{\mathbb{R}^{#1}}

\newcommand{\method}{\textsc{MedType}}

\newcommand{\ctakes}{cTAKES}
\newcommand{\scispacy}{ScispaCy}
\newcommand{\metamap}{MetaMap}
\newcommand{\metamaplite}{MetaMapLite}
\newcommand{\quickumls}{QuickUMLS}

\newcommand{\ncbi}{NCBI}
\newcommand{\cdr}{Bio CDR}
\newcommand{\sharecorpus}{ShARe}
\newcommand{\medmen}{MedMentions}
\newcommand{\wiki}{\textsc{WikiMed}}
\newcommand{\pubmed}{\textsc{PubMedDS}}

\newcommand{\basecnn}{Type-CNN}
\newcommand{\basefc}{DeepType-FC}
\newcommand{\basernn}{DeepType-RNN}
\newcommand{\baseattn}{AttentionNER}

\newcommand{\problem}{DD}
\newcommand{\problemfull}{Document Dating}

\newtheorem{theorem}{Theorem}[section]
\newtheorem{claim}[theorem]{Claim}

\newcommand{\tensor}{\mathcal{X}}
\newcommand{\Real}{\mathbb{R}}

\newcommand{\tuples}{\mathbb{T}}

\newcommand{\argmax}{arg\,max}

\newcommand\norm[1]{\left\lVert#1\right\rVert}

\newcommand{\note}[1]{\textcolor{blue}{#1}}

\newcommand*{\Scale}[2][4]{\scalebox{#1}{$#2$}}%
\newcommand*{\Resize}[2]{\resizebox{#1}{!}{$#2$}}%

\newcommand{\medlinker}{MedLinker}

%%% Tensor
%\DeclareMathAlphabet\ten{OMS}{cmsy}{b}{n} %%usage: \mathbfcal{W}
%% Matrix
\def\mat#1{\mbox{\bf #1}}%% usage: \mat{W}.

	\maketitle
	\beginsupplement
	\appendix
	\appendix

\section{Implementation Details}

\textbf{Online Demo \& medtype-as-service:} Along with providing a step-by-step guide for reproducing all the results reported in the paper, we also provide code for running an online demo of \method. We also provide a scalable implementation of \method{} called medtype-as-service which is based on \textit{bert-as-service} \cite{bert_as_service} for processing thousands of documents simultaneously. 

\textbf{Hyperparameters:} We use pre-trained weights of BioBERT \cite{biobert} for initializing BERT component of \method{}. \method{} is implemented using HuggingFace Transformers library \cite{huggingface}. For training, we utilize Adam optimizer \cite{adam} with a learning rate in range $(10^{-3}, 10^{-5})$. The window size of context ($k$) is chosen from $\{48,64,128\}$. The best hyperparameters were selected based on the performance on the validation split of the datasets. We use the default hyperparameters for all the entity linkers and components of \method{}. A grid search over the validation split was performed for deciding a threshold for each semantic type from the range of (0.001, 1). The area under the Precision-Recall curve was used for choosing the best threshold. 

\textbf{Training Details:} All training was performed on NVIDIA-GTX 1080Ti GPUs. Each training epoch of \method{} takes from 5 mins to 2 days depending on the size of the dataset. In terms of number of parameters, \method{} has around 110 million parameters (same as BERT-base model). 

\section{Medical Entity Linkers}
\label{sec:linker_desc}
In this section, we present a short description of the different medical entity linking systems used in this work for probing the impact of incorporating a type-based candidate concepts filtering step. 
\begin{itemize}[itemsep=1pt,parsep=0pt,partopsep=0pt,leftmargin=10pt,topsep=1pt]
	
	\item \textbf{\metamap{}} \cite{metamap}  leverages a knowledge-intensive approach based on symbolic NLP and linguistic techniques to map biomedical mentions in text to UMLS concepts. 
	\item \textbf{\ctakes{}} \cite{ctakes} uses a terminology-agnostic dictionary look-up algorithm for mapping named entities to UMLS concepts. We utilize the Clinical Pipeline of \ctakes{} augmented with LVG Annotator\footnote{https://cwiki.apache.org/confluence/display/\ctakes{}/ \ctakes{}+4.0+-+LVG}. 
	\item \textbf{\metamaplite{}} \cite{metamaplite} re-implements the basic functionalities of \metamap{} with an additional emphasis on real-time  processing and competitive performance. 
	\item \textbf{\quickumls{}} \cite{quickumls}  is a fast, unsupervised algorithm that leverages approximate, dictionary-matching techniques for mapping biomedical entities in text.
	\item \textbf{\scispacy{}} \cite{scispacy} builds upon the robust spaCy library \cite{spacy} for several biomedical and scientific text-processing applications such as parsing, named entity recognition, and entity linking.
	\item \textbf{MedLinker} \cite{Medlinker} utilize approximate dictionary matching heuristics with
	specialized neural language models for medical entity typing and
	entity linking in biomedical literature.
\end{itemize}

\begin{table}
	\centering
	\small
	\begin{tabular}{lcccc}
		\toprule
		\multicolumn{1}{c}{}  & \multicolumn{1}{c}{\textbf{\ncbi{}}}         & \multicolumn{1}{c}{\textbf{\cdr{}}}         & \multicolumn{1}{c}{\textbf{\sharecorpus{}}}             & \multicolumn{1}{c}{\textbf{MedMent}} \\
		\midrule
		\textbf{MetaMap} & 9.0 & 20.7 & 8.1 & 39.8 \\
		\midrule
		Oracle (Fine) & +0.5 & +0.3 & +0.3 & +6.9 \\
		Oracle (Coarse) & +0.5 & +0.3 & +0.3 & +6.1 \\
		\basecnn{} & +0.5 & +0.2 & +0.2 & +3.8 \\
		\midrule
		\method{}  & +0.5 & +0.2 & +0.3 & +4.2 \\
		\midrule \midrule
		\textbf{cTakes} & 21.6 & 42.1 & 18.7 & 18.3 \\
		\midrule
		Oracle (Fine) & +0.2 & +0.1 & +0.1 & +0.2 \\
		Oracle (Coarse) & +0.2 & +0.1 & +0.1 & +0.2 \\
		\basecnn{} & +0.2 & +0.1 & +0.0 & +0.1 \\
		\midrule
		\method{}  & +0.2 & +0.1 & +0.0 & +0.1 \\
		\midrule \midrule
		\textbf{MetaMapLite} & 7.3 & 18.9 & 7.5 & 35.2 \\
		\midrule
		Oracle (Fine) & +1.6 & +1.5 & +1.5 & +7.8 \\
		Oracle (Coarse) & +1.6 & +1.5 & +1.5 & +6.5 \\
		\basecnn{} & +1.6 & +1.3 & +1.3 & +4.0 \\
		\midrule
		\method{}  & +1.6 & +1.4 & +1.4 & +4.6 \\
		\midrule \midrule
		\textbf{QuickUMLS} & 7.2 & 13.2 & 5.8 & 10.1 \\
		\midrule
		Oracle (Fine) & +0.9 & +3.7 & +3.1 & +2.9 \\
		Oracle (Coarse) & +0.9 & +3.2 & +3.1 & +2.4 \\
		\basecnn{} & +0.6 & +2.6 & +2.4 & +1.3 \\
		\midrule
		\method{}  & +0.8 & +3.1 & +2.8 & +1.7 \\
		\midrule \midrule
		\textbf{ScispaCy} & 12.7 & 24.4 & 10.2 & 40.6 \\
		\midrule
		Oracle (Fine) & +2.0 & +2.0 & +2.0 & +9.4 \\
		Oracle (Coarse) & +2.0 & +2.0 & +2.0 & +7.8 \\
		\basecnn{} & +1.8 & +1.1 & +1.6 & +3.9 \\
		\midrule
		\method{}  & +1.9 & +1.8 & +1.8 & +4.6 \\
		\midrule \midrule
		\textbf{MedLinker} & 8.7 & 25.9 & 7.5 & 26.5 \\
		\midrule
		Oracle (Fine) & +1.5 & +3.5 & +2.6 & +3.6 \\
		Oracle (Coarse) & +1.5 & +3.3 & +2.6 & +2.8 \\
		\basecnn{} & +0.8 & +1.1 & +1.8 & +0.2 \\
		\midrule
		\method{}  & +1.3 & +2.8 & +2.1 & +0.9\\
		\bottomrule
	\end{tabular}
	\caption{\label{tab:partial_full}
		We report the F1-score for partial mention and entity identifier match (Section \ref*{sec:evaluation}). For each method, the first row is its base performance and following rows indicates the change in scores with different entity disambiguation methods.}
\end{table}

\section{Discussion}
\label{sec:discussion}

\noindent \textbf{Combining \method{} with Biomedical WSD methods}
Disambiguating the candidate concepts produced by medical entity extraction pipelines has been a long-standing area of research, with several tools developed to integrate with existing pipelines.  The YTEX suite of algorithms \cite{Garla2011,Garla2012} extends both \metamap{} and \ctakes{} with a disambiguation module that helps to reduce noise considerably, although \citet{Osborne2013} found that it often over-filtered correct concepts.  These methods can be combined with \method{} to create a multi-stage filtering approach for disambiguation.  \method{} performs coarse filtering to a high-confidence set based on predicted type, a key step for narrowing down over-generated candidate sets in open-ended deep learning systems; disambiguation methods can then perform a fine-grained selection of the correct candidate to improve further entity linking performance.  We highlight this as an important direction for future work on medical entity linking.

\textbf{Entity Linking with \method{}: }
We present the results on entity linking with \method{} using \textit{Partial\_mention\_id\_match} evaluation metric as described in Section \ref*{sec:evaluation}.
The results are reported in Table \ref{tab:partial_full}. Similar to Section \ref*{sec:results_main}, we found that \method{} helps to improve medical entity linking across all the settings and is  comparable to the coarse-grained oracle and the best performing baseline \basecnn{}.

\textbf{Type-wise Analyzing of Improvement from \pubmed{}:}
Here, we present the type-wise results for demonstrating the benefit of utilizing our proposed datasets: \wiki{} and \pubmed{} for medical entity linkers. The results are reported for all four dataset: \ncbi{} \cdr{}, \sharecorpus{}, and MedMentions in Table \ref{tab:typewise_perf}. Overall, we find that utilizing \pubmed{} help improve performance across most of the semantic types. The gains are more substantial for the  types which have less coverage in the original training data. 

\begin{table*}
	\centering
	\small
		\begin{tabular}{lrrrrr}
			\hline
			\toprule
			
			\textbf{Categories}    & \textbf{\ncbi{}}    & \textbf{\cdr{}}    & \textbf{\sharecorpus{}}    & \textbf{\medmen{}}    & \textbf{\pubmed{}} \\
			\midrule
			Activities \& Behaviors             &    4      & 7      & 1      & 12,249 & 2,725,161 \\
			Anatomy                             &    3      & 29     & 4      & 19,098 & 10,688,138 \\
			Chemicals \& Drugs                  &    0      & 32,436 & 1      & 46,420 & 44,476,957 \\
			Concepts \& Ideas                   &    0      & 0      & 1      & 60,475 & 5,274,354 \\
			Devices                             &    0      & 0      & 0      & 2,691  & 242,599 \\
			Disease or Syndrome                 &    10,760 & 22,603 & 5,895  & 11,709 & 9,846,667 \\
			Disorders                           &    664    & 1,853  & 997    & 3,575  & 1,115,186 \\
			Finding                             &    749    & 2,220  & 500    & 15,666 & 1,778,023 \\
			Functional Concept                  &    0      & 0      & 1      & 23,672 & 48,553 \\
			Genes \& Molecular Sequences        &    20     & 0      & 0      & 5,582  & 281,662 \\
			Living Beings                       &    0      & 43     & 7      & 31,691 & 21,339,662 \\
			Mental or Behavioral Dysfunction    &    293    & 3,657  & 410    & 2,463  & 2,353,547 \\
			Neoplastic Process                  &    4,022  & 2,301  & 323    & 4,635  & 1,476,843 \\
			Objects                             &    0      & 129    & 2      & 10,357 & 5,184,355 \\
			Occupations                         &    0      & 0      & 0      & 1,443  & 654,604 \\
			Organic Chemical                    &    0      & 90,428 & 1      & 10,258 & 50,248,085 \\
			Organizations                       &    0      & 0      & 0      & 2,276  & 298,119 \\
			Pathologic Function                 &    143    & 3,290  & 2,285  & 4,121  & 1,895,835 \\
			Pharmacologic Substance             &    0      & 90,872 & 1      & 11,935 & 50,696,769 \\
			Phenomena                           &    4      & 163    & 2      & 7,210  & 1,722,873 \\
			Physiology                          &    15     & 166    & 3      & 24,753 & 10,674,561 \\
			Procedures                          &    5      & 73     & 4      & 37,616 & 7,471,434 \\
			Qualitative Concept                 &    0      & 0      & 7      & 32,564 & 1,211,747 \\
			Sign or Symptom                     &    211    & 9,844  & 2,687  & 1,809  & 3,750,734 \\
			\bottomrule
		\end{tabular}
%	}
	\caption{Category distribution of different datasets. Overall, we find that our proposed dataset, \pubmed{}, gives more extensive coverage across all semantic types.}
	\label{tab:cat_distribution}
\end{table*}

\begin{table*}
	\centering
	\small
	\begin{tabular}{lcccccccc}
		\toprule
		\multicolumn{1}{c}{}  & \multicolumn{2}{c}{\textbf{\ncbi{}}}         & \multicolumn{2}{c}{\textbf{\cdr{}}}         & \multicolumn{2}{c}{\textbf{\sharecorpus{}}}             & \multicolumn{2}{c}{\textbf{MedMentions}} \\
		\cmidrule(r){2-3}  \cmidrule(r){4-5} \cmidrule(r){6-7} \cmidrule(r){8-9}
		\multicolumn{1}{c}{}  & MT & w/ PM         & MT & w/ PM & MT & w/ PM & MT & w/ PM \\
		\midrule
		Activities \& Behaviors              &    0      & 0      & 0      & 0      & 0      & 0      & 0.719  & 0.744 \\
		Anatomy                             &    0      & 0      & 0      & 0      & 0      & 0      & 0.813  & 0.865 \\
		Chemicals \& Drugs                   &    0      & 0      & 0.83   & 0.915  & 0      & 0      & 0.778  & 0.822 \\
		Concepts \& Ideas                    &    0      & 0      & 0      & 0      & 0      & 0      & 0.805  & 0.828 \\
		Devices                             &    0      & 0      & 0      & 0      & 0      & 0      & 0.522  & 0.555 \\
		Disease or Syndrome                 &    0.945  & 0.972  & 0.878  & 0.932  & 0.846  & 0.923  & 0.79   & 0.844 \\
		Disorders                           &    0.589  & 0.69   & 0.824  & 0.858  & 0.507  & 0.799  & 0.621  & 0.679 \\
		Finding                             &    0      & 0.468  & 0.596  & 0.861  & 0.475  & 0.825  & 0.548  & 0.585 \\
		Functional Concept                  &    0      & 0      & 0      & 0      & 0      & 0      & 0.767  & 0.772 \\
		Genes \& Molecular Sequences         &    0      & 0      & 0      & 0      & 0      & 0      & 0.678  & 0.72 \\
		Living Beings                       &    0      & 0      & 0      & 0.571  & 0      & 0      & 0.881  & 0.901 \\
		Mental or Behavioral Dysfunction    &    0.174  & 0.833  & 0.588  & 0.926  & 0.484  & 0.788  & 0.767  & 0.807 \\
		Neoplastic Process                  &    0.917  & 0.942  & 0.909  & 0.946  & 0.715  & 0.909  & 0.856  & 0.874 \\
		Objects                             &    0      & 0      & 0      & 0.464  & 0      & 0      & 0.723  & 0.757 \\
		Occupations                         &    0      & 0      & 0      & 0      & 0      & 0      & 0.467  & 0.584 \\
		Organic Chemical                    &    0      & 0      & 0.919  & 0.943  & 0      & 0      & 0.719  & 0.806 \\
		Organizations                       &    0      & 0      & 0      & 0      & 0      & 0      & 0.73   & 0.756 \\
		Pathologic Function                 &    0      & 0.824  & 0.596  & 0.902  & 0.746  & 0.859  & 0.656  & 0.701 \\
		Pharmacologic Substance             &    0      & 0      & 0.92   & 0.933  & 0      & 0      & 0.636  & 0.708 \\
		Phenomena                           &    0      & 0      & 0.333  & 0.938  & 0      & 0      & 0.511  & 0.615 \\
		Physiology                          &    0      & 0      & 0      & 0.637  & 0      & 0      & 0.727  & 0.773 \\
		Procedures                          &    0      & 0      & 0      & 0.444  & 0      & 0      & 0.771  & 0.803 \\
		Qualitative Concept                 &    0      & 0      & 0      & 0      & 0      & 0      & 0.828  & 0.841 \\
		Sign or Symptom                     &    0      & 0.833  & 0.464  & 0.899  & 0.806  & 0.947  & 0.721  & 0.751 \\
		\bottomrule
	\end{tabular}
	\caption{\label{tab:typewise_perf}
		Type-wise analysis of the impact on using \method{} with \pubmed{} on \ncbi{}, \cdr{}, \sharecorpus{}, and \medmen{} datasets. We report F1-score for each semantic type. MT denotes \method{} and w/ PM indicates \method{} pre-trained on \pubmed{}.}
\end{table*}
	
	\newpage
	\balance
	\bibliography{references}
	\bibliographystyle{aaai21}
	
	%\section{ Acknowledgments}